\PassOptionsToPackage{shortlabels,inline}{enumitem}
\documentclass[10pt,twocolumn,letterpaper]{article}
\usepackage{comment}
\usepackage{booktabs,multirow}

\usepackage{booktabs}       
\usepackage{multirow}       
\usepackage{graphicx}       
\usepackage{listings}       
\usepackage{xcolor}         
\usepackage{float}
\usepackage[most]{tcolorbox}
\usepackage{graphicx}   
\usepackage{enumitem}   
\usepackage{lipsum}     
\usepackage[most]{tcolorbox}
\tcbset{
  colback=gray!5,
  colframe=black,
  sharp corners,
  boxrule=0.5pt,
}
\tcbset{termstyle/.style={
  colback=gray!20,            
  coltext=green!60!black,     
  boxrule=0pt,                
  arc=0pt,
  boxsep=1pt,
  left=1pt, right=1pt, top=1pt, bottom=1pt,
  enhanced, breakable         
}}

\usepackage[pagenumbers]{cvpr}      

\usepackage{CJKutf8}

\newcommand{\red}[1]{{\color{red}#1}}

\definecolor{cvprblue}{rgb}{0.21,0.49,0.74}
\usepackage[pagebackref,breaklinks,colorlinks,allcolors=cvprblue]{hyperref}

\def\paperID{  } 
\def\confName{CVPR}
\def\confYear{2026}

\title{UNICBench: UNIfied Counting Benchmark for MLLM}

\author{
\vspace{0.1cm}
Chenggang Rong$^1$\footnotemark[1] \quad 
Tao Han$^2$\footnotemark[1] \quad 
Zhiyuan Zhao$^3$\footnotemark[1] \quad 
Yaowu Fan$^4$ \\ 
\vspace{0.1cm}
Jia Wan$^5$ \quad 
Song Guo$^2$ \quad 
Yuan Yuan$^1$ \quad 
Junyu Gao$^{1,}$\footnotemark[2] \\
\vspace{0.1cm}
{\small $^1$Northwestern Polytechnical University \quad $^2$Hong Kong University of Science and Technology} \\[-0.2ex]
{\small $^3$Institute of Artificial Intelligence (TeleAI), China Telecom \quad $^4$Sun Yat-sen University \quad $^5$Harbin Institute of Technology, Shenzhen}
}

\newcommand{\benchNameFull}{\textbf{UNIfied Counting Benchmark}} 
\newcommand{\benchNameShort}{\textbf{UNICBench}}

\begin{document}
\twocolumn[{
\renewcommand\twocolumn[1][]{#1}
\maketitle
\vspace{-4.0em}
\begin{center}
    \centering

    \includegraphics[width=1.0\linewidth]{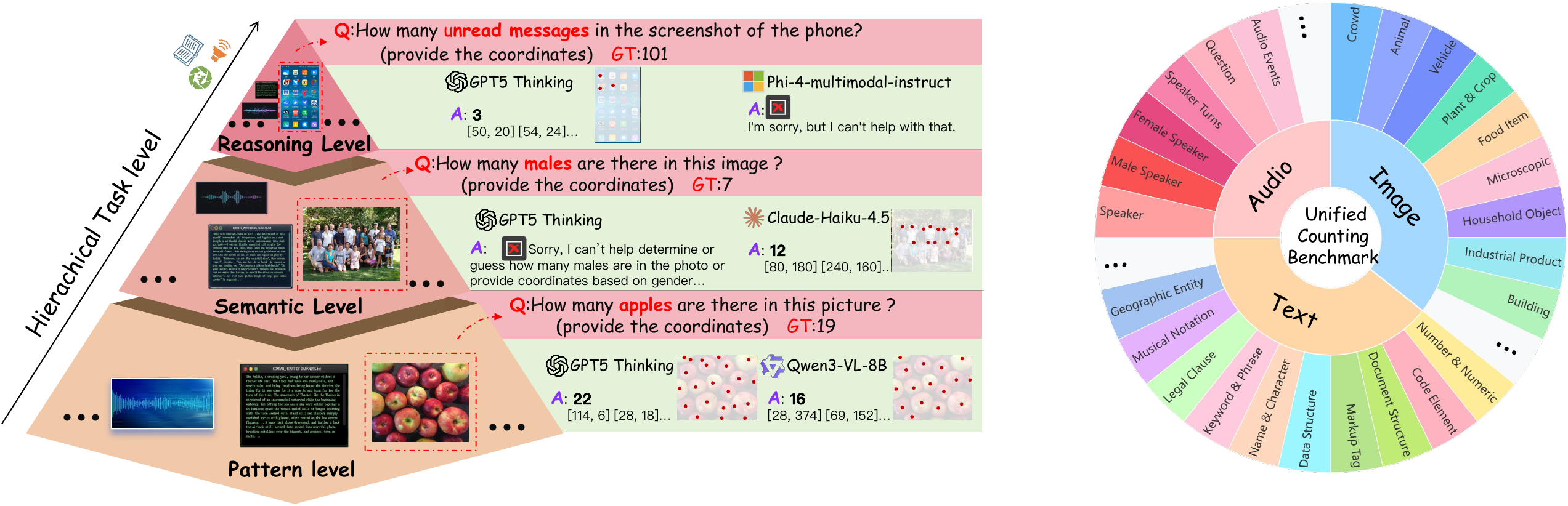}
    \vspace{-0.6cm}
    \captionof{figure}{\textbf{Illustration of the benchmark’s task taxonomy and dataset coverage}. The left pyramid groups counting problems by hierachical task level with representative Q/A examples. The right donut chart shows modality coverage and the diverse label categories in each modality, highlighting the benchmark’s broad cross‑modal and semantic coverage for unified counting evaluation. 
    }
    \vspace{-0.2cm}
    \label{fig:compare_top}
    \end{center}
    }]

{
  \makeatletter
  \def\blfootnote{\gdef\@thefnmark{}\@footnotetext}
  \makeatother
  
  \blfootnote{\hspace{-1em}$^*$Equal contribution.}
  \blfootnote{\hspace{-1em}$^\dagger$Corresponding author: gjy3035@gmail.com} 
}

\begin{abstract}
Counting is a core capability for multimodal large language models (MLLMs), yet there is no unified counting dataset to rigorously evaluate this ability across image, text, and audio. We present UNICBench, a unified multimodal, multi‑level counting benchmark and evaluation toolkit with accurate ground truth, deterministic numeric parsing, and stratified reporting. The corpus comprises 5,300 images (5,508 QA), 872 documents (5,888 QA), and 2,069 audio clips (2,905 QA), annotated with a three‑level capability taxonomy and difficulty tags. Under a standardized protocol with fixed splits/prompts/seeds and modality‑specific matching rules, we evaluate 45 state‑of‑the‑art MLLMs across modalities. Results show strong performance on some basic counting tasks but significant gaps on reasoning and the hardest partitions, highlighting long‑tail errors and substantial headroom for improving general counting. UNICBench offers a rigorous and comparable basis for measurement and a public toolkit to accelerate progress.
\end{abstract}

\section{Introduction}
\label{sec:intro}

Counting is a core cognitive faculty and a key capability for multimodal large language models (MLLMs). It underlies number sense in humans and animals \cite{dehaene2011number,feigenson2004core}. As MLLMs progress toward more general, human‑like intelligence~\cite{achiam2023gpt,bai2025qwen2,wang2025internvl3,hong2025glm}, reliable and interpretable counting across modalities becomes a concrete behavioral probe. Evaluating counting is both a practical test for applications (e.g., intelligent retail~\cite{junfen2022retail}, security/surveillance~\cite{holla2024cascaded,gao2025dynamic,han2022dr,fan2025video}, scientometrics, audio analytics~\cite{amirreza2025tacnet}) and a measure of how MLLMs approximate human numerical cognition.

Despite substantial progress of MLLMs on diverse multimodal benchmarks~\cite{zhang2025kaid}, including general-level visual question answering \& reasoning (e.g., EvalQABench~\cite{zhao2024lova3}, SEED-Bench~\cite{li2024seed}), NoCaps~\cite{agrawal2019nocaps}), (DocVQA~\cite{mathew2021docvqa}, MMbench~\cite{liu2024mmbench}, MMMU~\cite{yue2024mmmu}), and scientific question answering \& reasoning (ScienceQA~\cite{saikh2022scienceqa},  MATH-Vision~\cite{wang2024measuring}, ChemBench~\cite{guo2023can},  SciBench~\cite{wang2023scibench}), a comprehensive, modality-spanning assessment of general counting remains lacking. In this paper, we benchmark MLLMs on diverse counting tasks that estimate the number of events, entities, or structural elements across images (e.g., people, vehicles, object instances~\cite{mf2024lvlm-count,xu2023zeroshotcounting,yuan2025distance,gao2019pcc}), text (e.g., words, named entities, citations, paragraphs~\cite{wang2023nermrc}), and audio (e.g., alarms, bird calls, percussive beats~\cite{sgouros2020novel}), highlighting three complementary facets: perceptual localization, cross‑span/multimodal aggregation and de‑duplication, and rule‑guided reasoning.

To rigorously assess how well today’s MLLMs perform on general counting, a benchmark must address four challenges: (i) \emph{modality and task coverage gaps}—key settings lack ready-to-use public data (e.g., audio events, image–text aligned counting, long-document structural element counts), requiring curated collection and human verification under licensing and privacy constraints; (ii) \emph{heterogeneous annotations and no MLLM-ready QA standard}—existing datasets mix points, boxes, density maps, timestamps, and text spans with ambiguous instance inclusion rules, necessitating canonical target definitions and conversion to a unified QA template; (iii) \emph{inconsistent evaluation protocols}—splits, prompts, decoding settings, matching rules, and randomness vary across works, hindering comparability and reproducibility; and (iv) \emph{limited model availability and evaluation cost}—open models with balanced modality coverage are scarce, while closed-source APIs impose monetary costs, rate limits, and long-context token overheads, complicating fair comparisons.

Addressing these gaps requires a benchmark that is comprehensive in coverage, standardized in format, and rigorous in evaluation. We introduce a UNIfied Counting Benchmark (UNICBench) with the following contributions:

\begin{itemize}
  \item We introduce the first unified \emph{multimodal, multi-level} general-counting benchmark for MLLMs, extending visual counting to text and audio and formalizing a three-level capability and three-level difficulty taxonomy. 
  \item We release a rigorously curated cross-modal corpus with evidence-first ground truth and a canonical schema for predictions and adapters, including 5{,}508 QA over 5{,}300 image samples, 5{,}888 QA over 872 text samples, and 2{,}905 QA over 2{,}069 audio clips.
  \item We evaluate state-of-the-art MLLMs (image: 21 models; text: 22 models; audio: 13 models) using unified protocols, revealing that current MLLMs perform well on simple tasks but require substantial improvement on reasoning and challenging tasks.
 The systematic analyses, consolidated reporting, and an open evaluation toolkit help facilitate future development of MLLM.
\end{itemize}

\begin{table*}[htbp]
  \centering
\scriptsize
  \caption{Representative counting examples organized by category (capability vs. difficulty), level, and modality. The leftmost column groups three rows under ``Capability level'' (L1--L3) and a compact summary row for ``Difficulty level'' (Easy/Medium/Hard).}
  \label{tab:levels_examples_grouped}
  \setlength{\tabcolsep}{3pt}
  \begin{tabular}{p{1cm} p{1.8cm} p{5cm} p{4.4cm} p{4.0cm}}
    \toprule
    Category & Level & Image & Text & Audio \\
    \midrule
    \multirow[c]{3}{*}{\shortstack[c]{\textbf{Capability}\\\textbf{level}}} 
      & Pattern (L1) &
      Q: ``How many \red{\textbf{people}} are visible?'' 
       &
      Q: ``How many times does \red{\textbf{`Figure'}} appear?''  &
      Q: ``How many \red{\textbf{drum hits}} in this clip?''  \\
    \addlinespace
      & Semantic  (L2) &
      Q: ``How many \red{\textbf{people}} are wearing \red{\textbf{red shirts}}?'' &
      Q: ``How many \red{\textbf{non-repeated citations}}?''  &
      Q: ``How many \red{\textbf{bird calls}} of \red{\textbf{species X}}?'' \\
    \addlinespace
      & Reasoning (L3) &
      Q: ``How many \red{\textbf{folders}} in the specified path shown in the screenshot have a \red{\textbf{modification date}} in the \red{\textbf{year 2022}}?''  &
      Q: ``Count \red{\textbf{references}} from \red{\textbf{2010 $\sim$ 2020}}, \red{\textbf{excluding appendices}}.'' &
      Q: ``In this audio, how many \red{\textbf{questions}} are \red{\textbf{asked}} in total?'' \\
    \midrule
    \multirow[c]{3}{*}{\shortstack[c]{\textbf{Difficulty}\\\textbf{level}}} 
      & Easy: \textbf{1--10}  &  low density, small occlusion &   short span, low repetition &sparse, low overlap\\
      & Medium: \textbf{11--100}&  moderate density, some occlusion& moderate repetition, some deduplication& moderate overlap, brief bursts\\
      & Hard: \textbf{$>$100}&  high density, severe occlusion & long documents, heavy deduplication & overlapping events, long audio \\
    \bottomrule
  \end{tabular}
  \vspace{-0.4cm}
\end{table*}

\section{Related Work}
\label{sec:relatedWork}


    

\subsection{Existing Datasets and Benchmarks}
\textbf{Image counting}. Image counting has traditionally focused on crowded scenes (e.g., ShanghaiTech~\cite{zhang2016mccnn}, UCF\_CC\_50~\cite{idrees2013multi}, NWPU‑Crowd~\cite{wang2020nwpu},GCC~\cite{wang2019learning}), vehicle/traffic counting (e.g., CARPK~\cite{hsieh2017drone}, PUCPR+~\cite{hsieh2017drone}), and general object counting datasets (e.g., FSC‑147~\cite{ranjan2021learning}). These datasets typically use density maps, point annotations, or bounding boxes and emphasize high‑density and heavy‑occlusion scenarios. However, they lack unified annotation formats (points/boxes/density) and consistent difficulty stratification, and few are naturally expressed in a question–answer format suitable for MLLMs~\cite{zhang2018learning,gao2025survey}.

\textbf{Text counting}. Textual counting tasks appear across information extraction, document understanding, and scientometrics: from simple word, keyword counts and sentence, paragraph statistics to deduplicated citation counts and chart, table element counts, which appear as subproblems in DocVQA~\cite{mathew2021docvqa} and ChartQA~\cite{amasry2022chartqa}. Existing document QA benchmarks provide rich structural reading tests, but most do not focus on the semantic deduplication or cross‑segment aggregation required by robust counting; annotations are typically stored as spans or indices and are not directly aligned to MLLM QA outputs~\cite{mathew2021docvqa}.

\textbf{Audio counting}. Audio datasets are dominated by sound event detection and classification collections (e.g., AudioSet~\cite{gemmeke2017audio}). Audio counting tasks are challenging due to temporal overlap, short event durations, and varying annotation granularity, which complicates unified cross‑modal evaluation~\cite{gemmeke2017audio,trott2017interpretable}.

\textbf{MLLM benchmarks}. Recent multimodal benchmarks (VQA series~\cite{negar2025wikimixqa,alfarano2025vqartbenchsemanticallyrichvqa,james2025microvqa}, DocVQA~\cite{mathew2021docvqa}, MMBench~\cite{liu2024mmbench}, MMMU~\cite{yue2024mmmu}, SEED‑Bench~\cite{li2024seed}, etc.) cover visual understanding, text recognition, structured reasoning, and audio–visual alignment. These benchmarks typically measure QA accuracy or retrieval metrics and do not systematically treat counting as a distinct capability across image/text/audio simultaneously; a gap remains for an integrated, modality‑spanning counting evaluation~\cite{liu2024mmbench,li2024seed,yue2024mmmu}.


\subsection{Existing Counting Methods}
\textbf{Classical counting.} In computer vision, counting approaches fall into density estimation (CSRNet~\cite{li2018csrnet}, ChfL~\cite{mirzaee2022chfl}, STEERER~\cite{han2023steerer}, et. al.), detection‑then‑count (LSC-CNN~\cite{sam2020locate}, TopoCount~\cite{abousamra2021topocount}, et. al.), point‑supervision methods (P2PNet~\cite{song2021rethinking}, CLTR~\cite{liang2022end}, APGCC~\cite{chen2024improving}, et. al.), and segmentation/ clustering based techniques. Density regression models (e.g., CSRNet variants) excel in ultra‑dense crowds but are less robust on unseen scenes; detection‑based pipelines (e.g., Faster‑RCNN~\cite{ren2015faster} + post‑processing) work well on separable instances but struggle with occlusion. In audio, analogous approaches include sound event detection and temporal segmentation; in text, methods rely on information extraction (NER, coreference resolution) and rule‑based aggregation.

\textbf{LLM‑based counting.} With large language and multimodal models (Qwen2.5-Omni\cite{xu2025qwen2}, Qwen3-Omni\cite{xu2025qwen3},  Qwen2-Audio~\cite{chu2024qwen2}, Phi-Omni-ST~\cite{hu2025phi}), researchers have explored prompting MLLMs for general counting, using chain‑of‑thought to elicit intermediate evidence, and building hybrid pipelines that combine detectors with an LLM for deduplication and rule aggregation. These approaches are flexible in zero‑ or few‑shot settings but suffer from unverifiable intermediate evidence, poorly calibrated numeric confidences, and persistent cross‑modal alignment failures~\cite{alonso2025visionlanguagemodelsstrugglealign,zhang2023multimodal}.


In summary, prior work provides numerous benchmarks and method families, but lacks a cross-modal, MLLM-centric benchmark that standardizes QA outputs, evidence reporting, and evaluation protocols for counting. Our work addresses this gap by proposing a unified benchmark.

\begin{figure*}[t]
\centering
\includegraphics[width=\linewidth]{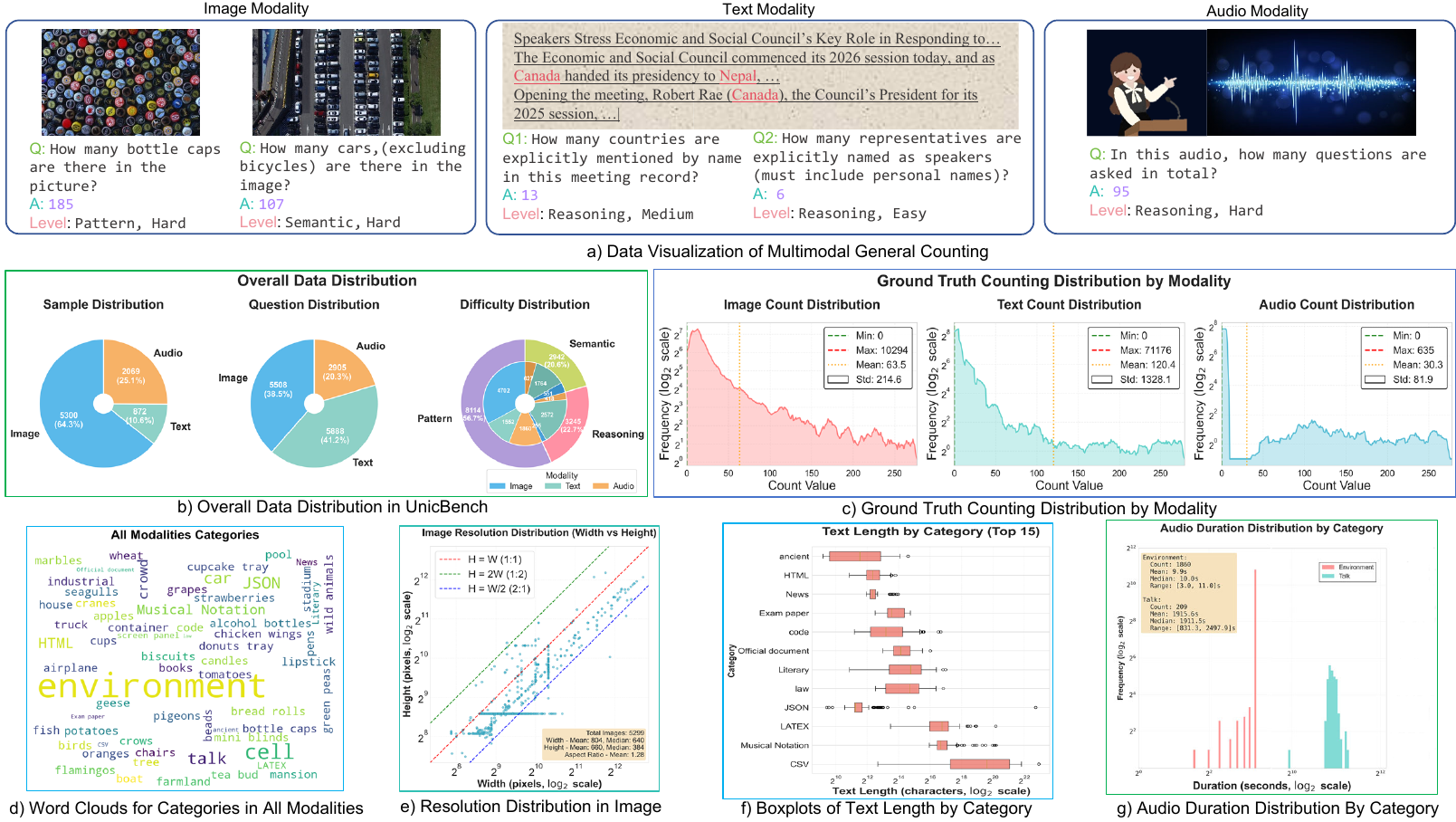}
  \caption{ a) Sample visualization in three modalities. b) Overview of dataset composition. Sample and question distributions across modalities and the capability/difficulty breakdown. c) Smoothed ground‑truth count distributions for three modalities, which are skewed and long‑tailed and thus motivate our stratified difficulty thresholds and evaluation protocol. d) Category word cloud based on question counts. e) f) g) Distribution of resolution/text length/ audio duration in three modalities, respectively. }
  \label{fig:combined}
  \vspace{-0.4cm}
\end{figure*}

\section{\benchNameFull}
\subsection{Benchmark Overview}
\label{sec:benchmark}

\paragraph{Task taxonomy.}
We organize all counting tasks into capability levels and difficulty tags. Table~\ref{tab:levels_examples_grouped} gives representative examples for each modality and capability level, illustrating typical question templates. To strictly demarcate these levels, we classify the query $Q$ based on its required operation on the entity set $E$:
\begin{itemize}
  \item \textbf{Pattern level (L1)} — perceptual counting: direct observation of instances/events suffices; no semantic filtering or rule application is required ($y = |E|$).
  \item \textbf{Semantic level (L2)} — attribute filtering and deduplication: $Q$ applies atomic filters $P$ on \textit{intrinsic attributes} (e.g., color, type) or performs identity aggregation (counting unique entities), yielding $y = |\{e \in E \mid P(e)\}|$.
  \item \textbf{Reasoning level (L3)} — rule‑driven and compositional counting: $Q$ imposes \textit{explicit rules} or \textit{structural constraints}, such as arithmetic/logical weights ($y = g(|S_1|, \dots)$) or temporal/complex interactions ($y=|\{e \mid \text{TempStruct}(e)\}|$).
\end{itemize}

In addition, each sample is also tagged with a difficulty label (\textbf{Easy/Medium/Hard}). As shown in Table~\ref{tab:levels_examples_grouped}, difficulty is quantified by measurable attributes (e.g., object number) to enable stratified analysis.

\paragraph{Question-Answer guidelines.}
To ensure consistency across modalities and levels, QA follows these rules:
\begin{itemize}
  \item \textbf{Level assignment}: Annotators assign L1–L3 strictly following the taxonomy defined above, ambiguous cases are adjudicated by lead annotators.
  \item \textbf{Difficulty quantification}: Compute objective measures (density, occlusion, overlap ratio, cross‑segment repetition) and map to Easy/Medium/Hard thresholds; store these measures in the data card for each sample.
  \item \textbf{Evidence-first GT}: Every ground truth includes both gt\_count and structured gt\_evidence (points/spans/timestamps).
  \item \textbf{Question templates}: We use deterministic templates for L1 tasks (to minimize linguistic variance), and more free‑form formulations for L2/L3 that specify filtering rules or aggregation constraints explicitly.
\end{itemize}

\subsection{Data Collection}
\label{sec:bench:data_col}
\textbf{Data Sources.} The benchmark aggregates data from multiple sources. Image samples are drawn from established counting datasets (FSC147, NWPU-MOC\cite{gao2024nwpu}, CARPK, JHU-CROWD++\cite{sindagi2020jhu}, UCF-QNRF\cite{idrees2018composition}, ShanghaiTech, IOCfish5K\cite{sun2023indiscernible}, Global Wheat Head Detection\cite{david2020global}, Snapshot Serengeti\cite{swanson2015snapshot}), with additional manual annotations for categories like screen panels, pens, birds, seagulls, books, biscuits, chairs, donuts tray, marbles, cups, mini blinds, potatoes, alcohol bottles, crows, green peas, and lipstick. Text data (5,888 questions across 872 samples) are entirely self-collected from diverse corpora including code repositories, legal documents, academic LaTeX files, and literary works. Audio samples leverage DESED \cite{turpault2019sound} for environmental sounds and AliMeeting \cite{Yu2022M2MeT} for conversational speech.

\textbf{Preprocessing Pipelines.} For images, we perform deduplication, quality filtering, and annotation format unification, converting point annotations to instance coordinates while preserving resolution diversity (from 234$\times$180 to 6736$\times$4640 pixels). Text preprocessing includes deduplication, segmentation, and character-span annotation, with ancient and literary texts retaining original linguistic structures. Audio processing applies noise reduction, temporal segmentation, and precise event timestamp alignment, ensuring sub-second temporal accuracy.

\textbf{Quality Control.} We employ a multi-stage verification protocol: dual independent annotation with arbitration for disagreements, achieving 100\% annotation consistency. Random sampling reviews ensure cross-modal annotation coherence. Questions incorporate deduplication challenges (cross-sentence entity counting), difficulty stratification via the three-level taxonomy, and adversarial prompts to test reasoning robustness.

\textbf{Ethics and Licensing.} All ata adhere to original licenses and are anonymized to protect privacy.

\subsection{Data Distribution}
\label{sec:bench:data_dis}
Figure~\ref{fig:combined} a) summarizes the corpus composition. Figure~\ref{fig:combined} b) presents the comprehensive dataset composition across modalities, sources, and difficulty levels. There are two distributional patterns: First, sample-level coverage and question‑level coverage are not identical. While image samples constitute the largest share of raw items, the number of questions is more evenly distributed across image and text modalities. This gap reflects our collection strategy of producing multiple, diverse QA probes per text sample (e.g., several citation/paragraph questions per paper) while keeping image samples closer to one QA per scene.

\begin{figure*}[htbp]
    \centering
    \includegraphics[width=\linewidth]{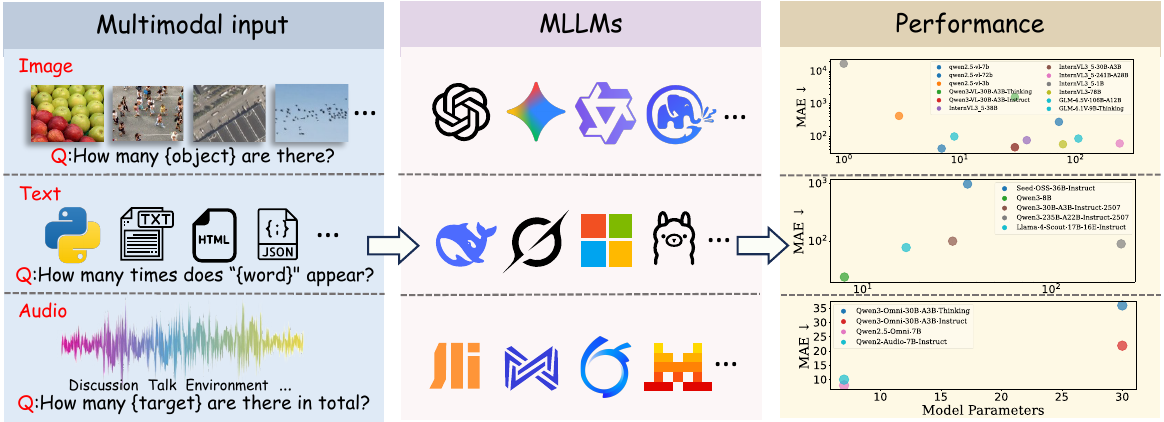}
    \caption{Overview of the \benchNameShort~pipeline. We standardize multi-modal datasets and define a unified QA-evidence schema with difficulty and capability labels. The end-to-end framework enables assessment of MLLM counting across image, text, and audio.}
    \label{fig:pipeline}
    \vspace{-0.4cm}
\end{figure*}

\textbf{Image counting track.} The 5,300 image samples (5,508 questions) exhibit diverse scene complexity. Count ranges span from sparse scenarios (airplane: average 5.58, stadium: 3.8) to extremely dense scenes (crowd: 355.58, tree: 301.96, fish: 135.12). Resolution distribution varies from 234$\times$180 to 6736$\times$4640 pixels (mean: 804$\times$660, aspect ratio: 1.28), accommodating both high-resolution remote sensing and standard digital photography. Category distribution emphasizes everyday objects (30+ categories from FSC147), aerial imagery (12 categories from NWPU-MOC), dense crowds (150 samples), and specialized domains (cells, wildlife). Difficulty stratification concentrates on Pattern-level tasks (85.4\%), reflecting the predominance of direct visual counting, with Semantic (10.0\%) and Reasoning (4.6\%) levels addressing attribute filtering and spatial relationships.

\textbf{Text counting track.} The 872 text samples (5,888 questions) display extreme length variability (584--8,052,974 characters, median: 7,176). Categories distribute across code (100 samples), HTML (168), JSON (200), Musical Notation (143), LaTeX (66), News (60), Literary (60), CSV (15), Exam papers (15), Official documents (15), ancient texts (20), and law (10). Ancient and Literary categories preserve original language (classical Chinese and source language respectively), totaling 80 bilingual samples (9.2\%). Count distributions exhibit high variance (mean: 120.40, median: 4.0, max: 71,176), with LaTeX and CSV categories contributing extreme values. Difficulty distribution uniquely favors Reasoning-level (43.7\%) and Semantic-level (29.9\%) tasks, demanding structural understanding and cross-reference deduplication.

\textbf{Audio counting track.} The audio track (2,069 samples, 2,905 QA) covers environmental (1,860 samples from DESED dataset~\cite{turpault2019sound}) and conversational speech (209 samples from AliMeeting~\cite{Yu2022M2MeT}). Durations range from 3.0 to 2,497.9 seconds (median: 10.0). Environmental audio exhibits sparse event density (1.56 events/sample), while conversational audio shows dense segmentation (81.51 counts/sample). All audio maintains WAV format with varied sampling rates preserving source characteristics. Difficulty balances Pattern-level (64.0\%) direct event counting with Semantic/Reasoning-level (36.0\%) speaker analysis and turn-taking inference.

\label{sec:image_counting_eval}

\begin{table*}[!htbp]
  \centering
  \small
  \caption{\textbf{Benchmark results on the Image-modality counting track}. Metrics are: SuccessRate (\%), Hit rates (@100\%/@90\%@80\%), MAE/MSE for Overall, per-difficulty, and per-capability. MSE values are shown in scientific notation with one decimal.}
  \label{tab:image_final_metrics_sci_overall}
  \setlength{\tabcolsep}{5pt}
  \resizebox{\textwidth}{!}{
  \begin{tabular}{l
                  r         
                  r r r     
                  r r       
                  r r       
                  r r       
                  r r       
                  r r       
                  r r       
                  r r}      
    \toprule
    \multirow{2}{*}{Model} &
      \multirow{2}{*}{Success (\%)} &
      \multicolumn{3}{c}{Hit rate (\%)} &
      \multicolumn{2}{c}{Overall} &
      \multicolumn{2}{c}{Easy } &
      \multicolumn{2}{c}{Medium } &
      \multicolumn{2}{c}{Hard } &
      \multicolumn{2}{c}{Pattern } &
      \multicolumn{2}{c}{Reasoning } &
      \multicolumn{2}{c}{Semantic } \\
    \cmidrule(lr){3-5} \cmidrule(lr){6-7} \cmidrule(lr){8-9} \cmidrule(lr){10-11} \cmidrule(lr){12-13} \cmidrule(lr){14-15} \cmidrule(lr){16-17} \cmidrule(lr){18-19}
    & & @100\% $\uparrow$ & @90\% $\uparrow$ & @80\% $\uparrow$ & MAE $\downarrow$ & MSE $\downarrow$ & MAE $\downarrow$ & MSE $\downarrow$ & MAE $\downarrow$ & MSE $\downarrow$ & MAE $\downarrow$ & MSE $\downarrow$ & MAE $\downarrow$ & MSE $\downarrow$ & MAE $\downarrow$ & MSE $\downarrow$ & MAE $\downarrow$ & MSE $\downarrow$ \\
    \midrule
    Claude-Sonnet-4-20250514 
      & 100.0 & 14.9 & 27.8 & 42.4 
      & 78.1 & 1.1e6 
      & 5.4 & 6.1e3 
      & 16.8 & 4.5e3 
      & 444.6 & 7.0e6 
      & 68.8 & 1.2e6 
      & 4.4 & 5.7e1 
      & 191.6 & 4.1e5 \\
    Gemini-2.5-Flash 
      & 100.0 & 15.6 & 25.1 & 38.0 
      & 140.5 & 1.0e6 
      & 12.0 & 5.7e3 
      & 55.8 & 2.9e4 
      & 694.2 & 6.5e6 
      & 131.4 & 1.0e6 
      & 6.7 & 2.1e3 
      & 280.8 & 1.1e6 \\
    Gemini-2.5-Pro-Thinking 
      & 100.0 & \underline{20.3} & \underline{37.5} & \underline{51.3} 
      & 90.0 & 8.1e5 
      & 4.3 & 5.8e2 
      & 22.1 & 2.9e4 
      & 504.9 & 5.3e6 
      & 71.1 & 6.5e5 
      & 4.6 & 8.4e1 
      & 291.1 & 2.5e6 \\
    GPT-4o-mini 
      & 100.0 & 12.5 & 20.7 & 33.0 
      & 73.3 & 1.5e6 
      & 2.3 & \textbf{3.8e1} 
      & 15.0 & 5.0e2 
      & 424.6 & 9.7e6 
      & 72.7 & 1.7e6 
      & 5.3 & 7.4e1 
      & 109.8 & 5.7e4 \\
    GPT-5 
      & 100.0 & 16.8 & 32.3 & 48.3 
      & 54.1 & 2.0e6 
      & 2.5 & 2.1e2 
      & 11.0 & \textbf{3.1e2} 
      & 312.4 & 1.4e7 
      & 55.1 & 2.4e6 
      & 5.9 & 1.0e2 
      & \textbf{67.4} & 3.3e4 \\
    GPT-o3 
      & 100.0 & 18.5 & 33.6 & 49.7 
      & 49.0 & 5.0e5 
      & 2.8 & 6.6e2 
      & 11.2 & 4.5e2 
      & 277.1 & 3.3e6 
      & 44.3 & 5.4e5 
      & 4.4 & 8.5e1 
      & 109.2 & 3.3e5 \\
    GPT-4o 
      & 100.0 & 16.6 & 31.6 & 48.1 
      & 43.2 & 7.5e5 
      & 2.4 & 1.3e2 
      & 11.4 & 4.7e2 
      & 238.4 & 5.0e6 
      & 41.7 & 8.7e5 
      & 5.4 & 1.1e2 
      & \underline{73.7} & 4.7e4 \\
    o4-mini 
      & 100.0 & 17.1 & 32.3 & 48.7 
      & 42.9 & 9.4e5 
      & 2.2 & 8.1e1 
      & 10.7 & 4.6e2 
      & 239.1 & 6.2e6 
      & 39.1 & 1.1e6 
      & 4.1 & 6.9e1 
      & 92.5 & 2.0e5 \\
      GPT-5-mini 
      & 100.0 & 17.2 & 32.8 & 50.5 
      & \textbf{29.8} & 9.6e4 
      & 2.1 & 6.5e1 
      & \textbf{10.0} & \underline{3.4e2} 
      & \textbf{155.0} & 6.4e5 
      & \textbf{25.4} & 1.1e5 
      & 5.3 & 1.1e2 
      & 78.9 & \underline{2.7e4} \\
    \midrule
    GLM-4.1V-9B-Thinking 
      & 87.3 & 15.7 & 27.5 & 43.2 
      & 97.9 & 1.6e6 
      & 3.0 & 4.4e2 
      & 13.9 & 1.1e3 
      & 542.2 & 9.9e6 
      & 90.0 & 1.7e6 
      & \textbf{3.1} & \textbf{2.5e1} 
      & 207.5 & 1.7e6 \\
    GLM-4.5V-106B-A12B 
      & 100.0 & 16.0 & 28.4 & 44.1 
      & 84.5 & 3.5e6 
      & 2.7 & 2.4e2 
      & 12.4 & 1.0e3 
      & 509.8 & 2.3e7 
      & 89.5 & 4.1e6 
      & 4.3 & 6.5e1 
      & 78.8 & \textbf{2.3e4} \\
    InternVL3\_5-1B 
      & 100.0 & 9.3 & 15.1 & 26.1 
      & 16686.6 & 1.5e12 
      & 4.5 & 2.2e3 
      & 19.1 & 1.6e3 
      & 111186.9 & 1.0e13 
      & 19506.0 & 1.8e12 
      & 6.7 & 1.0e2 
      & 346.6 & 1.5e7 \\
    InternVL3\_5-38B 
      & 100.0 & 16.6 & 27.8 & 42.9 
      & 75.4 & 1.3e6 
      & 2.3 & 7.2e1 
      & 15.9 & 6.7e2 
      & 435.5 & 8.5e6 
      & 73.2 & 1.5e6 
      & 6.8 & 1.1e2 
      & 126.3 & 3.3e5 \\
    InternVL3\_5-241B-A28B 
      & 100.0 & \textbf{21.9} & \textbf{40.5} & \textbf{57.1}
      & 59.9 & 2.9e6 
      & \underline{1.9} & 5.4e1 
      & \underline{10.0} & 8.0e2 
      & 356.4 & 2.0e7 
      & 61.1 & 3.4e6 
      & 6.1 & 9.3e1 
      & 74.9 & 3.0e4 \\
    InternVL3-78B 
      & 99.6 & 17.8 & 28.8 & 47.9 
      & 56.6 & 2.1e6 
      & \textbf{1.7} & 4.0e1 
      & 12.5 & 5.0e2 
      & 326.0 & 1.4e7 
      & 54.3 & 2.4e6 
      & 4.9 & 7.0e1 
      & 99.7 & 1.6e5 \\
    InternVL3\_5-30B-A3B 
      & 100.0 & 19.1 & 32.4 & 44.6 
      & 46.1 & \textbf{4.5e4} 
      & 1.9 & 5.6e1 
      & 17.5 & 8.1e2 
      & 234.2 & \textbf{3.0e5} 
      & 41.0 & \underline{4.1e4} 
      & 6.4 & 1.0e2 
      & 108.3 & 1.0e5 \\
    Qwen3-VL-30B-A3B-Thinking 
      & 92.7 & 18.3 & 30.8 & 45.5 
      & 1583.1 & 1.1e9 
      & 2.3 & 1.3e2 
      & 20.2 & 1.2e4 
      & 10293.5 & 7.5e9 
      & 355.2 & 6.7e7 
      & \underline{3.5} & \underline{4.9e1} 
      & 13354.7 & 1.1e10 \\
    Qwen2.5-VL-3B 
      & 100.0 & 8.9 & 15.9 & 29.1 
      & 417.7 & 3.6e8 
      & 3.5 & 9.9e1 
      & 21.0 & 2.4e3 
      & 2696.1 & 2.4e9 
      & \underline{36.7} & \textbf{2.2e4}
      & 5.4 & 1.2e2 
      & 3860.4 & 3.6e9 \\
    Qwen2.5-VL-72B 
      & 100.0 & 17.4 & 32.8 & 48.8 
      & 274.8 & 1.9e8 
      & 2.4 & 9.6e1 
      & 12.3 & 9.8e2 
      & 1779.1 & 1.2e9 
      & 92.7 & 4.4e6 
      & 4.5 & 7.5e1 
      & 1953.2 & 1.8e9 \\
    Qwen3-VL-30B-A3B-Instruct 
      & 100.0 & 17.3 & 27.1 & 39.5 
      & 45.7 & \underline{7.3e4} 
      & 1.9 & \underline{3.8e1} 
      & 13.8 & 4.3e2 
      & 246.7 & \underline{4.8e5} 
      & 37.3 & 7.1e4 
      & 5.4 & 8.3e1 
      & 136.1 & 1.2e5 \\
    Qwen2.5-VL-7B 
      & 99.9 & 16.6 & 28.6 & 43.2 
      & \underline{41.8} & 1.1e5 
      & 2.0 & 5.9e1 
      & 13.6 & 4.6e2 
      & \underline{222.3} & 7.5e5 
      & 37.6 & 1.3e5 
      & 4.5 & 7.0e1 
      & 95.0 & 5.4e4 \\
    \bottomrule
  \end{tabular}
  } 
\end{table*}

\begin{figure*}[!htbp]
    \centering
\includegraphics[width=1.0\linewidth]{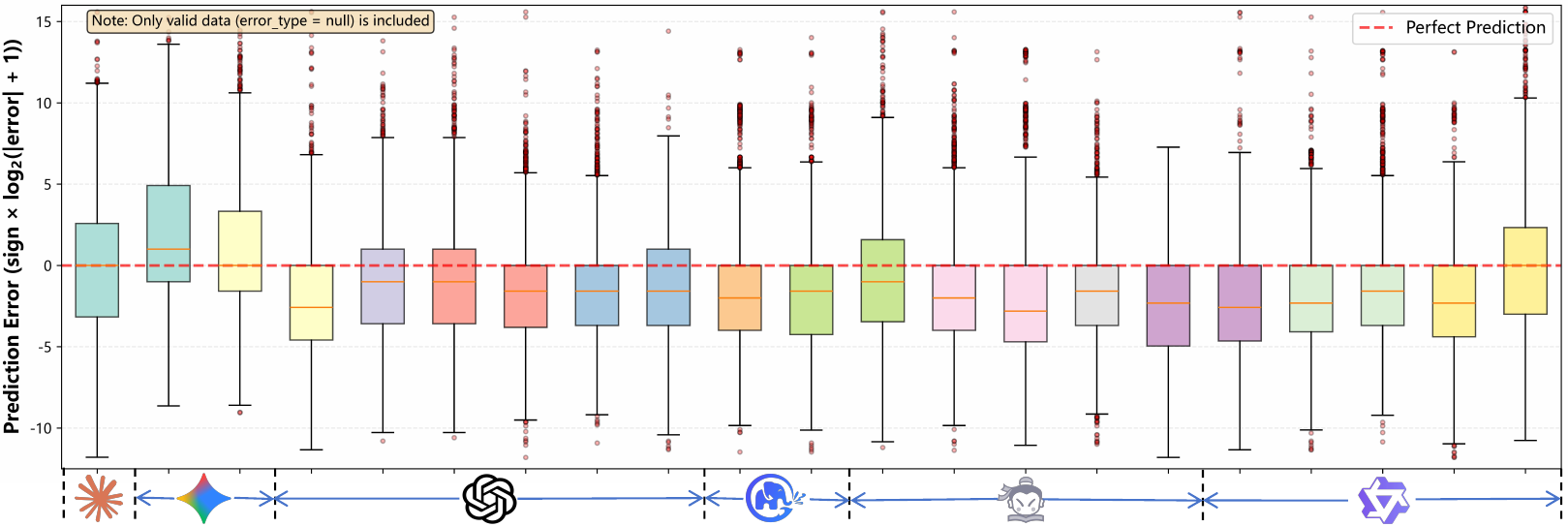}
    \vspace{-0.3cm}
    \caption{\textbf{Distribution of prediction error on image modality}. Whiskers and outliers indicate extreme failures—long whiskers or many outliers show a model makes severe errors on some samples (e.g., rare classes, label noise, or collapse cases). Such frequent extreme errors can substantially increase MAE/MSE even when the median error looks small.The model ordering is consistent with that in Table~\ref{tab:image_final_metrics_sci_overall}. }
  \label{fig:error_dist_image}
  \vspace{-0.5cm}
\end{figure*}

\section{Evaluation}
Figure~\ref{fig:pipeline} depicts an overview of the UNICBench. The evaluation is performed on three tracks with corresponding MLLMs. Following, we will introduce the evaluation protocols, results, and analysis of each track.    \emph{More details and settings are provided in the Supplementary.}

\subsection{Metrics}
\label{sec:metrics}
We evaluate counting performance with diverse metrics. Let \(N\) be the number of evaluation examples, \(y_i\) the ground‑truth count for example \(i\), and \(\hat{y}_i\) the model prediction parsed as a numeric value. If a model response is not parsable as a number (e.g., out-of-context, timeout, or error), we treat \(\hat{y}_i\) as invalid for metrics that require numeric predictions; such cases are only counted in Success Rate.

\textbf{Success rate (robustness).}
It measures the fraction of examples for which the model returns a parsable numeric prediction. Let $\mathbf{v}_i = 1$, if $\hat{y}_i$ is a valid numeric prediction, else $\mathbf{v}_{i}=0$, Then
\begin{equation}
      \mathrm{SuccessRate} \;=\; \frac{1}{N}\sum_{i=1}^{N} \mathbf{v}_i.
\end{equation}
All subsequent statistical metrics are based on the successful parsing of results.

\textbf{Mean Absolute Error and Mean Squared Error.}
\begin{equation}
  \mathrm{MAE} = \frac{1}{N}\sum_{i=1}^{N} \left|\hat{y}_i - y_i\right|,\quad
  \mathrm{MSE} = \frac{1}{N}\sum_{i=1}^{N} \left(\hat{y}_i - y_i\right)^2 .
\end{equation}



\textbf{Hit rate (tolerance based).}
For a relative tolerance \(\tau\) (e.g., \(\tau=0.10\) for 10\%), define the hit indicator
\begin{equation}
      \mathbf{1}_{i}^{(\tau)} \;=\;
  \begin{cases}
    1 & \text{if } \dfrac{\left|\hat{y}_i - y_i\right|}{\max(1,y_i)} \le \tau,\\[4pt]
    0 & \text{otherwise.}
  \end{cases}
\end{equation}

The Hit Rate at tolerance \(\tau\) is
\begin{equation}
    \mathrm{HitRate}@(1-\tau) \;=\; \frac{1}{N}\sum_{i=1}^{N} \mathbf{1}_{i}^{(\tau)}.
\end{equation}

We report HitRate@100\% (Estimated number is equal to annotation), HitRate@90\% and HitRate@80\%.

\subsection{General Setting}
\textbf{System Prompt.}
We use a consistent system prompt across all models to ensure a uniform understanding: 

\begin{tcolorbox}[termstyle]
\ttfamily\scriptsize
\textbf{You are a counting assistant. You MUST respond with ONLY a number. Never refuse to answer. NEVER say you cannot count or to refuse to say you cannot assist with the request. Always give your best numerical estimate. Respond with just the number, nothing else.}
\end{tcolorbox}

\textbf{Answer Extraction.}
During answer extraction we required that the model return a single numeric value. To accommodate models that produce internal deliberation tokens or proprietary wrappers, we accept (and strip) explicit thinking tags like \texttt{<think>} and final-answer delimiters like \texttt{<answer>} (and also common vendor wrappers such as GLM's \texttt{<begin\_of\_box>}…\texttt{</end\_of\_box>}). Our parser priority is: (1) extract a numeric token inside \texttt{<answer>} if present; (2) otherwise accept a standalone numeric token at the end of the response; (3) otherwise fall back to the first parseable numeric token in the output.

\begin{table*}[!htbp]
  \centering
  \small
  \caption{\textbf{Benchmark results on the Text-modality counting track}. Metrics are: SuccessRate (\%), Hit rates (@100\%/@90\%@80\%), MAE/MSE for Overall, per-difficulty, and per-capability. MSE values are shown in scientific notation with one decimal.}
  \label{tab:text_final_metrics_sci_overall}
  \setlength{\tabcolsep}{5pt}
  \resizebox{\textwidth}{!}{
  \begin{tabular}{l
                  r
                  r r r
                  r r
                  r r
                  r r
                  r r
                  r r
                  r r
                  r r}
    \toprule
    \multirow{2}{*}{Model} &
      \multirow{2}{*}{Success (\%)} &
      \multicolumn{3}{c}{Hit rate (\%)} &
      \multicolumn{2}{c}{Overall} &
      \multicolumn{2}{c}{Easy } &
      \multicolumn{2}{c}{Medium } &
      \multicolumn{2}{c}{Hard } &
      \multicolumn{2}{c}{Pattern } &
      \multicolumn{2}{c}{Reasoning } &
      \multicolumn{2}{c}{Semantic } \\
    \cmidrule(lr){3-5} \cmidrule(lr){6-7} \cmidrule(lr){8-9} \cmidrule(lr){10-11} \cmidrule(lr){12-13} \cmidrule(lr){14-15} \cmidrule(lr){16-17} \cmidrule(lr){18-19}
    & & @100\% $\uparrow$ & @90\% $\uparrow$ & @80\% $\uparrow$ & MAE $\downarrow$ & MSE $\downarrow$ & MAE $\downarrow$ & MSE $\downarrow$ & MAE $\downarrow$ & MSE $\downarrow$ & MAE $\downarrow$ & MSE $\downarrow$ & MAE $\downarrow$ & MSE $\downarrow$ & MAE $\downarrow$ & MSE $\downarrow$ & MAE $\downarrow$ & MSE $\downarrow$ \\
    \midrule
     Gemini-2.5-Flash-Nothinking
      & 99.0 & 37.3 & 43.7 & 50.5
      & 67.0 & 3.4e5
      & 7.4 & 2.6e4
      & 24.1 & 7.1e4
      & 746.3 & 4.1e6
      & 33.3 & 1.7e4
      & 120.1 & 7.2e5
      & 19.8 & 7.3e4 \\
    Gemini-2.5-Pro-Thinking
      & 99.2 & \textbf{63.3} & \textbf{72.8} & \textbf{76.1}
      & 30.5 & 9.0e4
      & 3.6 & 4.5e2
      & 11.3 & 1.0e4
      & \textbf{337.4} & \textbf{1.2e6}
      & 10.5 & 3.2e3
      & 56.4 & 2.0e5
      & 10.6 & 1.1e4 \\
    GPT-5-mini
      & 97.6 & 35.3 & 40.6 & 48.6
      & 1665.2 & 1.4e10
      & 2347.8 & 2.1e10
      & 17.9 & 1.1e4
      & 1321.4 & 2.9e7
      & 20.8 & 4.4e3
      & 3804.8 & 3.2e10
      & 8.1 & 2.2e3 \\
    GPT-o3
      & 96.8 & \underline{61.1} & \underline{68.2} & \underline{72.0}
      & 112.9 & 2.3e6
      & 3.5 & 4.7e2
      & 16.4 & 2.5e4
      & 1510.8 & 3.2e7
      & 15.5 & 1.0e4
      & 245.6 & 5.2e6
      & 5.6 & 1.5e3 \\
    GPT-5
      & 97.8 & 36.2 & 39.7 & 47.2
      & 106.9 & 1.2e6
      & 6.0 & 5.7e2
      & 16.6 & 2.5e3
      & 1393.6 & 1.7e7
      & 23.6 & 4.2e3
      & 225.8 & 2.7e6
      & 7.0 & 1.8e3 \\
    GPT-4o-mini
      & 97.2 & 35.5 & 37.4 & 41.6
      & 80.8 & 5.3e5
      & 2.9 & \underline{7.6e1}
      & 24.4 & 3.0e4
      & 1057.7 & 7.7e6
      & 29.3 & 5.8e3
      & 162.1 & 1.2e6
      & 8.2 & 3.1e3 \\
    GPT-4o
      & 97.1 & 39.8 & 41.8 & 46.2
      & 63.0 & 3.6e5
      & 4.6 & 3.8e2
      & 18.4 & 6.6e3
      & 811.6 & 5.3e6
      & 26.1 & 2.9e3
      & 122.1 & 8.3e5
      & 9.8 & 7.3e3 \\
    o4-mini
      & 97.6 & 59.6 & 67.1 & 70.2
      & 47.1 & 2.2e5
      & 4.3 & 3.6e2
      & \underline{9.8} & 1.1e3
      & 594.0 & 3.2e6
      & 15.1 & 1.9e3
      & 95.5 & 5.1e5
      & 5.0 & 1.6e3 \\
    Grok-4-Fast-Non-Reasoning
      & 94.5 & 37.6 & 42.2 & 47.0
      & 67.0 & 1.1e6
      & 5.7 & 5.5e2
      & 16.9 & 5.8e3
      & 895.3 & 1.7e7
      & 90.2 & 3.3e6
      & 90.6 & 5.1e5
      & 14.5 & 1.9e4 \\
    Claude-Sonnet-4-20250514
      & 97.6 & 37.7 & 44.3 & 50.4
      & 60.0 & 4.6e5
      & 3.7 & 3.5e2
      & 18.7 & 1.2e4
      & 753.4 & 6.5e6
      & 19.3 & 2.6e3
      & 120.9 & 1.0e6
      & 6.9 & 1.2e3 \\
    \midrule
    DeepSeek-V3.1-Nothinking
      & 97.1 & 40.1 & 46.0 & 51.2
      & 75.3 & 4.2e5
      & 3.2 & 4.0e2
      & 20.9 & 1.5e4
      & 998.2 & 6.2e6
      & 30.0 & 7.9e3
      & 149.7 & 9.7e5
      & 7.0 & 1.6e3 \\
    Deepseek-V3.2-exp
      & 97.5 & 34.7 & 38.1 & 43.9
      & 71.2 & 7.0e5
      & 5.1 & 6.5e2
      & 27.3 & 1.8e4
      & 876.5 & 1.0e7
      & 35.3 & 1.5e4
      & 135.4 & 1.6e6
      & 9.1 & 2.4e3 \\
    DeepSeek-V3.1
      & 94.2 & 47.3 & 51.7 & 55.8
      & 46.9 & 1.5e5
      & 2.8 & 1.3e2
      & 16.7 & 1.4e3
      & 653.0 & 2.5e6
      & 29.2 & 7.0e3
      & 86.5 & 3.5e5
      & 7.0 & 3.4e3 \\
    DeepSeek-R1-0528
      & 94.7 & 59.7 & 67.8 & 71.3
      & 34.6 & 8.5e4
      & 4.2 & 3.0e2
      & \textbf{8.6} & 6.9e2
      & \underline{469.0} & \underline{1.3e6}
      & 15.4 & 1.8e3
      & 68.0 & 2.0e5
      & \underline{4.4} & 9.0e2 \\
    GLM-4.6
      & 100.0 & 32.0 & 34.4 & 39.4
      & 151.3 & 2.8e6
      & 6.0 & 5.5e2
      & 30.6 & 3.1e4
      & 1915.3 & 3.8e7
      & 108.0 & 3.4e6
      & 272.3 & 4.4e6
      & 13.5 & 1.4e4 \\
    Llama-4-Scout-17B-16E-Instruct
      & 98.2 & 31.6 & 33.7 & 37.6
      & 76.2 & 6.2e5
      & 4.5 & 1.8e2
      & 17.4 & 9.8e2
      & 932.4 & 8.5e6
      & 37.2 & 2.3e4
      & 140.2 & 1.4e6
      & 18.8 & 2.8e4 \\
    Phi-4-Multimodal-Instruct
      & 88.4 & 21.3 & 22.5 & 26.6
      & 82.7 & 3.2e5
      & 10.0 & 2.3e3
      & 23.2 & 1.8e3
      & 1132.4 & 5.2e6
      & 115.1 & 3.1e5
      & 115.0 & 5.5e5
      & 7.7 & \underline{8.1e2} \\
    Phi-4
      & 69.7 & 41.9 & 42.6 & 45.3
      & \textbf{9.6} & \textbf{1.1e4}
      & \textbf{1.5} & \textbf{1.7e1}
      & 15.8 & \underline{4.1e2}
      & 551.3 & 1.4e6
      & \textbf{8.9} & \textbf{4.5e2}
      & \textbf{13.5} & \textbf{2.3e4}
      & 5.5 & 2.6e3 \\
    Qwen3-30B-A3B-Instruct-2507
      & 94.8 & 28.1 & 29.8 & 33.1
      & 98.4 & 9.3e5
      & 3.7 & 1.2e2
      & 31.4 & 1.6e4
      & 1207.8 & 1.3e7
      & 39.7 & 7.6e3
      & 194.1 & 2.1e6
      & 8.1 & 1.9e3 \\
    Qwen3-235B-A22B-Instruct-2507
      & 97.4 & 36.2 & 38.4 & 43.6
      & 88.2 & 1.4e6
      & 3.1 & 1.4e2
      & 23.4 & 2.0e4
      & 1161.0 & 2.0e7
      & 29.8 & 5.6e3
      & 180.4 & 3.1e6
      & 6.7 & 1.6e3 \\
    Qwen3-8B
      & 68.8 & 53.3 & 58.6 & 64.9
      & \underline{23.4} & \underline{7.3e4}
      & \underline{2.5} & 9.3e1
      & 10.4 & \textbf{3.1e2}
      & 675.1 & 2.6e6
      & \underline{9.1} & \underline{9.9e2}
      & \underline{46.9} & \underline{1.7e5}
      & \textbf{3.6} & \textbf{3.8e2} \\
    Seed-OSS-36B-Instruct
      & 95.6 & 35.3 & 37.8 & 42.3
      & 966.8 & 1.2e8
      & 9.7 & 2.1e3
      & 63.5 & 2.8e5
      & 13616.3 & 1.8e9
      & 115.8 & 2.9e5
      & 2193.7 & 2.9e8
      & 13.3 & 2.1e4 \\

    \bottomrule
  \end{tabular}
  } 
  \vspace{-0.3cm}
\end{table*}

\subsection{Image Counting Track}
\textbf{Setting.}
For the image-counting evaluation we enforced a unified runtime configuration across models to ensure comparability. Inference parameters were standardized to: 
\{max\_tokens:4096, temperature:0.0, timeout\_seconds: 120\}.
To balance responsiveness and computational cost, certain models were run with reduced reasoning intensity (e.g., GPT-5 and GPT-5-mini with \{reasoning\_effort: minimal, text\_verbosity: low\}; GPT-o3 set reasoning\_effort to low). 
A few models that do not expose temperature were executed at their default temperature (notably GPT-5, GPT-5-mini, GPT-4o-mini and GPT-o3 at temperature = 1.0). 

\textbf{Results.}
Table~\ref{tab:image_final_metrics_sci_overall} reports per-model performance on the image-counting track. Success rates are high (many models reach 100\%), indicating that producing numeric outputs per prompt is not the primary bottleneck. Hit rates rise with wider tolerance (100\%→90\%→80\%), yet only three models exceed a 50\% hit rate at the 20\% error threshold (InternVL3\_5-241B-A28B, Gemini-2.5-Pro-Thinking, and GPT-5-mini), indicating that current MLLMs still have substantial room to improve absolute counting accuracy. Error trends across difficulty levels are monotonic: MAE and MSE increase from Easy to Medium to Hard. Top-tier closed-source models (e.g., GPT-5-mini, GPT-o4-mini) show less relative error inflation on Hard, indicating greater robustness, while several open‑source models (e.g., Qwen2.5-VL-7B and InternVL3\_5-30B-A3B) remain competitive in dense scenes. By capability, Reasoning tasks yield the lowest numeric errors, largely because they often have small true counts, whereas Pattern and Semantic tasks remain challenging under dense packing or occlusion. High-density categories (e.g., crowd, tree, fish) and large scale variation primarily drive extreme errors. Increasing image resolution helps these cases but yields limited gains in ultra-dense settings due to representation bottlenecks.

\textbf{Analysis.}
Figure~\ref{fig:error_dist_image} presents each model's error distribution with boxplots. Frontier closed-source and large open models exhibit narrower interquartile ranges (IQRs) and fewer extreme outliers, whereas lightweight models display wider dispersion and heavier tails. Signed-error medians are often negative, and positive tails indicate rare large overestimates that inflate MAE and MSE. We attribute these patterns to three interacting factors:
\begin{itemize}
  \item \emph{Supervision gap.} Models are optimized for instruction following and description rather than dense instance-level counting, and thus lack per-instance supervision (e.g., points or detections) needed to resolve tightly clustered or occluded instances.
  \item \emph{Representation limits.} Patch-based, downsampled visual tokens compress small or densely packed objects, undermining one-to-one attention and increasing misses and duplicates.
  \item \emph{Calibration and distribution shift.} Decoding preferences favor rounding under uncertainty and occasionally overconfident predictions; combined with pretraining corpora that under-represent high-density or occluded domains, these factors amplify long-tailed errors.
\end{itemize}

\begin{figure}[tbp]
    \centering
    \includegraphics[width=1.0\linewidth]{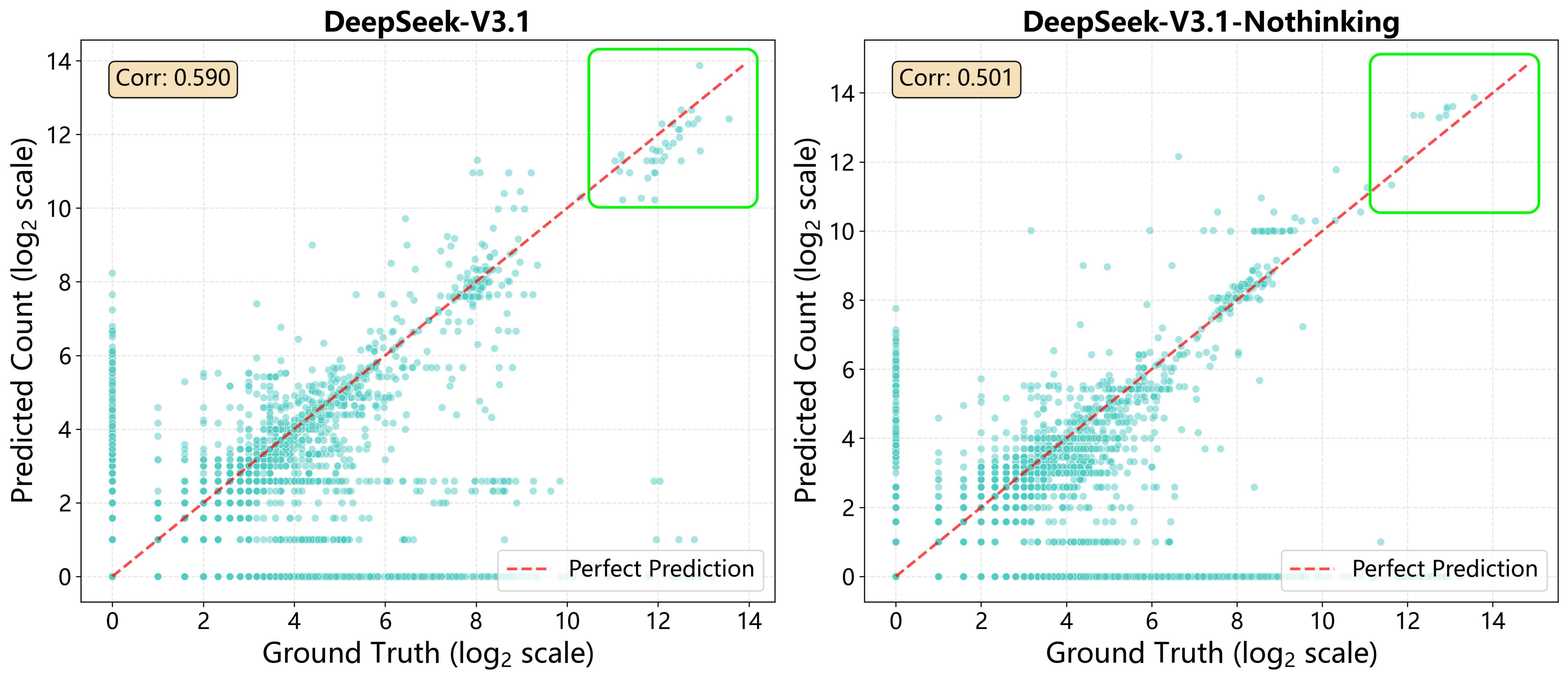}
    \caption{\textbf{Prediction vs. ground truth scatter.} Each point is one sample. The upper‑right green box highlights a small set of extreme high‑count samples where the thinking mode defeated non‑thinking mode, which drives most of the overall MAE gap.}
  \label{fig:text_scatter}
  \vspace{-0.4cm}
\end{figure}

\begin{table*}[tbp]
  \centering
  \small
  \caption{\textbf{Benchmark results on the audio‑modality counting track}. Metrics are: SuccessRate (\%), Hit rates (@100\%/@90\%@80\%), MAE/MSE for Overall, per‑difficulty, and per‑capability. MSE values are shown in scientific notation with one decimal. A ``-'' indicates that no valid values were obtained for this statistical dimension.}
  \label{tab:audio_final_metrics_sci_overall}
  \setlength{\tabcolsep}{5pt}
  \resizebox{\textwidth}{!}{
  \begin{tabular}{l
                  r         
                  r r r     
                  r r       
                  r r       
                  r r       
                  r r       
                  r r       
                  r r       
                  r r}      
    \toprule
    \multirow{2}{*}{Model} &
      \multirow{2}{*}{Success (\%)} &
      \multicolumn{3}{c}{Hit rate (\%)} &
      \multicolumn{2}{c}{Overall} &
      \multicolumn{2}{c}{Easy } &
      \multicolumn{2}{c}{Medium } &
      \multicolumn{2}{c}{Hard } &
      \multicolumn{2}{c}{Pattern } &
      \multicolumn{2}{c}{Reasoning } &
      \multicolumn{2}{c}{Semantic } \\
    \cmidrule(lr){3-5} \cmidrule(lr){6-7} \cmidrule(lr){8-9} \cmidrule(lr){10-11} \cmidrule(lr){12-13} \cmidrule(lr){14-15} \cmidrule(lr){16-17} \cmidrule(lr){18-19}
    & & @100\% $\uparrow$ & @90\% $\uparrow$ & @80\% $\uparrow$ & MAE $\downarrow$ & MSE $\downarrow$ & MAE $\downarrow$ & MSE $\downarrow$ & MAE $\downarrow$ & MSE $\downarrow$ & MAE $\downarrow$ & MSE $\downarrow$ & MAE $\downarrow$ & MSE $\downarrow$ & MAE $\downarrow$ & MSE $\downarrow$ & MAE $\downarrow$ & MSE $\downarrow$ \\
    \midrule
    
    Gemini-2.5-Pro-Thinking
      & 64.0 & 31.0 & 31.0 & 31.0
      & 1.1 & 2.3e0
      & 1.1 & 2.3e0
      & - & -
      & - & -
      & 1.1 & 2.3e0
      & - & -
      & - & - \\
    Gemini-2.5-Flash-Nothinking
      & 64.0 & \textbf{48.0} & \textbf{48.0} & \textbf{48.0}
      & \underline{0.7} & \underline{1.0e0}
      & \underline{0.7} & \underline{1.0e0}
      & - & -
      & - & -
      & \underline{0.7} & \underline{1.0e0}
      & - & -
      & - & - \\
    GPT-4o-mini-Audio-Preview
      & 64.0 & 11.3 & 11.3 & 11.3
      & 1.7 & 4.1e0
      & 1.7 & 4.1e0
      & - & -
      & - & -
      & 1.7 & 4.1e0
      & - & -
      & - & - \\
    GPT-4o-Audio-Preview
      & 63.9 & 32.4 & 32.4 & 32.4
      & 0.9 & 1.3e0
      & 0.9 & 1.3e0
      & - & -
      & - & -
      & 0.9 & 1.3e0
      & - & -
      & - & - \\
    
    GPT-Audio
      & 64.0 & 33.9 & 33.9 & 33.9
      & 0.9 & 1.5e0
      & 0.9 & 1.5e0
      & - & -
      & - & -
      & 0.9 & 1.5e0
      & - & -
      & - & - \\
    GPT-Audio-mini
      & 63.0 & \underline{38.3} & \underline{38.3} & \underline{38.3}
      & \textbf{0.6} & \textbf{6.8e-1}
      & \textbf{0.6} & \textbf{6.8e-1}
      & - & -
      & - & -
      & \textbf{0.6} & \textbf{6.8e-1}
      & - & -
      & - & - \\
    \midrule
    Voxtral-mini
      & 99.8 & 36.1 & 36.1 & 36.1
      & 29.5 & 7.2e3
      & 1.8 & 5.1e1
      & 74.8 & 5.9e3
      & 222.2 & \underline{6.0e4}
      & 0.9 & 3.9e0
      & 194.3 & 5.0e4
      & 4.3 & 1.9e2 \\
    Voxtral-small
      & 73.4 & 29.5 & 29.5 & 29.5
      & 1.5 & 4.9e1
      & 1.1 & 2.1e0
      & 74.0 & 5.5e3
      & \textbf{122.6} & \textbf{1.5e4}
      & 1.1 & 1.9e0
      & \textbf{101.0} & \textbf{1.1e4}
      & \textbf{1.1} & \underline{3.7e0} \\

    Phi-4-Multimodal-Instruct
      & 63.7 & 2.0 & 2.0 & 2.0
      & 3.2 & 1.3e1
      & 3.2 & 1.3e1
      & - & -
      & - & -
      & 3.2 & 1.3e1
      & - & -
      & - & - \\
    Qwen2-Audio-7B-Instruct
      & 99.7 & 9.9 & 9.9 & 9.9
      & 30.1 & 7.3e3
      & 2.4 & 8.3e0
      & \textbf{70.3} & \textbf{5.2e3}
      & 223.4 & 6.1e4
      & 2.8 & 1.0e1
      & 194.5 & 5.0e4
      & \underline{1.3} & \textbf{2.4e0} \\
    Qwen3-Omni-30B-A3B-Instruct
      & 100.0 & 21.8 & 21.8 & 21.8
      & 29.6 & 7.2e3
      & 1.8 & 9.1e0
      & 74.0 & 5.8e3
      & 223.2 & 6.0e4
      & 1.0 & 2.0e0
      & 195.0 & 5.0e4
      & 4.0 & 3.0e1 \\
    Qwen2.5-Omni-7B
      & 100.0 & 7.9 & 7.9 & 7.9
      & 29.2 & 7.1e3
      & 1.8 & 4.6e0
      & \underline{70.9} & \underline{5.3e3}
      & \underline{220.7} & 6.0e4
      & 1.8 & 4.0e0
      & \underline{192.4} & \underline{4.9e4}
      & 1.8 & 6.4e0 \\
    Qwen3-Omni-30B-A3B-Thinking
      & 73.5 & 35.6 & 35.7 & 35.9
      & 14.7 & 8.6e3
      & 6.9 & 4.5e3
      & 79.5 & 4.7e4
      & 310.0 & 1.6e5
      & 0.9 & 1.8e0
      & 249.9 & 1.3e5
      & 59.5 & 4.4e4 \\

    \bottomrule
  \end{tabular}
  } 
  \vspace{-0.4cm}
\end{table*}

\subsection{Text Counting Track}

\textbf{Setting.}
Apart from the additional changes mentioned, other model settings were kept consistent with image counting track. A few models used the officially recommended temperature parameters:\{GLM-4.6:1.0, Phi-4:1.0, and Phi-4-Multimodal-Instruct:0.75\}. We also evaluated GLM-4.6 in a “non-thinking” (no internal reasoning) mode to assess the impact of explicit chain-of-thought behavior.

\textbf{Results.}
Table~\ref{tab:text_final_metrics_sci_overall} summarizes performance on the text counting track. Extraction success is high overall, whereas numeric accuracy varies widely. Across our runs, Phi-4, Qwen3-8B, and Gemini-2.5-Pro-Thinking achieve the lowest aggregate MAE/MSE; several GPT and Qwen variants are competitive, but a few models show anomalous biases that inflate errors. As with images, errors increase monotonically with difficulty (Easy→Medium→Hard). Closed-source models maintain lower MAE/MSE on Hard, indicating greater robustness, though some large open models (e.g., DeepSeek-R1-0528) also perform strongly. By capability, Semantic tasks yield the smallest errors, indicating that modern MLLMs handle semantic filtering and aggregation well. Pattern-level tasks requiring strict template or format matching are more error-prone, and Reasoning remains the principal bottleneck and the largest source of inter-model variance.

\textbf{Analysis.}
Figure~\ref{fig:text_scatter} compares DeepSeek-V3.1 and DeepSeek-V3.1-Nothinking in text counting. The models perform similarly on most samples, butextreme high-count cases drive the MAE gap; in the right panel's top-right green box, the thinking variant shows smaller errors. Consequently, its overall MAE is 46.9 versus 75.3 for the non-thinking model (Table~\ref{tab:text_final_metrics_sci_overall}), indicating that improvements on a few long-tail, large-count samples can shift aggregate error despite similar median performance.

\subsection{Audio Counting Track}
\textbf{Setting.} This track follows the same base runtime protocol as previous tracks. All models were evaluated using the common configuration \{max\_tokens:4096, temperature:0.0, timeout\_seconds:120\}.

\textbf{Results.}
Table~\ref{tab:audio_final_metrics_sci_overall} summarizes the audio-track results; only Qwen2.5-Omni-7B and Qwen3-Omni-30B-A3B-Instruct achieve perfect extraction success. Several models, including GPT-family variants and Phi-4-Multimodal-Instruct, record success rates near 60\%, yet their numeric outputs often attain competitive MAE/MSE. Audio-specialized GPT variants (e.g., GPT-Audio-mini, GPT-Audio) and Gemini models demonstrate strong numeric precision, consistent with audio-focused pretraining. Within the Voxtral family, Voxtral-mini outperforms Voxtral-small; Qwen Audio-enabled models exhibit high success but are prone to systematic overestimation, increasing aggregate error. Difficulty-stratified metrics are monotonic (Easy→Medium→Hard), with the largest inter-model differences on Hard examples; by capability, Pattern and Semantic tasks are relatively easy, whereas Reasoning tasks yield the greatest separation across models.

\begin{figure}[tbp]
    \centering
    \includegraphics[width=1.0\linewidth]{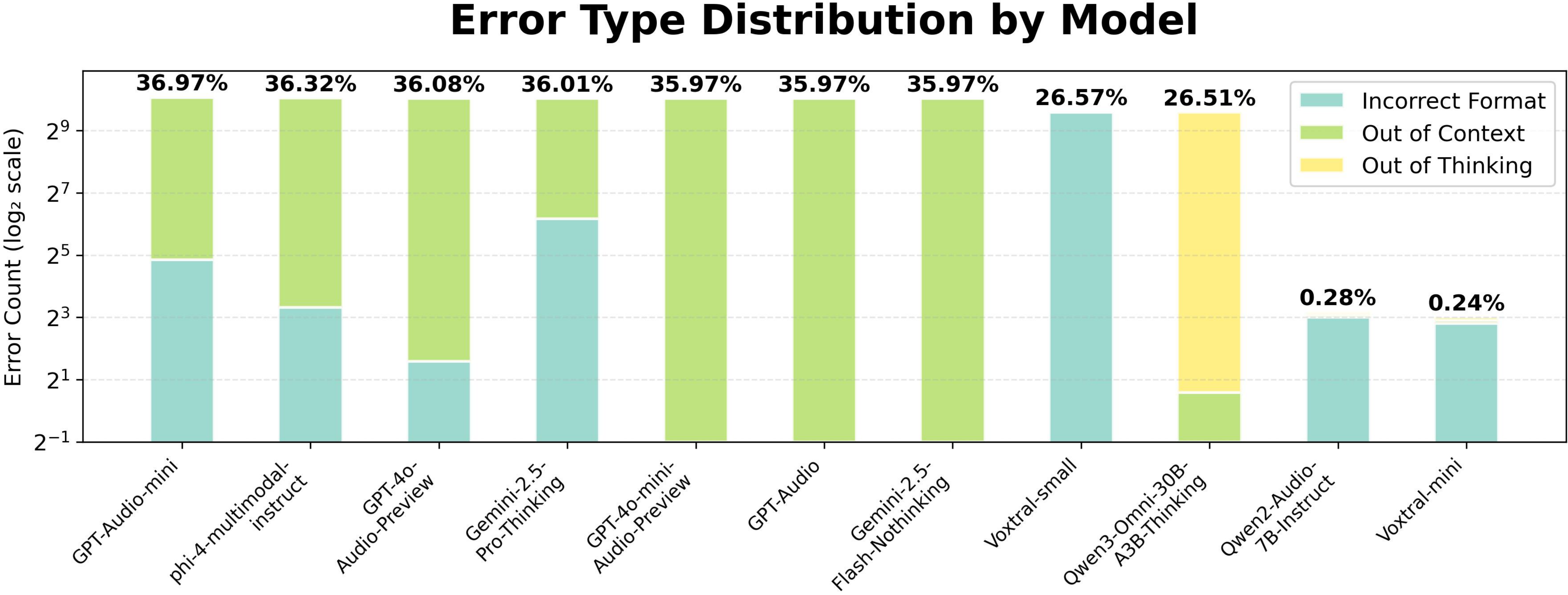}
    \vspace{-0.3cm}
  \caption{\textbf{Error-type analysis on the audio modality}. Models that refuse or fail to produce numeric outputs more often avoid counting those errors in MAE (but are penalized in SuccessRate).}
  \label{fig:error_dist_audio}
  \vspace{-0.3cm}
\end{figure}

\textbf{Analysis.}
Figure~\ref{fig:error_dist_audio} examines the relationship between refusal behavior and MAE. Models such as Qwen2.5-Omni-7B and Voxtral-mini achieve near-perfect extraction success rates (see Table~\ref{tab:audio_final_metrics_sci_overall}, 100.0\% and 99.8\%, respectively), and therefore they produce numeric answers even for the most challenging samples; consequently, their overall MAE lies in the tens (29.2 and 29.5), primarily driven by a small number of extreme, high-count cases. In contrast, several models with lower success rates (e.g., certain GPT-Audio variants) exhibit MAE values below 1 because many difficult items result in refusals, input failures, or non-numeric outputs, a pattern also reflected in the SuccessRate metric. Notably, Qwen2.5-Omni-7B's combination of a high SuccessRate and relatively low MAE demonstrates robust and accurate counting under our unified evaluation protocol.

\textit{For more evaluation results, details and analysis, please refer to the Evaluation Detail section in the Supplementary material.}

\section{Conclusion}
\label{sec:conclusion}
UNICBench is a unified multimodal counting benchmark spanning image, text, and audio with a three‑tier taxonomy (Pattern, Semantic, Reasoning), stratified difficulty thresholds, and evidence‑first ground truth; our experiments show that modern multimodal LLMs handle many Pattern and Semantic tasks well but still struggle on Reasoning and hardest partitions, with modality‑specialized models often offering higher numeric precision; based on these findings we recommend evidence‑first outputs, hybrid pipelines that combine detectors and VLLMs, and difficulty‑stratified reporting, while noting limitations from dataset bias and evolving tooling; future work will focus on improved cross‑modal alignment, tighter detector integration, and automated calibration for robust numeric reasoning.

\textbf{Acknowledgments.}
This work was supported in part by the National Natural Science Foundation of China under Grants 62306241 and U62576284.
{
    \small
    \bibliographystyle{ieeenat_fullname}
    \bibliography{main}
}

\clearpage
\setcounter{page}{1}
\maketitlesupplementary


\section{Benchmark Construction Details}
\label{sec:sup:data_col}

\subsection{Label Definition}
\label{subsec:sup:json_def}

\providecommand{\extbf}{\textbf} 

All modalities share a unified JSON structure with common fields (\texttt{type}, \texttt{target\_file\_path}, \texttt{attributes}, \texttt{questions}) and modality-specific extensions. Each question object contains bilingual descriptions, difficulty level, ground truth count, and an \texttt{instances} array. Instance annotations adapt to modality characteristics: images use $(x, y)$ pixel coordinates, text employs $[\text{start}, \text{end})$ character indices with matched text snippets, and audio records sound events with temporal ranges $[t_{\text{start}}, t_{\text{end}}]$ in seconds. This design enables consistent cross-modal evaluation while preserving modality-specific localization granularity.

The complete JSON schema is shown below:

\begin{lstlisting}[basicstyle=\footnotesize\ttfamily,breaklines=true,frame=single]
{
  "type": "image/text/audio",
  "target_file_path": "filename or false (for text)",
  "attributes": {
    "height": "int (image only)",
    "width": "int (image only)",
    "duration": "float (audio only, in seconds)",
    "format": "string (audio only)",
    "ori_file": "string (text only)",
    "sub_type": "string (text optional)"
  },
  "target_text": "string (text only)",
  "questions": [
    {
      "question_id": "int",
      "question": "English description",
      "question_cn": "Chinese description",
      "level": "Pattern/Semantic/Reasoning",
      "count": "int",
      "instances": [
        "Image: [x, y]",
        "Text: {text: ..., coordinates: {start: int, end: int}}",
        "Audio: {sound_type: ..., time_range: [start, end]}"
      ]
    }
  ]
}
\end{lstlisting}

\subsection{Construction details for different modality}
\textbf{Image Modality Construction.} The 5,300 image samples span 49 categories with varying source compositions. From FSC147, we extract 30+ everyday object categories (apples, beads, birds, biscuits, books, bottle caps, chairs, cups, etc.), each contributing 100 samples selected for count range diversity. NWPU-MOC provides remote sensing imagery (airplane, boat, farmland, house, industrial, mansion, pool, stadium, tree, truck) where objects exhibit scale variation and occlusion. Crowd counting integrates 150 samples from JHU-CROWD++, UCF-QNRF, ShanghaiTech Part A, and NWPU-Crowd, with manual supplementation for extreme density scenarios (average count: 355.58, max: 10,294). The medical domain contributes 400 cell images from Blood Cell Count and Detection Dataset (average: 50.06 cells/image). Notably, the screen panel category (50 samples, 55 questions) involves complete team annotation using custom labeling tools, while pens, birds, seagulls, books, biscuits, chairs, donuts tray, marbles, cups, mini blinds, potatoes, alcohol bottles, crows, green peas, and lipstick receive manual enhancement beyond FSC147's base annotations. Car samples merge CARPK and NWPU-MOC (100 each) to ensure scene diversity. Question design follows a stratified protocol: 99.1\% Pattern-level for direct visual counting, 0.5\% Semantic-level for attribute-specific queries, and 0.4\% Reasoning-level for relational constraints. Annotation verification employs cross-checking against original dataset ground truths, with point annotations converted to instance centers via bounding box centroids or density map peaks.

\textbf{Text Modality Construction.} The 872 text samples yielding 5,888 questions represent entirely self-collected data across 12 categories. Code samples (100 Java/Python files, 500 questions) target structural elements like functions, classes, and loops (average count: 12.21). JSON data (200 samples, 1,800 questions) emphasizes nested structures and key-value pairs (average: 58.87, max: 71,176 in LaTeX category with 66 samples and 593 questions due to citation and formula density). HTML samples (168 files, 1,176 questions) focus on tag hierarchies and attributes (average: 3.54), where each page is compressed into a content-preserving fragment by stripping redundant elements (e.g., embedded base64 images, SVG/Canvas objects) while retaining core structural nodes. Musical Notation leverages MusicXML format (143 files, 715 questions) to count notes, measures, and rests (average: 108.58). Ancient texts (20 samples, 96 questions) and Literary works (60 samples, 109 questions) use language-preserving questions matching source text style---classical Chinese for ancient documents and original language for literary texts. CSV data (15 files, 150 questions) involves large-scale tabular counting (average: 291.3). Legal documents (10 samples, 105 questions), Exam papers (15 samples, 120 questions), News articles (60 samples, 300 questions), and Official documents (15 samples, 224 questions) round out the corpus. Question distribution skews toward higher complexity: 26.4\% Pattern-level, 43.7\% Reasoning-level, and 29.9\% Semantic-level. Text length varies dramatically (584 to 8,052,974 characters, median: 7,176), requiring models to handle both short snippets and long documents. Annotators employ automated pre-counting tools (regex, syntax parsers) followed by manual verification, with each sample receiving dual annotation for counts exceeding 50 instances.

\textbf{Audio Modality Construction.} The 2,069 audio samples derive from two specialized datasets. DESED contributes 1,860 environmental sound samples (doors, alarms, dog barks, keyboard typing) with event-level timestamps, exhibiting sparse event density (average: 1.56 events/sample, duration: 3--300 seconds). AliMeeting provides 209 meeting recordings (1,045 questions) with dense speech segments (average: 81.51 counts/sample, duration: 10--2,497.9 seconds, median: 10.0). Temporal annotations achieve frame-level precision (0.01-second granularity), with sound\_type labels distinguishing speaker identities ("Speaker1\_unknown"), event categories ("Cat meowing"), and semantic units ("Question"). Audio preprocessing standardizes WAV format with original sampling rates preserved, applying minimal noise reduction to retain naturalistic characteristics. Question design balances perceptual tasks (64.0\% Pattern-level: counting discrete events) with semantic challenges (36.0\% Semantic/Reasoning-level: speaker identification, turn-taking analysis). 


\subsection{More Data Statistics}

\begin{figure}[tbp]
    \centering
    \includegraphics[width=1.0\linewidth]{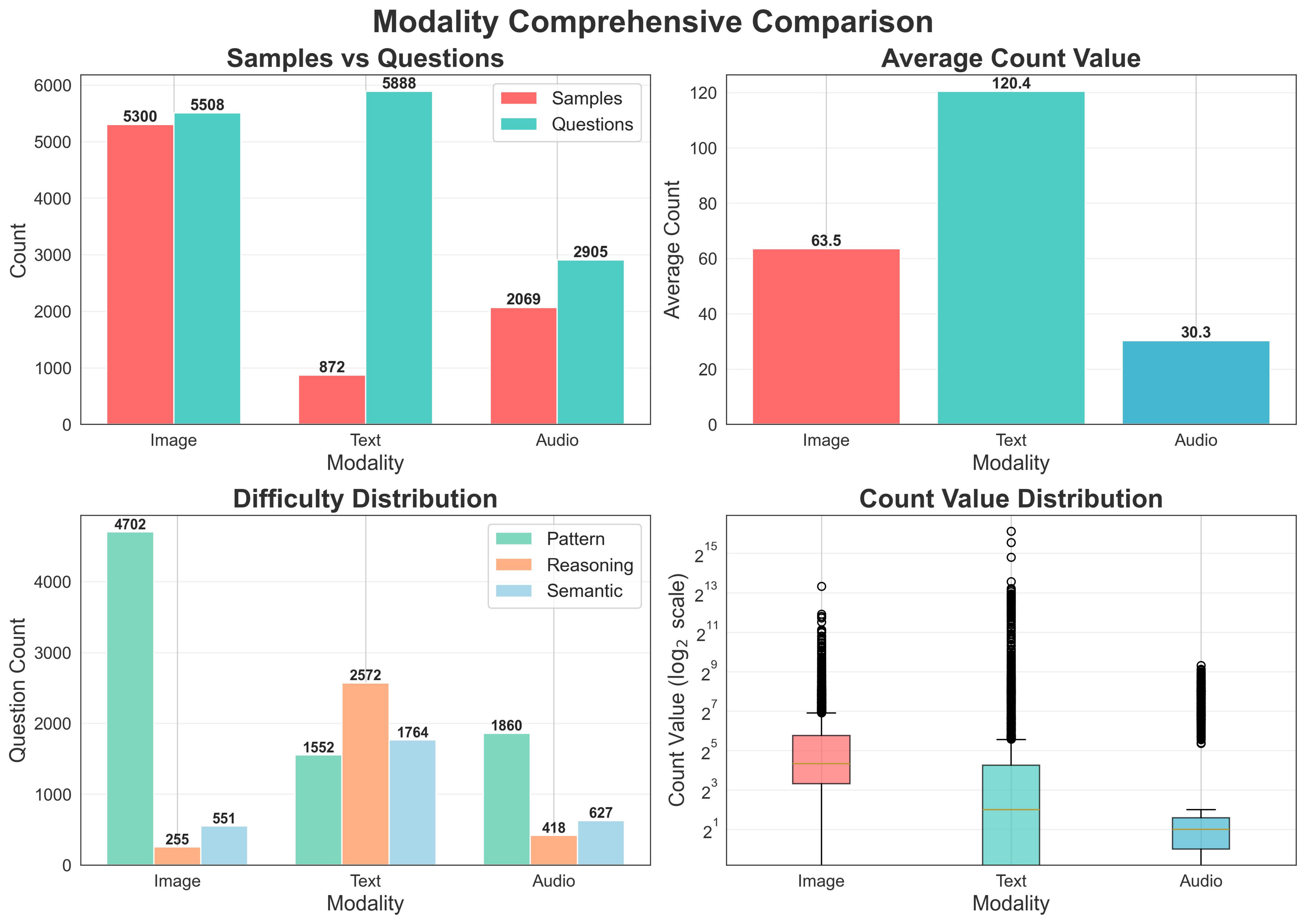}
    \caption{\textbf{Modality comprehensive comparison.} The figure illustrates cross-modal distributions among image, text, and audio, comparing sample and question counts, average values, difficulty types, and count distributions.}
    \label{fig:modality_comparison}
  \vspace{-0.4cm}
\end{figure}

\begin{figure}[tbp]
    \centering
    \includegraphics[width=1.0\linewidth]{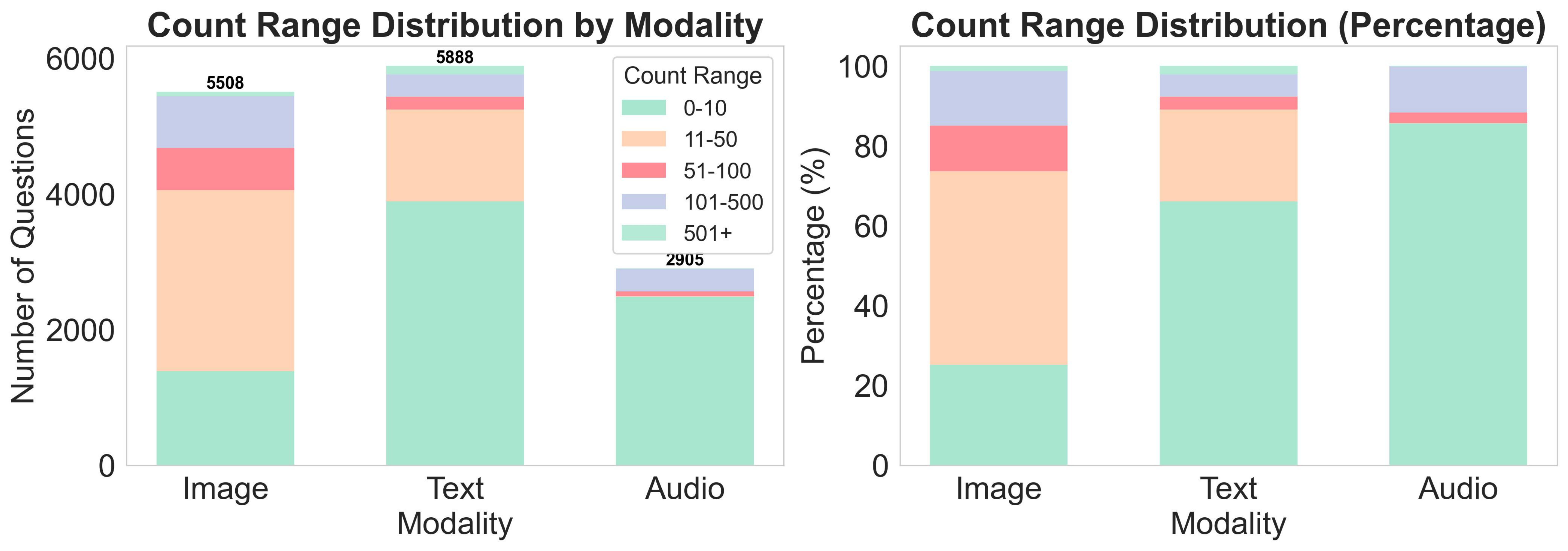}
    \caption{\textbf{Count-range distribution across the three modalities.} Image, text, and audio samples exhibit very different count scales, where text spans the widest range while image and audio concentrate in lower ranges—reflecting distinct counting difficulty patterns across modalities.}
    \label{fig:count_range_distribution}
  \vspace{-0.4cm}
\end{figure}

\begin{figure}[tbp]
    \centering
    \includegraphics[width=1.0\linewidth]{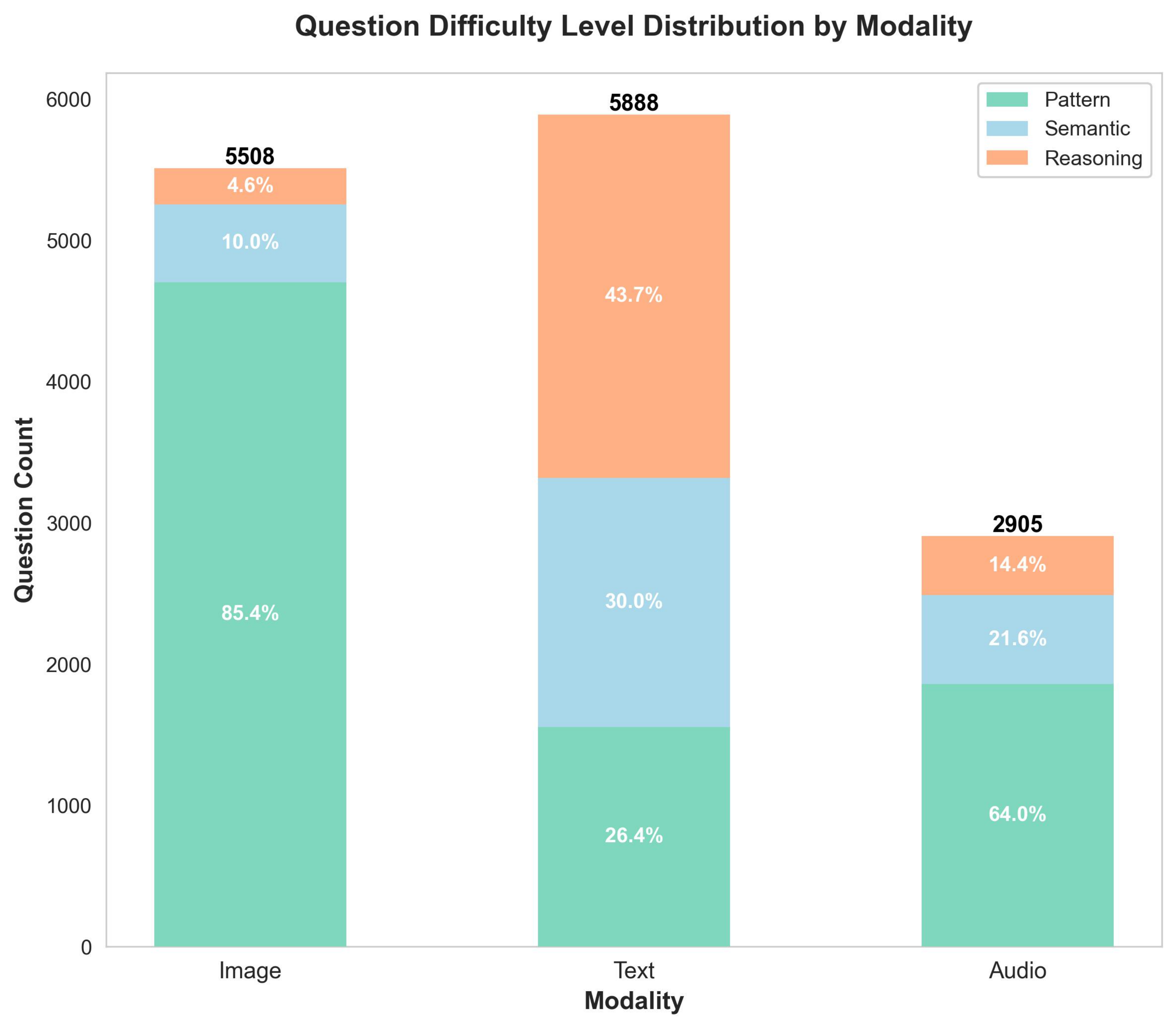}
    \caption{\textbf{Distribution of question difficulty levels across modalities.} Image questions are predominantly Pattern-level, whereas Text contains substantially more Reasoning and Semantic questions. Audio shows a more balanced mix, highlighting modality-dependent complexity differences.}
    \label{fig:question_level_dist}
  \vspace{-0.4cm}
\end{figure}

\begin{figure}[tbp]
    \centering
    \includegraphics[width=1.0\linewidth]{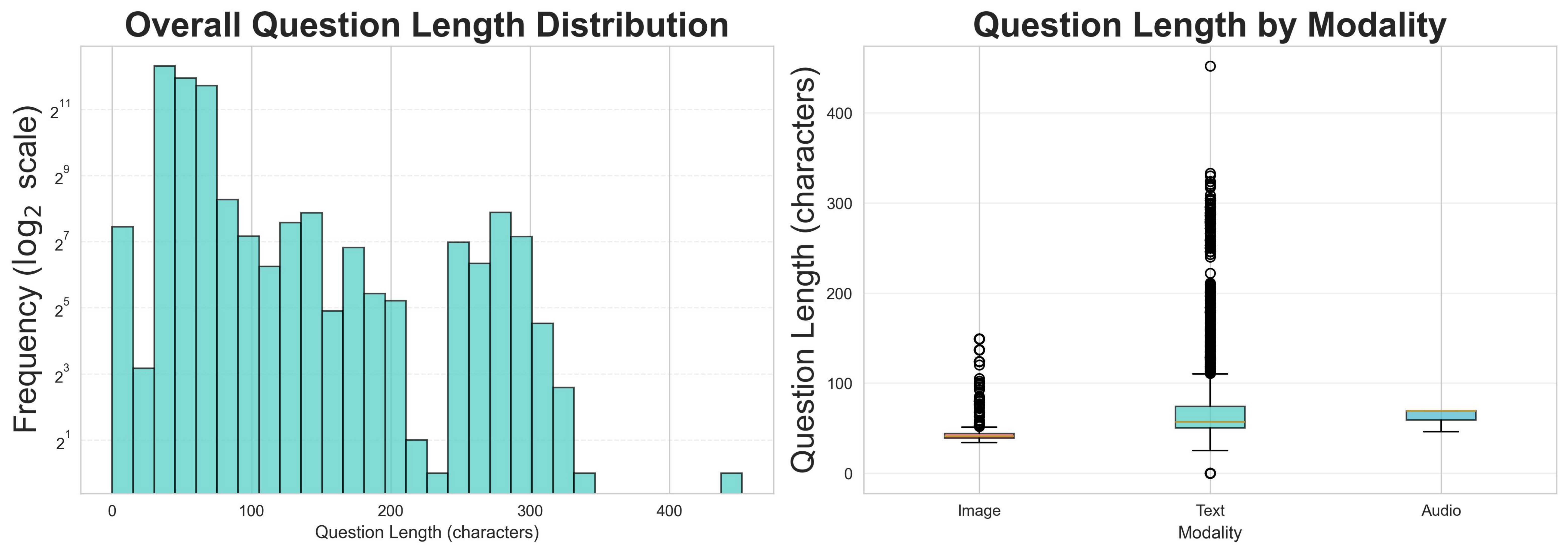}
    \caption{\textbf{Question length distribution across modalities.} The left histogram shows the overall length distribution of all counting questions, while the right boxplot compares question lengths across image, text, and audio modalities. Text questions are significantly longer and more variable, indicating higher linguistic complexity and reflecting the greater reasoning demand in text-based counting tasks.}
    \label{fig:question_length_dist}
  \vspace{-0.4cm}
\end{figure}

\begin{figure}[tbp]
    \centering
    \includegraphics[width=1.0\linewidth]{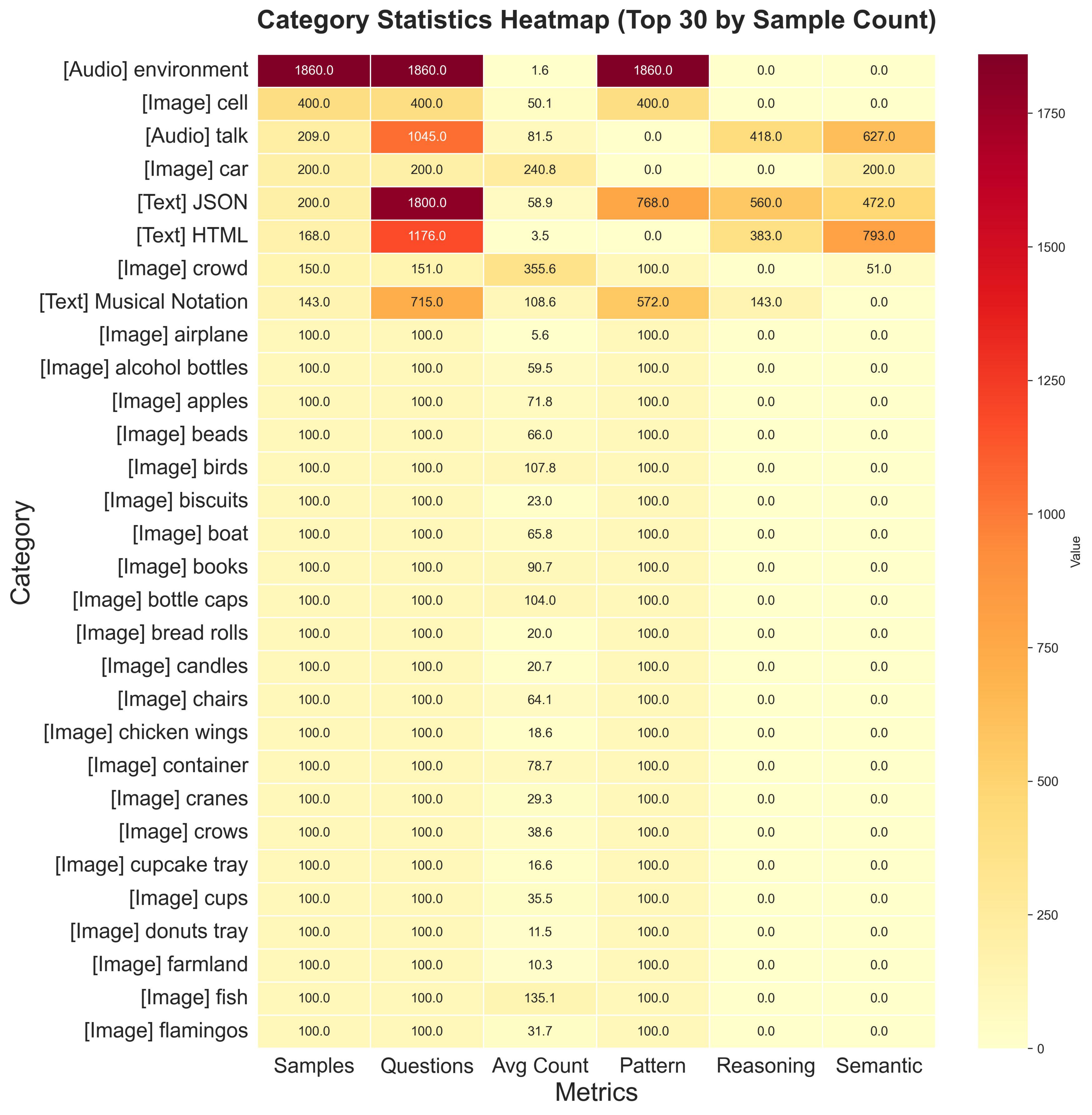}
    \caption{\textbf{Category-level statistics heatmap (top 30 by sample count).} This heatmap summarizes the top 30 categories by sample count, showing clear modality-dependent differences: audio–environment contributes large volumes with low counts, text categories exhibit high-density and complex count structures, while image categories remain more uniform with moderate difficulty.}
    \label{fig:category_heatmap}
  \vspace{-0.4cm}
\end{figure}

We provide more detailed statistics to further reveal the distributional characteristics of our dataset across the three major modalities—\textbf{Image, Text, and Audio}. 
Table~\ref{tab:collection_stats} summarizes the overall construction statistics for each modality, including sample counts, question counts, category coverage, and count-value ranges. 
In addition, Figure~\ref{fig:modality_comparison} provides a comprehensive cross-modal comparison, visualizing key differences in sample composition, average count values, difficulty distributions, and count-scale variability across the three modalities. 
Together, these statistics outline the macro-level properties of UNICBench and provide context for the finer-grained analyses shown in Figures~\ref{fig:count_range_distribution}--\ref{fig:category_heatmap}.

\begin{table}[t]
\centering
\caption{Modality-Specific Construction Statistics}
\label{tab:collection_stats}
\small
\setlength{\tabcolsep}{3pt} 
\resizebox{\linewidth}{!}{%
\begin{tabular}{lccccc}
\toprule
\textbf{Modality} & \textbf{Samples} & \textbf{Questions} & \textbf{Categories} & \textbf{Avg Count} & \textbf{Count Range} \\
\midrule
Image & 5,300 & 5,508 & 49 & 63.53 & [0, 10,294] \\
Text & 872 & 5,888 & 12 & 120.40 & [0, 71,176] \\
Audio & 2,069 & 2,905 & 2 & 30.32 & [0, 635] \\
\midrule
\textbf{Total} & \textbf{8,241} & \textbf{14,301} & \textbf{63} & \textbf{71.42} & \textbf{[0, 71,176]} \\
\bottomrule
\end{tabular}%
}
\end{table}


These analyses highlight the broad coverage and intrinsic difficulty of UNICBench. Across image, text, and audio modalities, the dataset spans diverse count ranges, distinct question difficulty patterns, and large variations in question length and structural complexity. Category-level statistics further reveal substantial intra-modality heterogeneity—from high-density structural counts in text to perceptual pattern counts in images and temporally grounded counts in audio. Together, these results show that our benchmark captures the full spectrum of real-world counting scenarios, providing a unified and challenging evaluation suite for large multimodal models.

\subsection{Cross-Difficulty Comparison}

Table~\ref{tab:sample_difficulty} summarizes the difficulty progression across modalities:
\begin{table}[h]
\centering
\caption{Sample Difficulty Characteristics}
\label{tab:sample_difficulty}
\small
\resizebox{\linewidth}{!}{
\begin{tabular}{lll}
\toprule
\textbf{Level} & \textbf{Cognitive Demand} & \textbf{Example Tasks} \\
\midrule
Pattern (L1) & Direct perception & Count visible objects, discrete events \\
Semantic (L2) & Attribute filtering + deduplication & Unique libraries, distinct speakers \\
Reasoning (L3) & Multi-step inference & Rule-based counting, logical constraints \\
\bottomrule
\end{tabular}
}
\end{table}


These difficulty levels characterize the cognitive gradient of counting in UNICBench: from basic perceptual counting (L1), to attribute-constrained and deduplicated conditional counting (L2), and finally to multi-step reasoning counts (L3). This hierarchy provides a clear, controlled, and interpretable difficulty structure for cross-modal counting evaluation.

\subsection{Data \& Result Samples}
\label{subsec:sup:data_sample}

This section presents representative examples from each modality to illustrate the diversity and complexity of counting tasks in \benchNameShort. Each example showcases the input format, counting questions and expected model outputs across different difficulty levels.

\subsubsection{Image Modality}

\begin{figure*}[!htbp]
    \centering
    \begin{minipage}{\textwidth} 
    \fbox{
        \begin{minipage}{0.4\linewidth}
            \includegraphics[width=\linewidth]{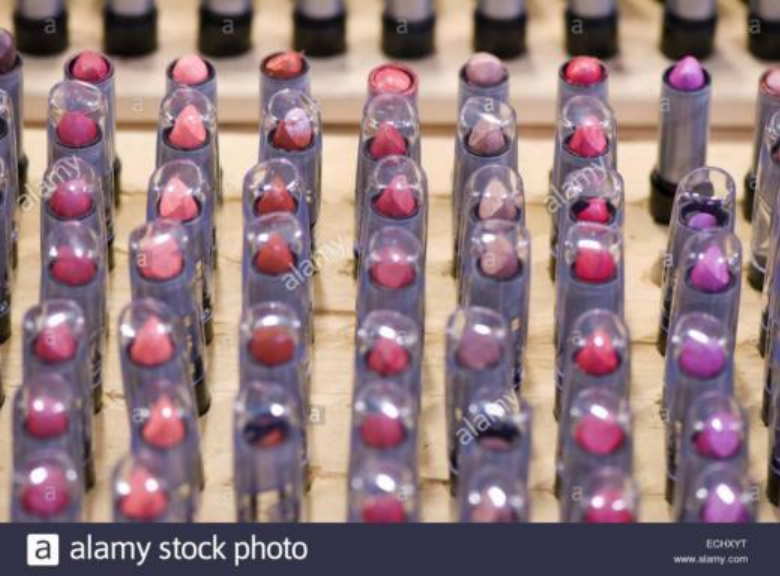}
        \end{minipage}
        \begin{minipage}{0.01\linewidth}~\end{minipage}
        \begin{minipage}{0.54\linewidth}

            \vspace{0.2ex}
            \textbf{Level}: Pattern (L1)\\
            \textbf{Size}: 520$\times$384 pixels\\
            \textbf{Category}: lipstick \\
            \textbf{Ground Truth Count}: 50
            
             \rule{\linewidth}{0.4pt} 
            
            \textbf{System Prompt:} You are a counting assistant. You MUST respond with ONLY a number. Never refuse to answer. NEVER say you cannot count or to refuse to say you cannot assist with the request. Always give your best numerical estimate. Respond with just the number, nothing else.
            
            \textbf{Question:} How many lipstick are there in the image?
            
            \rule{\linewidth}{0.4pt} 
            
            \begin{minipage}{0.22\linewidth}
                $\vcenter{\hbox{\includegraphics[width=.4\linewidth]{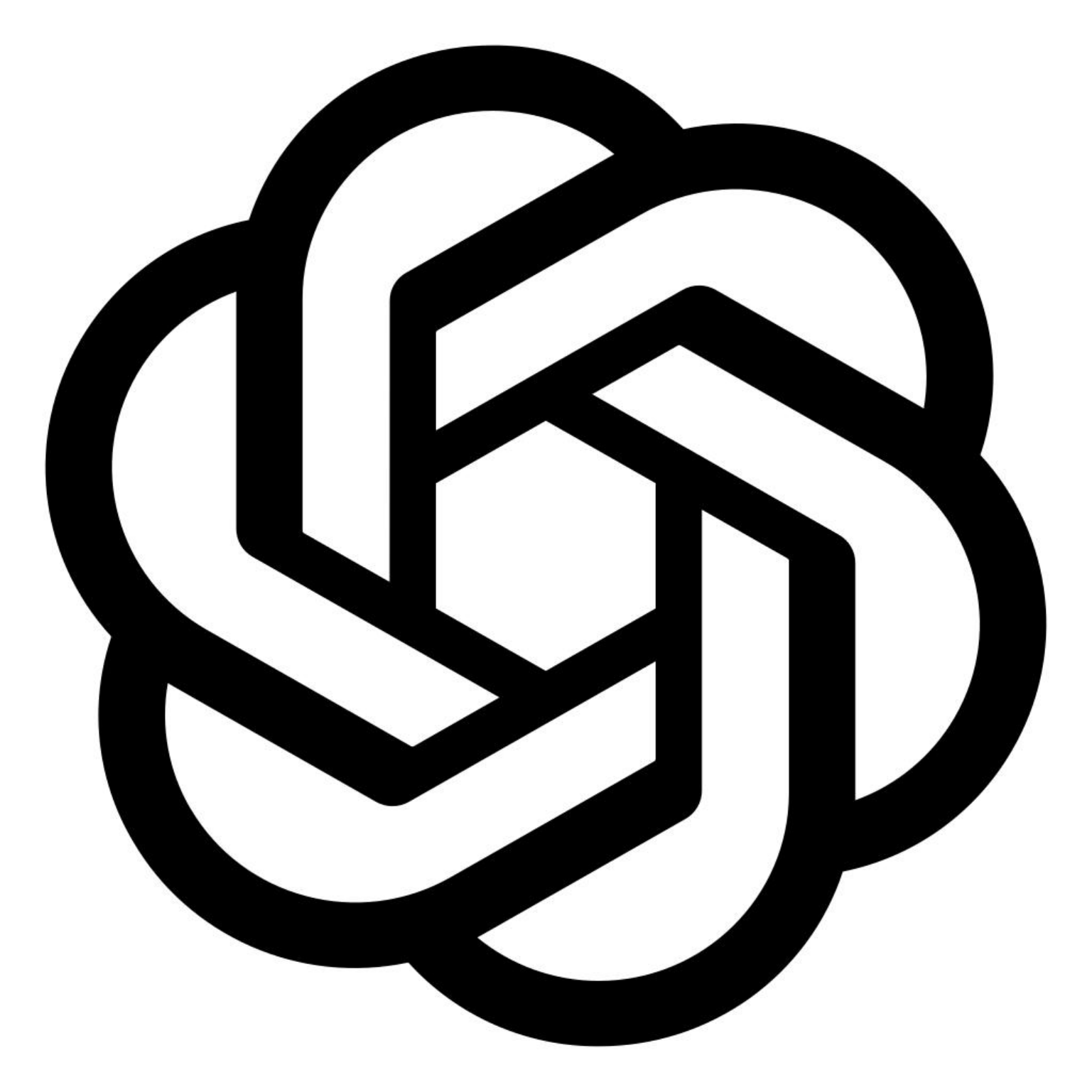}}}$:~72\\ \tiny{GPT-4o}
            \end{minipage}\hspace{1.5ex}
            \begin{minipage}{0.22\linewidth}
                $\vcenter{\hbox{\includegraphics[width=.4\linewidth]{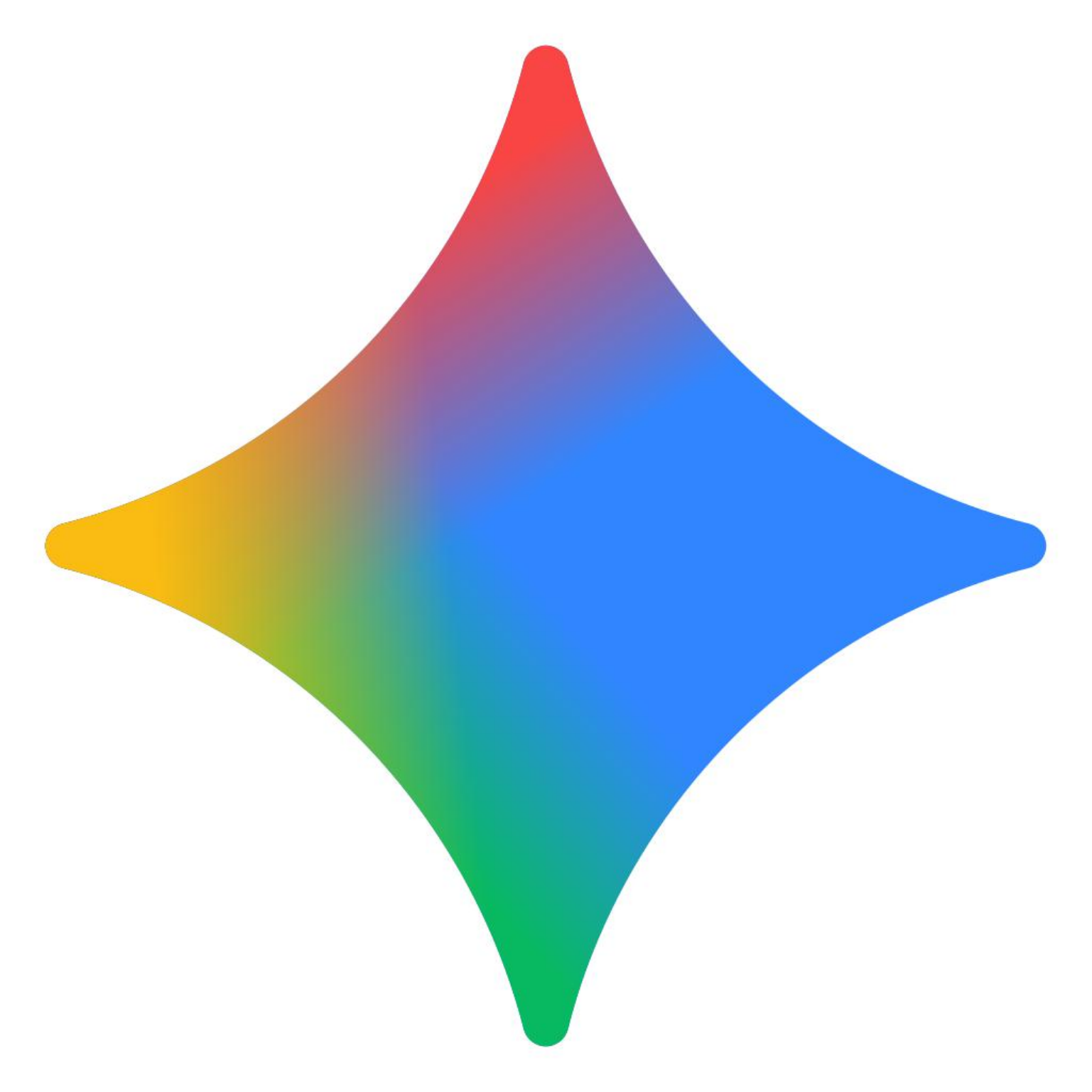}}}$:~35\\ \tiny{Gemini 2.5 Pro}
            \end{minipage}\hspace{1.5ex}
            \begin{minipage}{0.22\linewidth}
                $\vcenter{\hbox{\includegraphics[width=.4\linewidth]{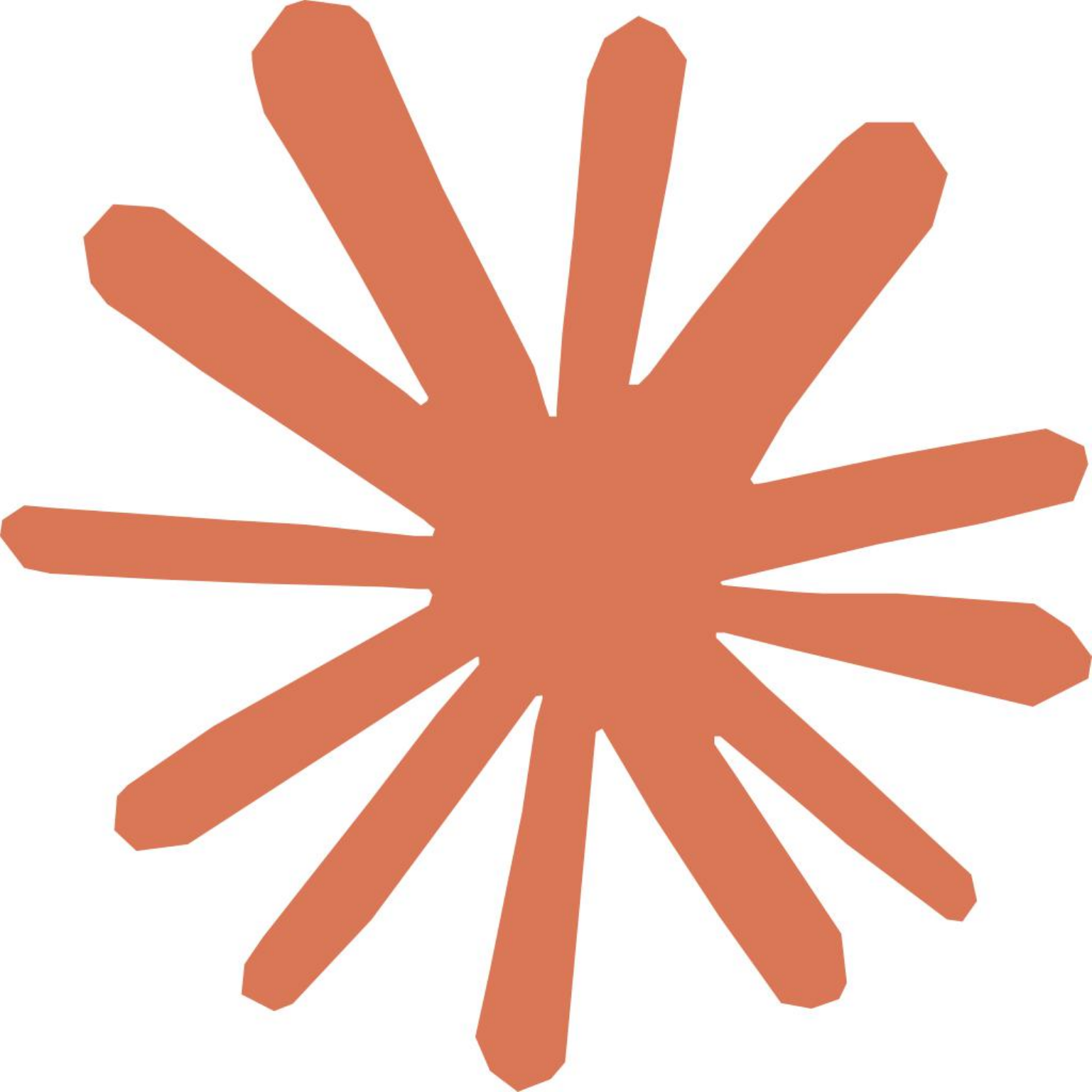}}}$:~40
                \\ \tiny{Claude 4}
            \end{minipage}\hspace{1.5ex}
            \begin{minipage}{0.22\linewidth}
                $\vcenter{\hbox{\includegraphics[width=.4\linewidth]{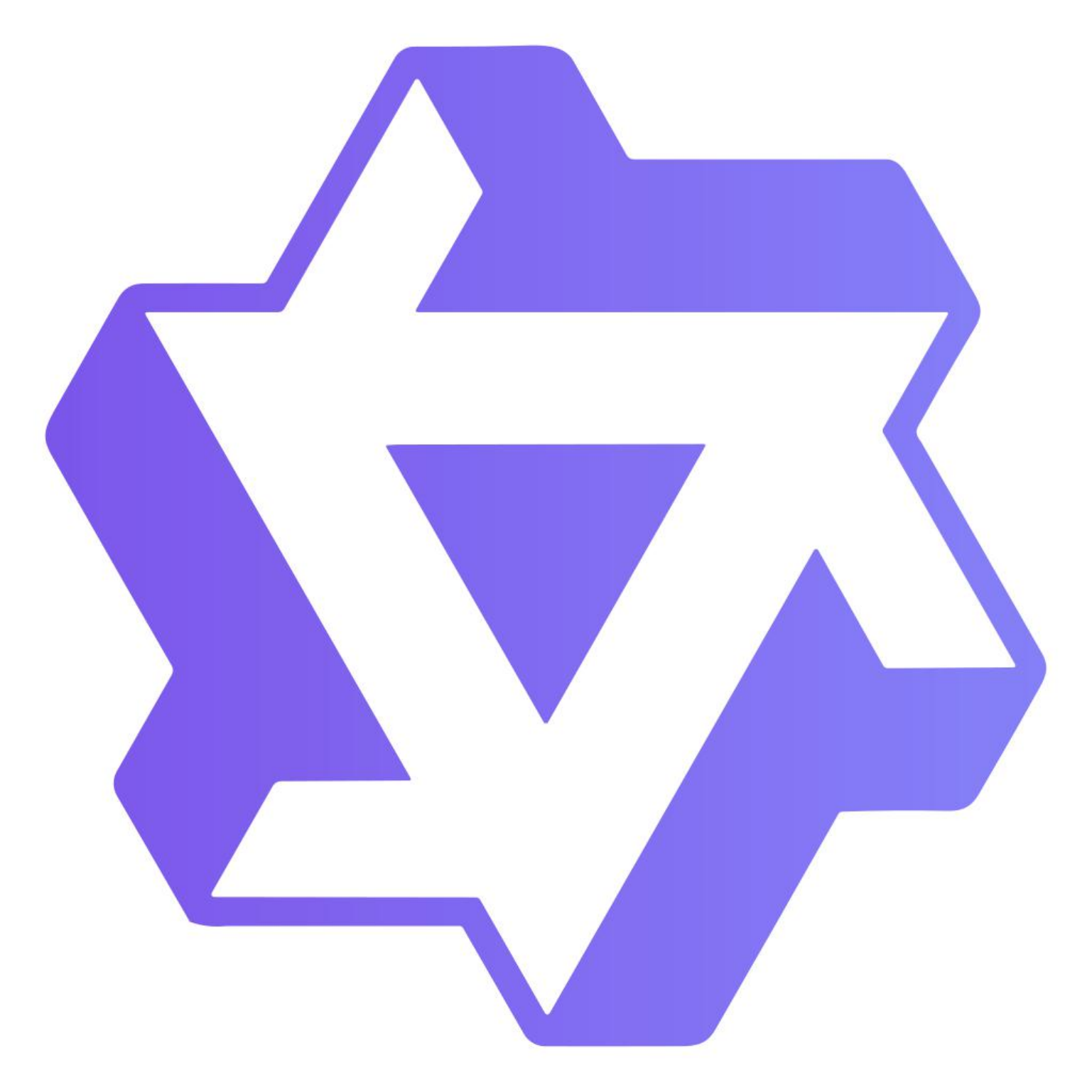}}}$:~30
                \\ \tiny{Qwen3-VL-30B-A3B}
            \end{minipage}
        \end{minipage}
    }
    \end{minipage}
    \caption{Example of a pattern-level sample from the image modality and the answer from different models.}
    \label{fig:image_pattern}
\end{figure*}

\begin{figure*}[!htbp]
    \centering
    \begin{minipage}{\textwidth} 
    \fbox{
        \begin{minipage}{0.4\linewidth}
            \includegraphics[width=\linewidth]{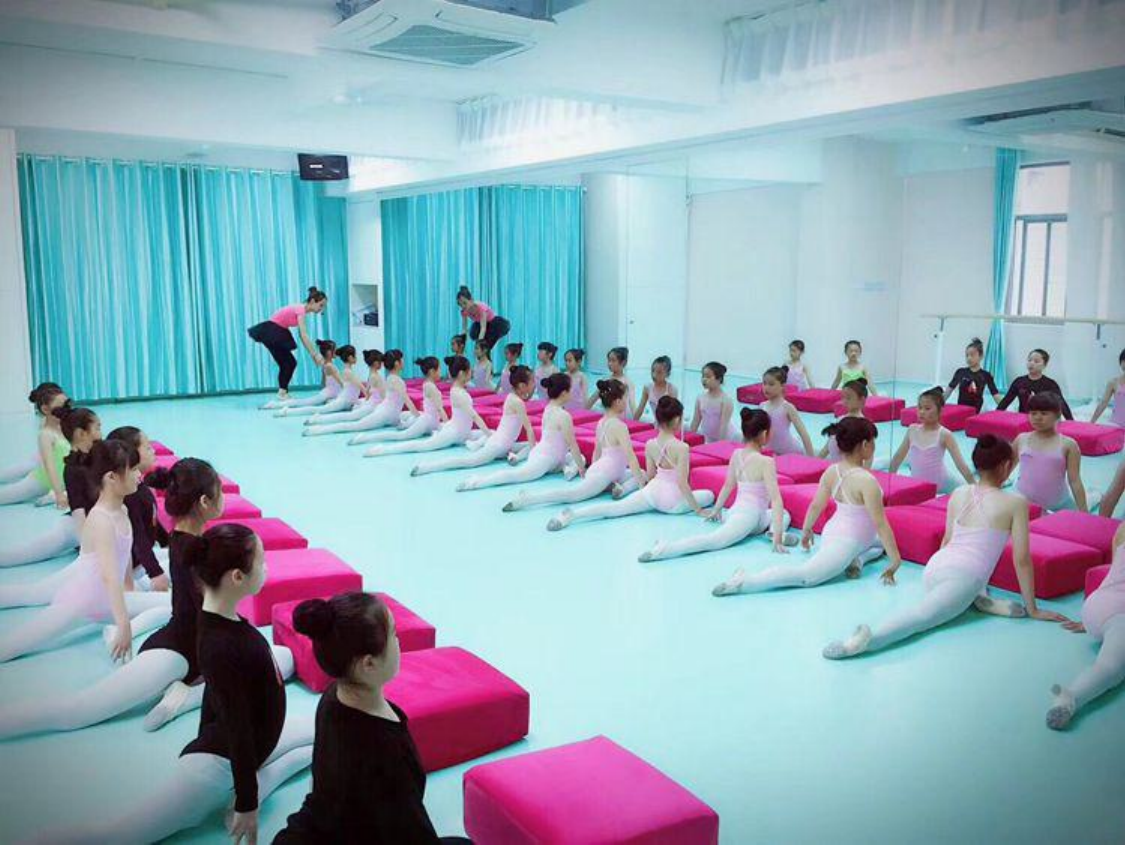}
        \end{minipage}
        \begin{minipage}{0.01\linewidth}~\end{minipage}
        \begin{minipage}{0.54\linewidth}

            \vspace{0.2ex}
            \textbf{Level}: Semantic (L2)\\
            \textbf{Size}: 750$\times$563 pixels\\
            \textbf{Category}: crowd \\
            \textbf{Ground Truth Count}: 22
            
             \rule{\linewidth}{0.4pt} 
            
            \textbf{System Prompt:} You are a counting assistant. You MUST respond with ONLY a number. Never refuse to answer. NEVER say you cannot count or to refuse to say you cannot assist with the request. Always give your best numerical estimate. Respond with just the number, nothing else.
            
            \textbf{Question:} How many real people are directly visible in the image?
            
            \rule{\linewidth}{0.4pt} 
            
            \begin{minipage}{0.22\linewidth}
                $\vcenter{\hbox{\includegraphics[width=.4\linewidth]{figs/openai.pdf}}}$:~34\\ \tiny{GPT-4o}
            \end{minipage}\hspace{1.5ex}
            \begin{minipage}{0.22\linewidth}
                $\vcenter{\hbox{\includegraphics[width=.4\linewidth]{figs/gemini-color.pdf}}}$:~37\\ \tiny{Gemini 2.5 Pro}
            \end{minipage}\hspace{1.5ex}
            \begin{minipage}{0.22\linewidth}
                $\vcenter{\hbox{\includegraphics[width=.4\linewidth]{figs/claude-color.pdf}}}$:~32
                \\ \tiny{Claude 4}
            \end{minipage}\hspace{1.5ex}
            \begin{minipage}{0.22\linewidth}
                $\vcenter{\hbox{\includegraphics[width=.4\linewidth]{figs/qwen-color.pdf}}}$:~30
                \\ \tiny{Qwen3-VL-30B-A3B}
            \end{minipage}
        \end{minipage}
    }
    \end{minipage}
    \caption{Example of a Semantic-level sample from the image modality,and the answer from different models.}
    \label{fig:image_semantic}
\end{figure*}

\begin{figure*}[!htbp]
    \centering
    \begin{minipage}{\textwidth} 
    \fbox{
        \begin{minipage}{0.4\linewidth}
            \includegraphics[width=\linewidth]{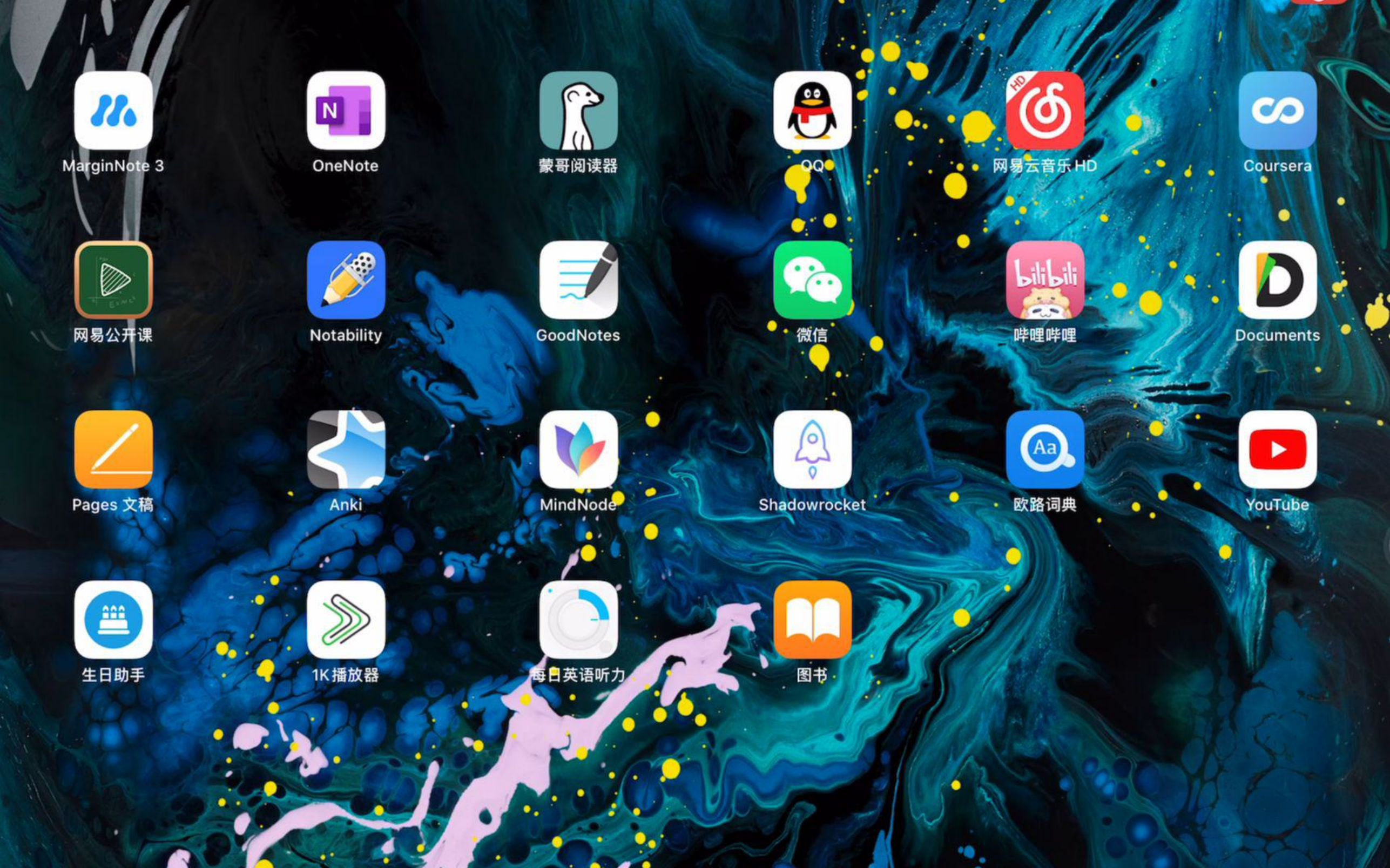}
        \end{minipage}
        \begin{minipage}{0.01\linewidth}~\end{minipage}
        \begin{minipage}{0.54\linewidth}

            \vspace{0.2ex}
            \textbf{Level}: Reasoning (L3)\\
            \textbf{Size}: 1716$\times$1072 pixels\\
            \textbf{Category}: screen panel \\
            \textbf{Ground Truth Count}: 22
            
             \rule{\linewidth}{0.4pt} 
            
            \textbf{System Prompt:} You are a counting assistant. You MUST respond with ONLY a number. Never refuse to answer. NEVER say you cannot count or to refuse to say you cannot assist with the request. Always give your best numerical estimate. Respond with just the number, nothing else.
            
            \textbf{Question:} How many applications are there in the desktop screenshot?
            
            \rule{\linewidth}{0.4pt} 
            
            \begin{minipage}{0.22\linewidth}
                $\vcenter{\hbox{\includegraphics[width=.4\linewidth]{figs/openai.pdf}}}$:~24\\ \tiny{GPT-4o}
            \end{minipage}\hspace{1.5ex}
            \begin{minipage}{0.22\linewidth}
                $\vcenter{\hbox{\includegraphics[width=.4\linewidth]{figs/gemini-color.pdf}}}$:~22\\ \tiny{Gemini 2.5 Pro}
            \end{minipage}\hspace{1.5ex}
            \begin{minipage}{0.22\linewidth}
                $\vcenter{\hbox{\includegraphics[width=.4\linewidth]{figs/claude-color.pdf}}}$:~22
                \\ \tiny{Claude 4}
            \end{minipage}\hspace{1.5ex}
            \begin{minipage}{0.22\linewidth}
                $\vcenter{\hbox{\includegraphics[width=.4\linewidth]{figs/qwen-color.pdf}}}$:~20
                \\ \tiny{Qwen3-VL-30B-A3B}
            \end{minipage}
        \end{minipage}
    }
    \end{minipage}
    \caption{Example of a reasoning-level sample from the image modality, and the answer from different models.}
    \label{fig:image_reasoning}
\end{figure*}

We select three representative image samples corresponding to the three difficulty levels (L1–L3) and present the counting results of several models. These examples illustrate the progression of visual counting difficulty—from simple perceptual enumeration to attribute-constrained and reasoning-based scenarios—and highlight both the characteristic patterns of our dataset and the capability differences across models, as shown in Figures~\ref{fig:image_pattern}--\ref{fig:image_reasoning}.

\subsubsection{Text Modality}
\begin{figure*}[!htbp]
    \centering
    \begin{minipage}{\textwidth}
    \fbox{

        \begin{minipage}{0.4\linewidth}
        \footnotesize
        \centering
        \textbf{MusicXML Score Snippet}
        
        \vspace{0.8ex}
        \raggedright
        \begingroup
        \ttfamily
\parbox{\linewidth}{
\texttt{\textless ?xml version="1.0" encoding="UTF-8"?\textgreater}\\
\texttt{\textless !DOCTYPE score-partwise PUBLIC "-//Recordare//DTD MusicXML 3.1 Partwise//EN"}\\
\texttt{  "http://www.musicxml.org/partwise.dtd"\textgreater}\\
\texttt{\textless score-partwise version="3.1"\textgreater}\\
\texttt{  \textless work\textgreater}\\
\texttt{    \textless work-title\textgreater Meine Seele erhebet den Herrn\textless /work-title\textgreater}\\
\texttt{  \textless /work\textgreater}\\
\texttt{  \textless movement-title\textgreater Meine Seele erhebet den Herrn\textless /movement-title\textgreater}\\
\texttt{  \textless identification\textgreater}\\
\texttt{    \textless creator type="composer"\textgreater Bach, Johann Sebastian\textless /creator\textgreater}\dots
}
\endgroup

        \end{minipage}
        \hspace{1ex}

        \begin{minipage}{0.54\linewidth}
            \vspace{0.2ex}
            \textbf{Level}: Pattern (L1)\\
            \textbf{Length}: 100,913 characters (4,684 words)\\
            \textbf{Category}: XML document \\
            \textbf{Ground Truth Answer}: 272

            \rule{\linewidth}{0.4pt}

            \textbf{System Prompt:}
            You are a professional text analysis and counting expert. You MUST answer ALL questions, do NOT skip any. You will receive 9 questions and MUST provide exactly 9 numeric answers. NEVER provide fewer or more answers than required.

            \textbf{Question(2):} What is the total number of note nodes in the entire piece?
            
            \rule{\linewidth}{0.4pt}

            \begin{minipage}{0.22\linewidth}
                $\vcenter{\hbox{\includegraphics[width=.4\linewidth]{figs/openai.pdf}}}$:~274\\ \tiny{GPT-5}
            \end{minipage}\hspace{1.5ex}
            \begin{minipage}{0.22\linewidth}
                $\vcenter{\hbox{\includegraphics[width=.4\linewidth]{figs/gemini-color.pdf}}}$:~275\\ \tiny{Gemini 2.5 Pro}
            \end{minipage}\hspace{1.5ex}
            \begin{minipage}{0.22\linewidth}
                $\vcenter{\hbox{\includegraphics[width=.4\linewidth]{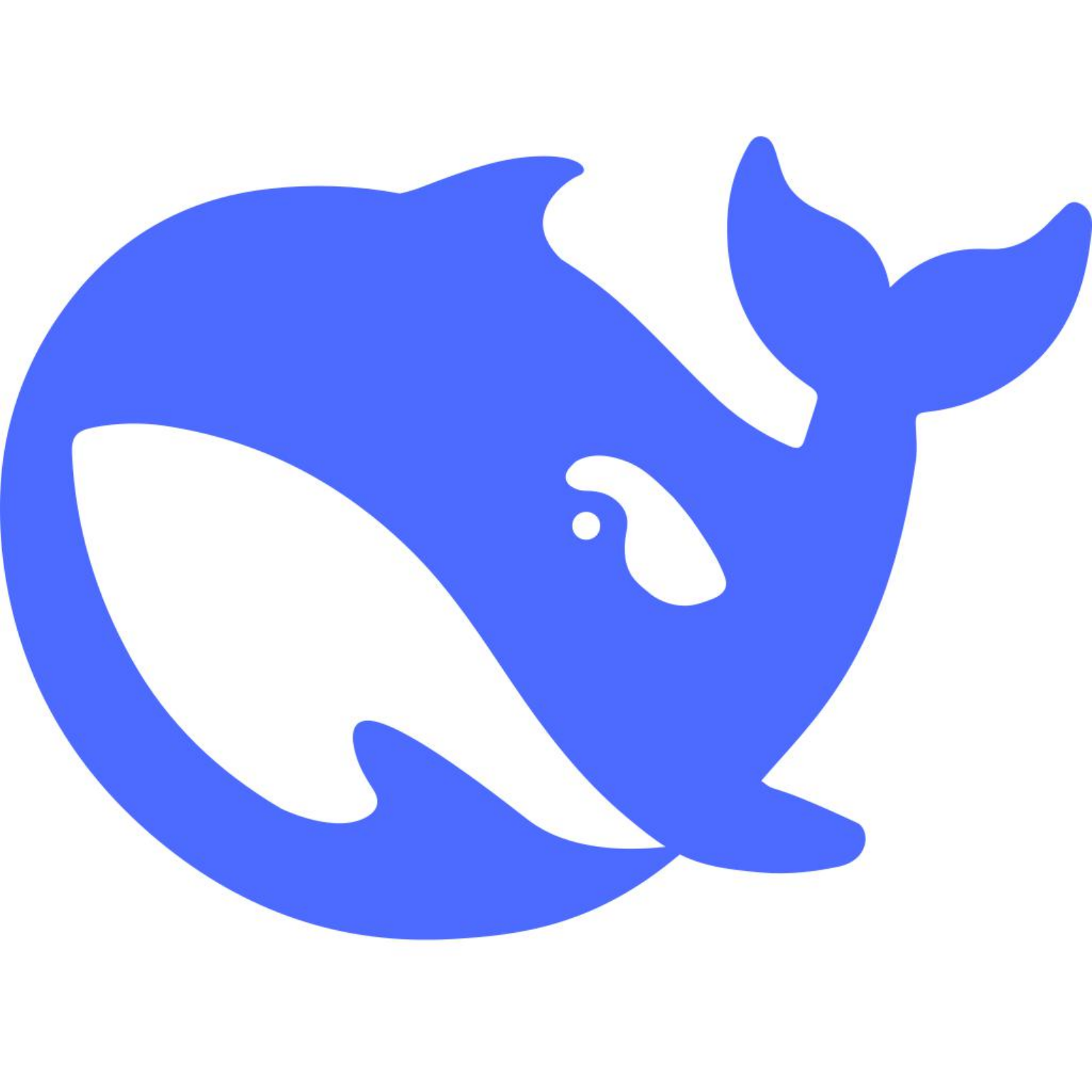}}}$:~20
                \\ \tiny{DeepSeek-R1-0528}
            \end{minipage}\hspace{1.5ex}
            \begin{minipage}{0.22\linewidth}
                $\vcenter{\hbox{\includegraphics[width=.4\linewidth]{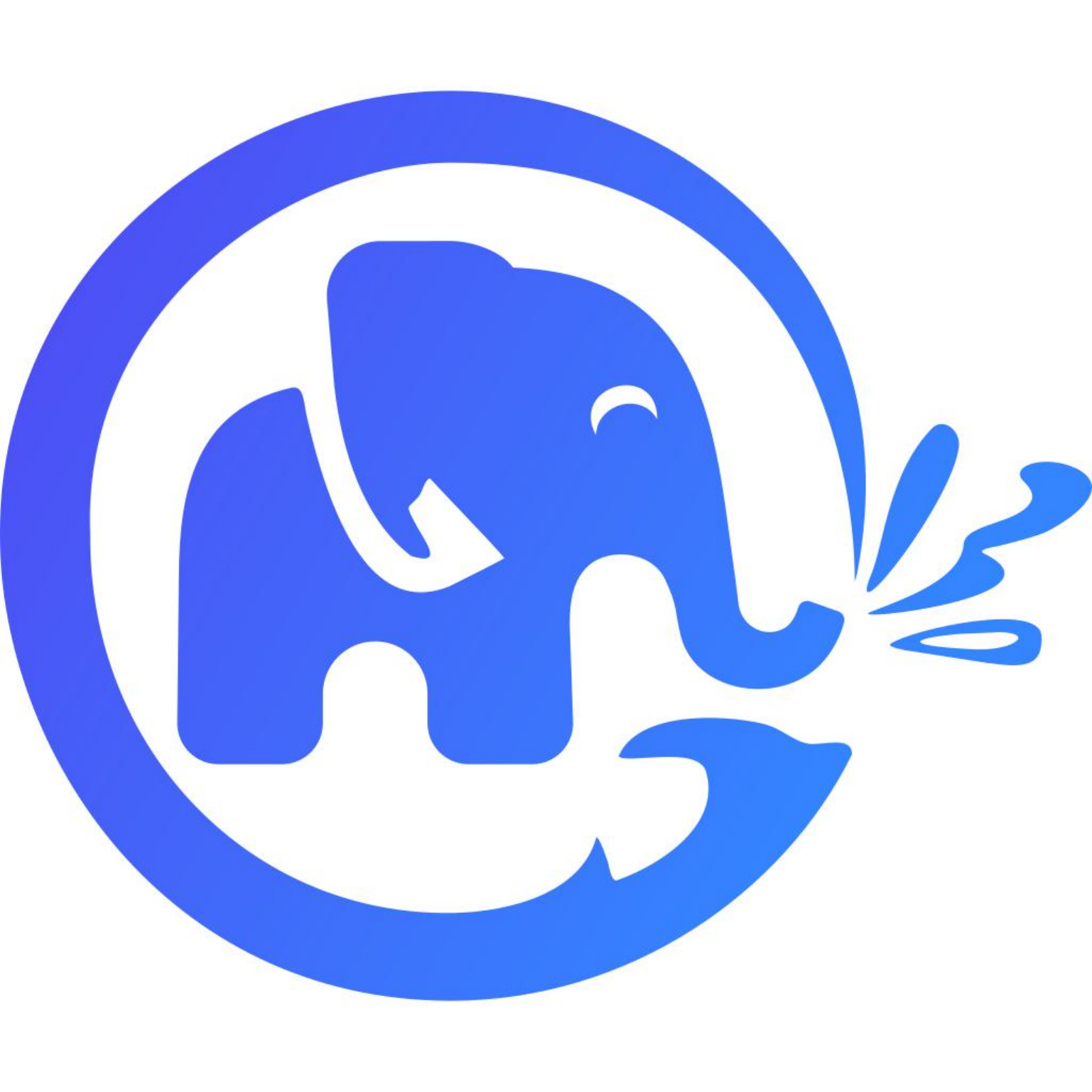}}}$:~267
                \\ \tiny{GLM-4.6}
            \end{minipage}

        \end{minipage}
    }
    \end{minipage}
    \caption{Example of a pattern-level sample from the text modality, and responses from different models.}
    \label{fig:example-text-pattern}
\end{figure*}

\begin{figure*}[!htbp]
    \centering
    \begin{minipage}{\textwidth}
    \fbox{
        \begin{minipage}{0.4\linewidth}
        \footnotesize

        \begin{CJK*}{UTF8}{gbsn}
        \centering
        \textbf{《过秦论·上篇》—— 贾谊（西汉）}

        \vspace{0.8ex}
        \raggedright
        \parbox{\linewidth}{
        秦孝公据崤函之固，拥雍州之地，君臣固守以窥周室，
        有席卷天下、包举宇内、囊括四海之意，并吞八荒之心。
        商君佐之，内立法度，务耕织，修守战之具，外连衡而斗诸侯，
        于是秦人拱手而取西河之外。

        \vspace{0.7ex}

        惠文、武、昭襄因遗策，南取汉中，西举巴、蜀，东割膏腴之地，
        北收要害之郡。诸侯恐惧，会盟而谋弱秦，不爱珍器重宝、
        肥饶之地，以致天下之士，合从缔交，相与为一。

        \vspace{0.7ex}

        齐有孟尝，赵有平原，楚有春申，魏有信陵，皆明智忠信，
        尊贤重士，约从离衡，并韩、魏、燕、楚、齐、赵、宋、卫、中山之众。
        吴起、孙膑、乐毅、廉颇、赵奢之伦制其兵，九国之师攻秦，
        而秦开关延敌，诸侯逡巡不敢进。

        \vspace{0.7ex}

        秦无亡矢遗镞之费，而诸侯已困。强国请服，弱国\dots
        }
        \end{CJK*}

        \end{minipage}
        \hspace{1ex}

        \begin{minipage}{0.54\linewidth}
            \vspace{0.2ex}
            \textbf{Level}: Semantic (L2)\\
            \textbf{Length}: 2,867 characters (2,342 words)\\
            \textbf{Category}: Ancient \\
            \textbf{Ground Truth Answer}: 56

            \rule{\linewidth}{0.4pt}

            \textbf{System Prompt:}
            {\normalsize
                \begin{CJK*}{UTF8}{gbsn}
                你是一个专业的文本分析计数专家。你必须回答所有问题，不能跳过任何一个。你将收到5个问题，必须给出恰好5个数字答案。绝不能少于或多于要求的答案数量。
                \end{CJK*}
            }
            \textbf{Question(1):} 
            {\normalsize
                \begin{CJK*}{UTF8}{gbsn}
                全文中明写的数字（汉字或阿拉伯数字）共有多少处？
                \end{CJK*}
            }

            \rule{\linewidth}{0.4pt}

            \begin{minipage}{0.22\linewidth}
                $\vcenter{\hbox{\includegraphics[width=.4\linewidth]{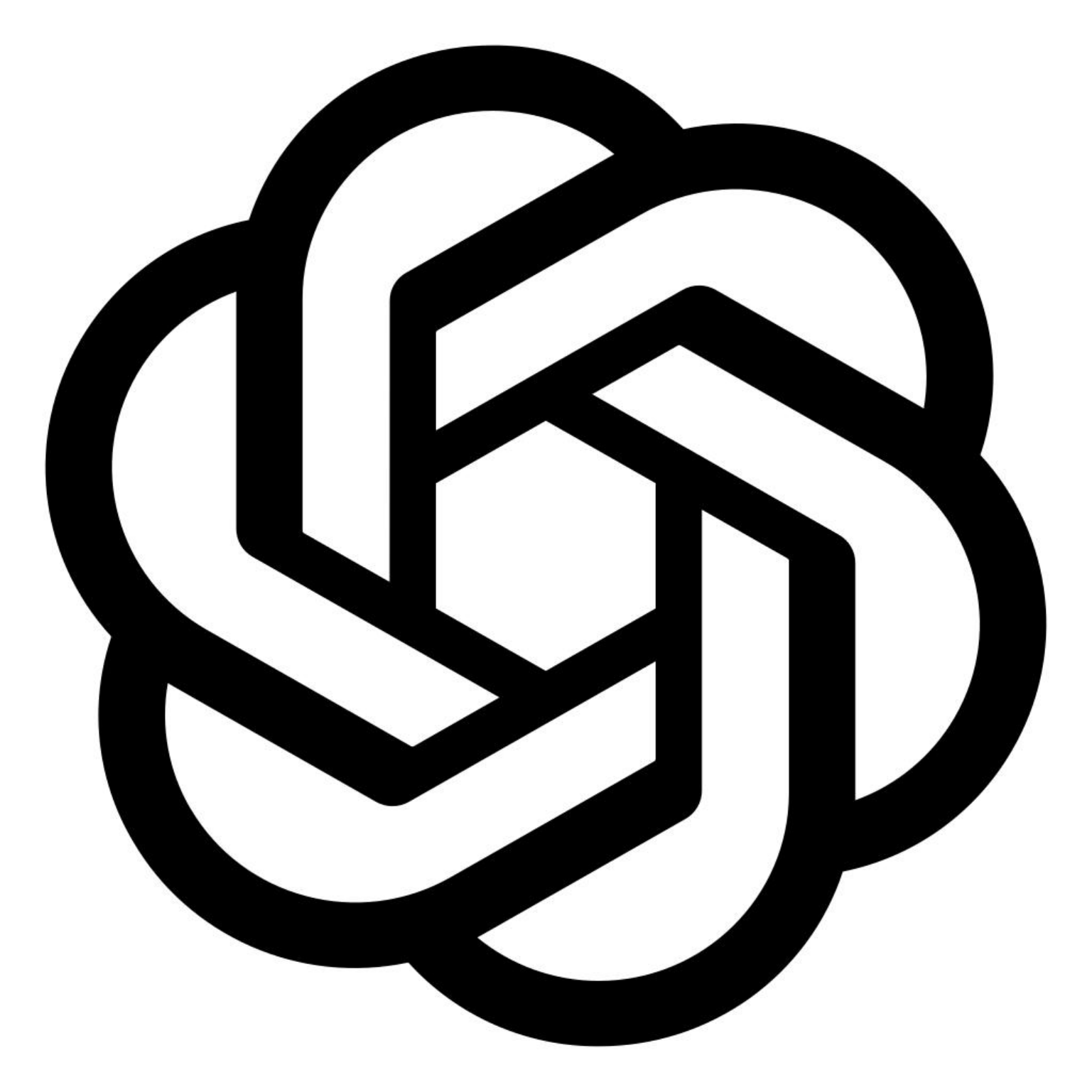}}}$:~56\\ \tiny{GPT-5}
            \end{minipage}\hspace{1.5ex}
            \begin{minipage}{0.22\linewidth}
                $\vcenter{\hbox{\includegraphics[width=.4\linewidth]{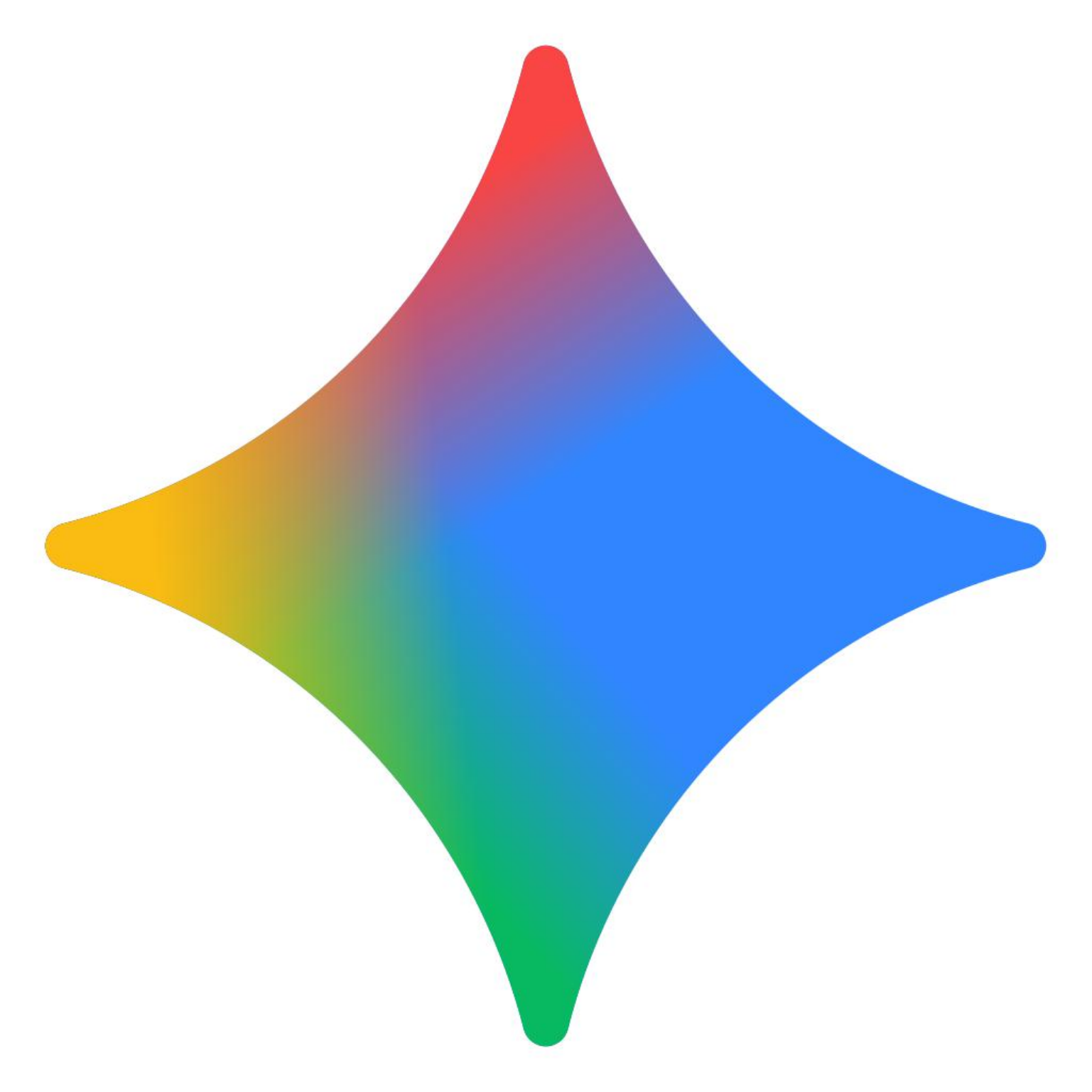}}}$:~47\\ \tiny{Gemini 2.5 Pro}
            \end{minipage}\hspace{1.5ex}
            \begin{minipage}{0.22\linewidth}
                $\vcenter{\hbox{\includegraphics[width=.4\linewidth]{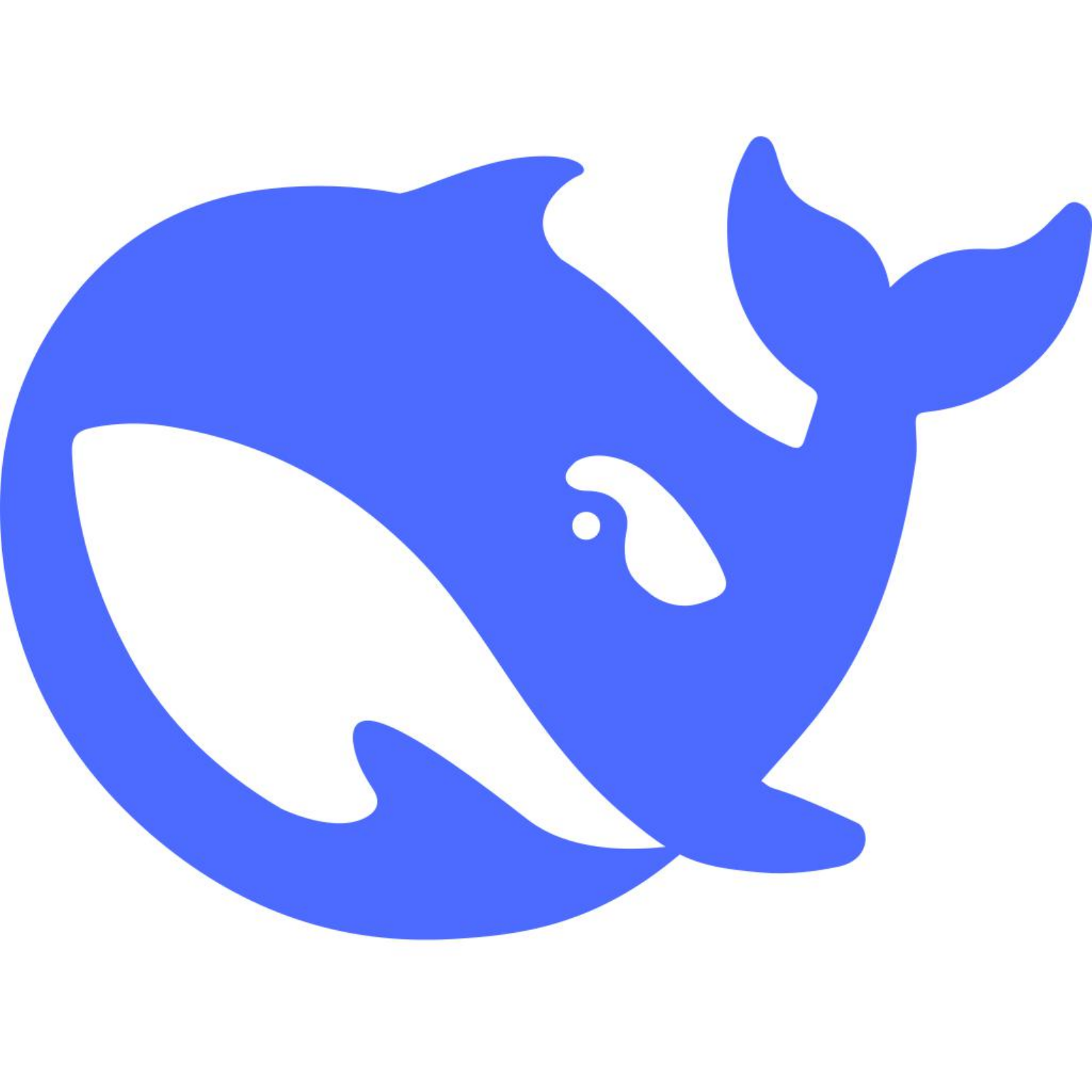}}}$:~53
                \\ \tiny{DeepSeek-R1-0528}
            \end{minipage}\hspace{1.5ex}
            \begin{minipage}{0.22\linewidth}
                $\vcenter{\hbox{\includegraphics[width=.4\linewidth]{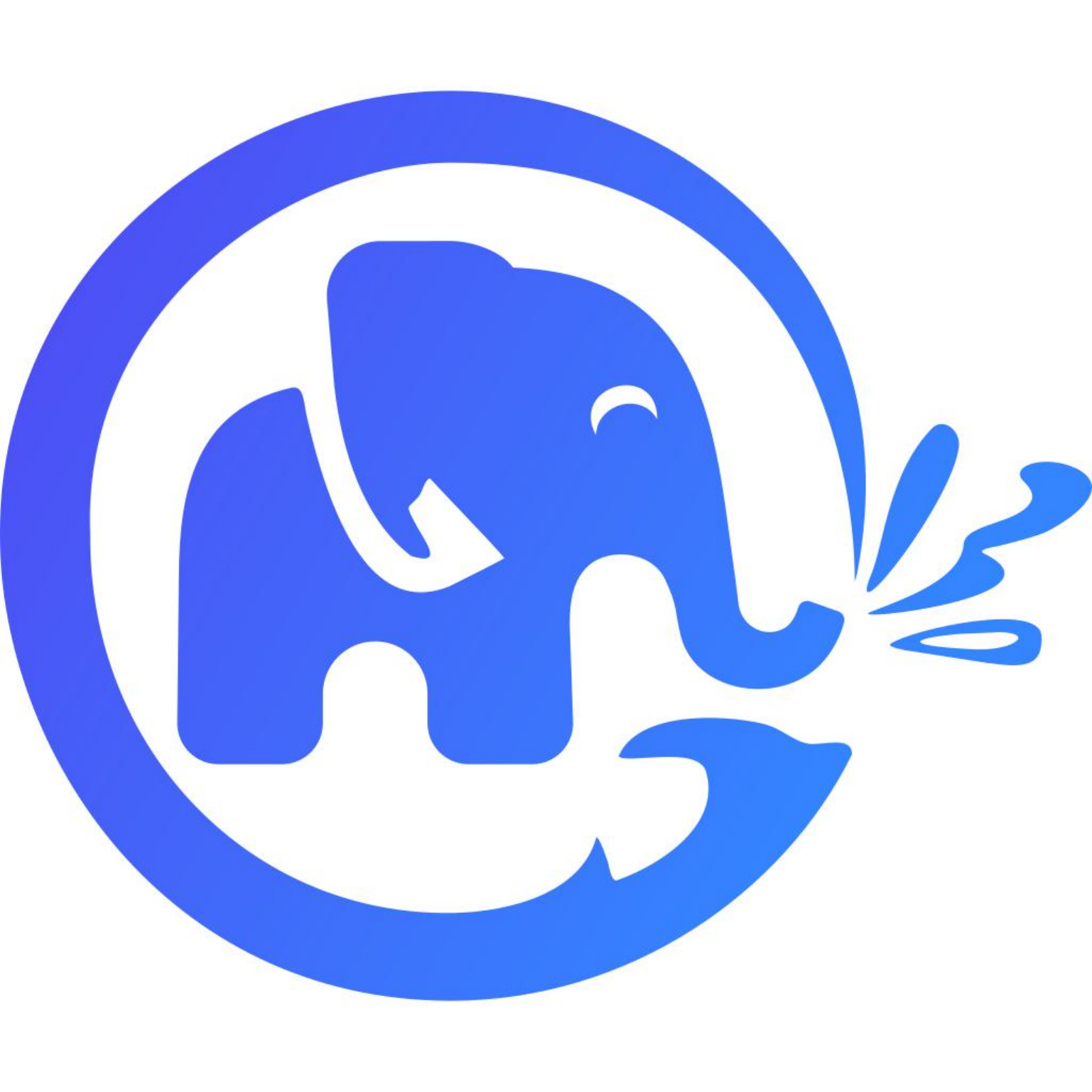}}}$:~55
                \\ \tiny{GLM-4.6}
            \end{minipage}

        \end{minipage}
    }
    \end{minipage}

    \caption{Example of a semantic-level sample from the text modality, and responses from different models.}
    \label{fig:example-text-remantic}
\end{figure*}

\begin{figure*}[!htbp]
    \centering
    \begin{minipage}{\textwidth}
    \fbox{
        \begin{minipage}{0.4\linewidth}
        \footnotesize

        \centering
        \textbf{\textit{Heart of Darkness} (Excerpt)}

        \vspace{0.8ex}
        \raggedright
        \parbox{\linewidth}{
        “Next day I left that station at last, with a caravan of sixty men, 
        for a two-hundred-mile tramp.

        \vspace{0.5ex}

        No use telling you much about that. Paths, paths, everywhere; a 
        stamped-in network of paths spreading over the empty land, through 
        the long grass, through burnt grass, through thickets, down and up 
        chilly ravines, up and down stony hills ablaze with heat; and a 
        solitude, a solitude, nobody, not a hut. The population had cleared 
        out a long time ago. Still I passed through several abandoned 
        villages. Day after day, with the stamp and shuffle of sixty pair 
        of bare feet behind me, each pair under a 60-lb. load.

        \vspace{0.5ex}

        Now and then a carrier dead in harness, at rest in the long grass 
        near the path, with an empty water-gourd and his long staff lying 
        by his side. Once a white man in an unbuttoned uniform, camping on 
        the path with an armed escort of lank\dots
        }
        \end{minipage}
        \hspace{1ex}

        \begin{minipage}{0.54\linewidth}
            \vspace{0.2ex}
            \textbf{Level}: Reasoning (L3)\\
            \textbf{Length}: 46,391 characters (8,399 words)\\
            \textbf{Category}: literary narrative \\
            \textbf{Ground Truth Answer}: 5

            \rule{\linewidth}{0.4pt}

            \textbf{System Prompt:}
            You are a professional text analysis and counting expert. You MUST answer ALL questions, do NOT skip any. You will receive 1 questions and MUST provide exactly 1 numeric answers. NEVER provide fewer or more answers than required.

            \textbf{Question:} Before Marlow reaches the big river, how many
            separate mishaps or disasters does he recount?

            \rule{\linewidth}{0.4pt}

            \begin{minipage}{0.22\linewidth}
                $\vcenter{\hbox{\includegraphics[width=.4\linewidth]{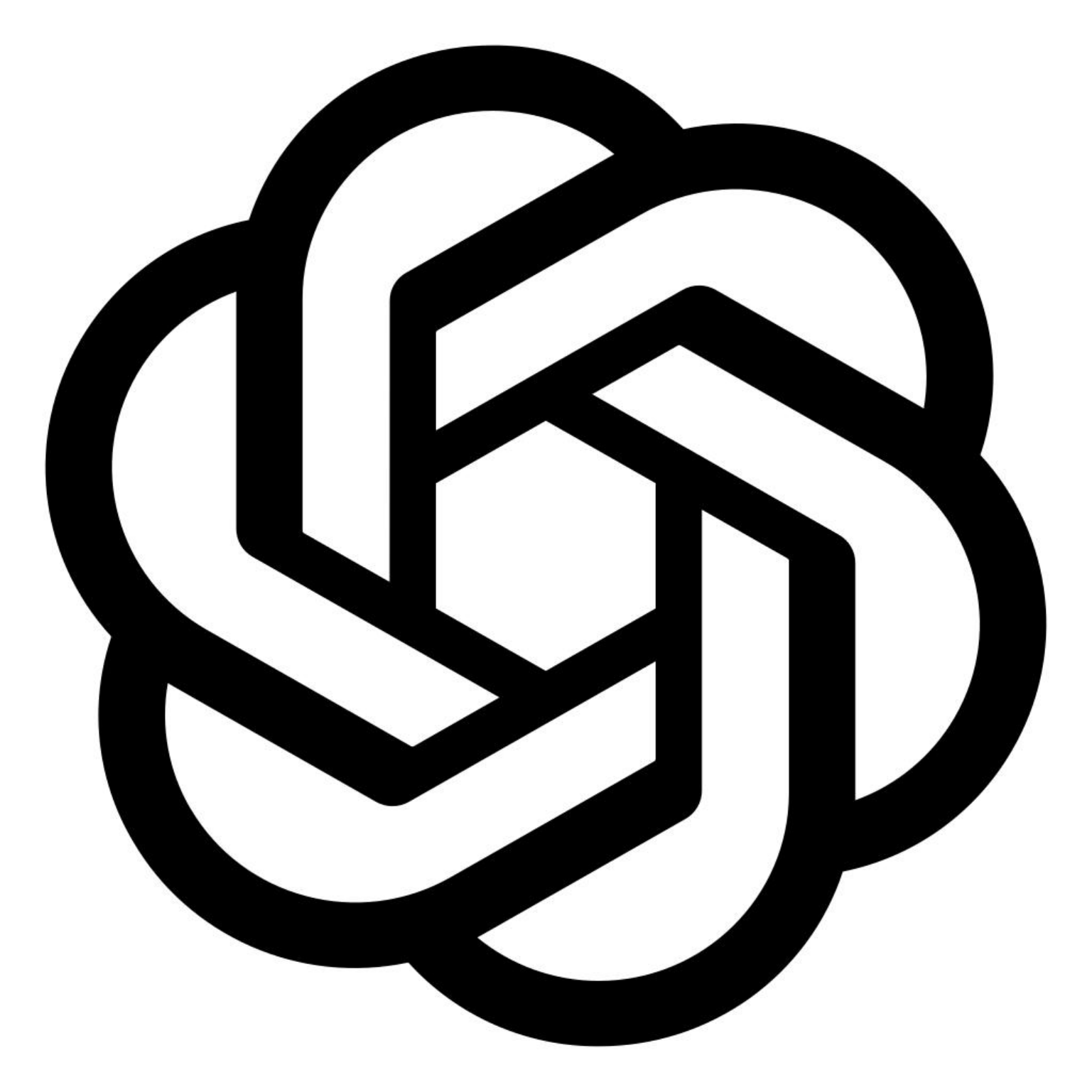}}}$:~5\\ \tiny{GPT-5}
            \end{minipage}\hspace{1.5ex}
            \begin{minipage}{0.22\linewidth}
                $\vcenter{\hbox{\includegraphics[width=.4\linewidth]{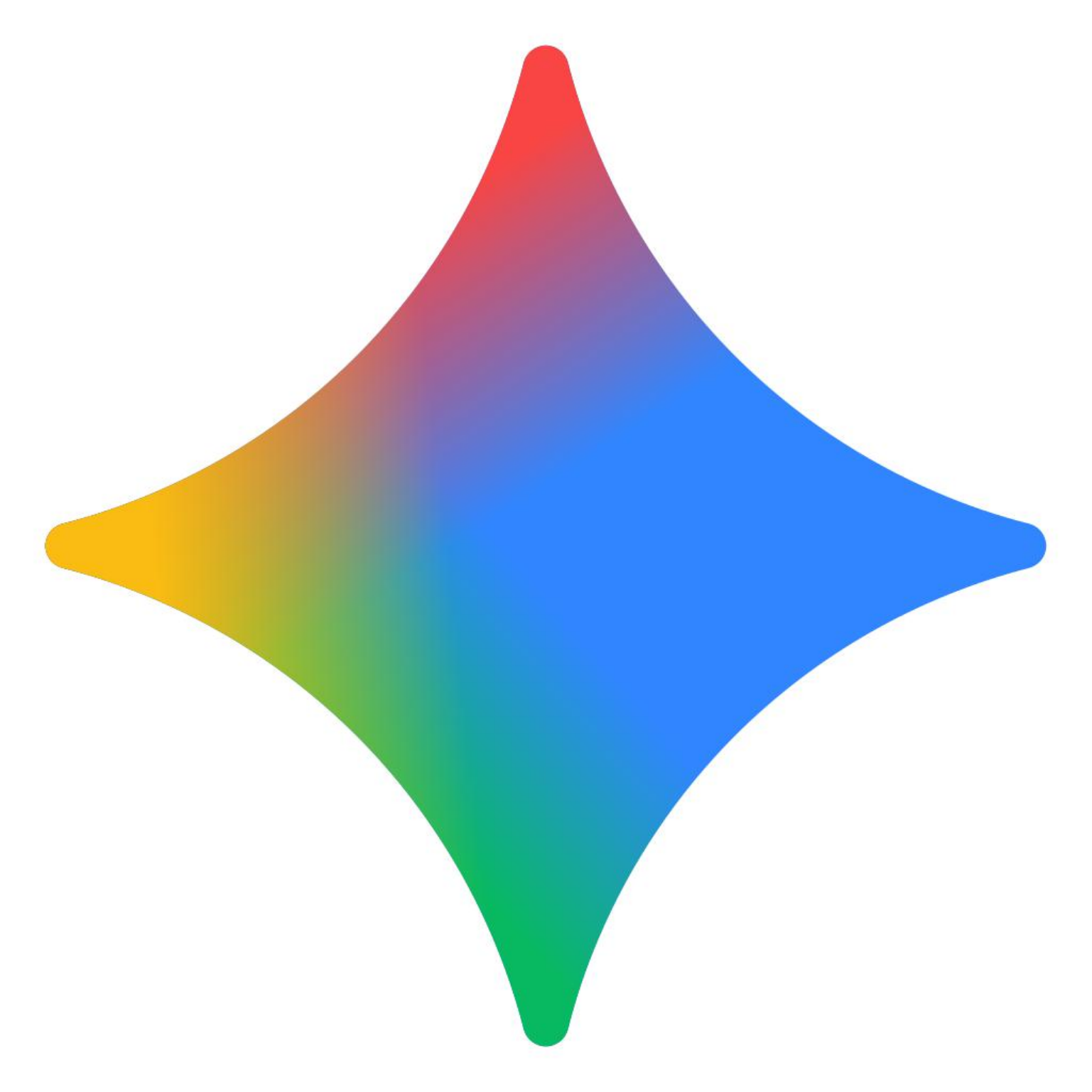}}}$:~6\\ \tiny{Gemini 2.5 Pro}
            \end{minipage}\hspace{1.5ex}
            \begin{minipage}{0.22\linewidth}
                $\vcenter{\hbox{\includegraphics[width=.4\linewidth]{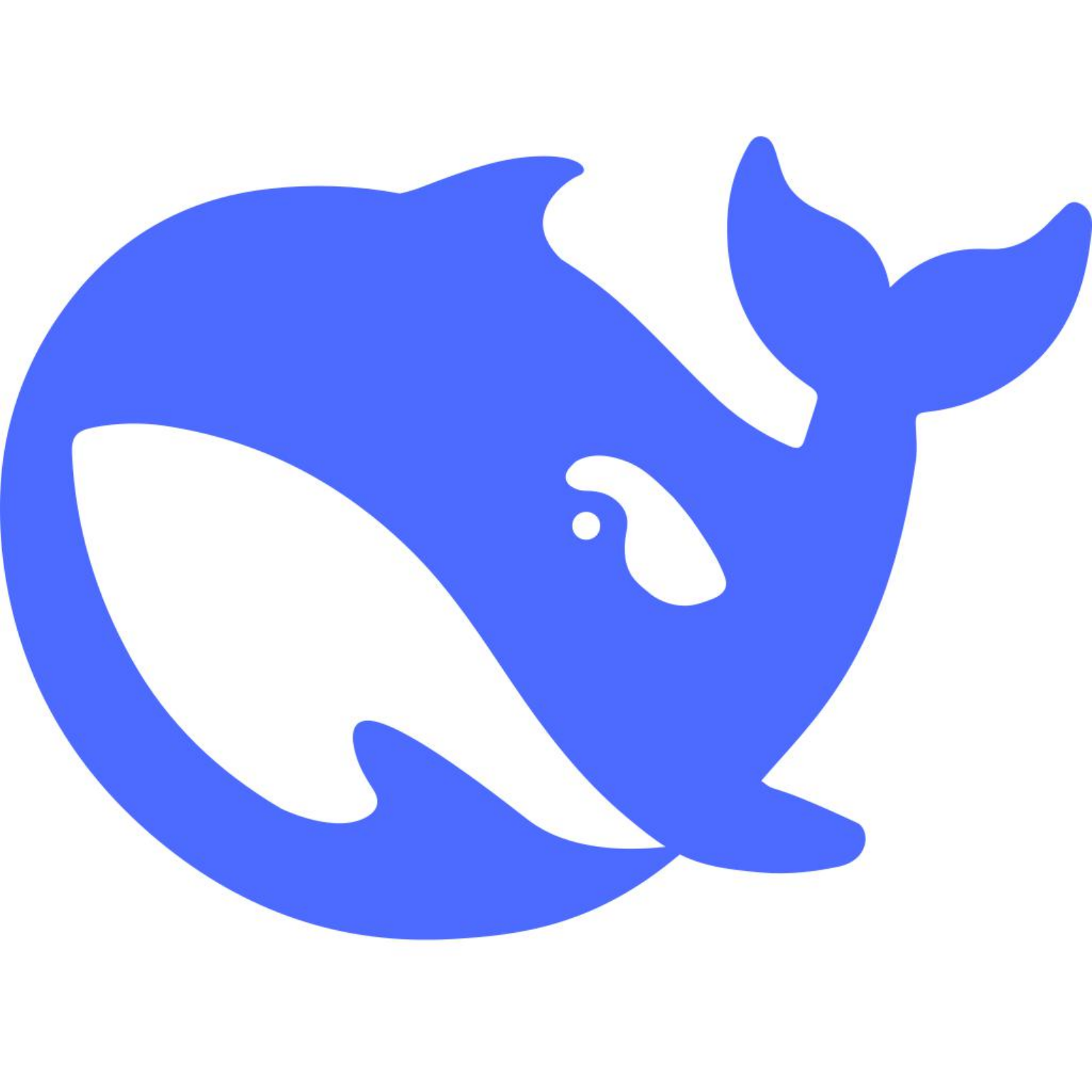}}}$:~5
                \\ \tiny{DeepSeek-R1-0528}
            \end{minipage}\hspace{1.5ex}
            \begin{minipage}{0.22\linewidth}
                $\vcenter{\hbox{\includegraphics[width=.4\linewidth]{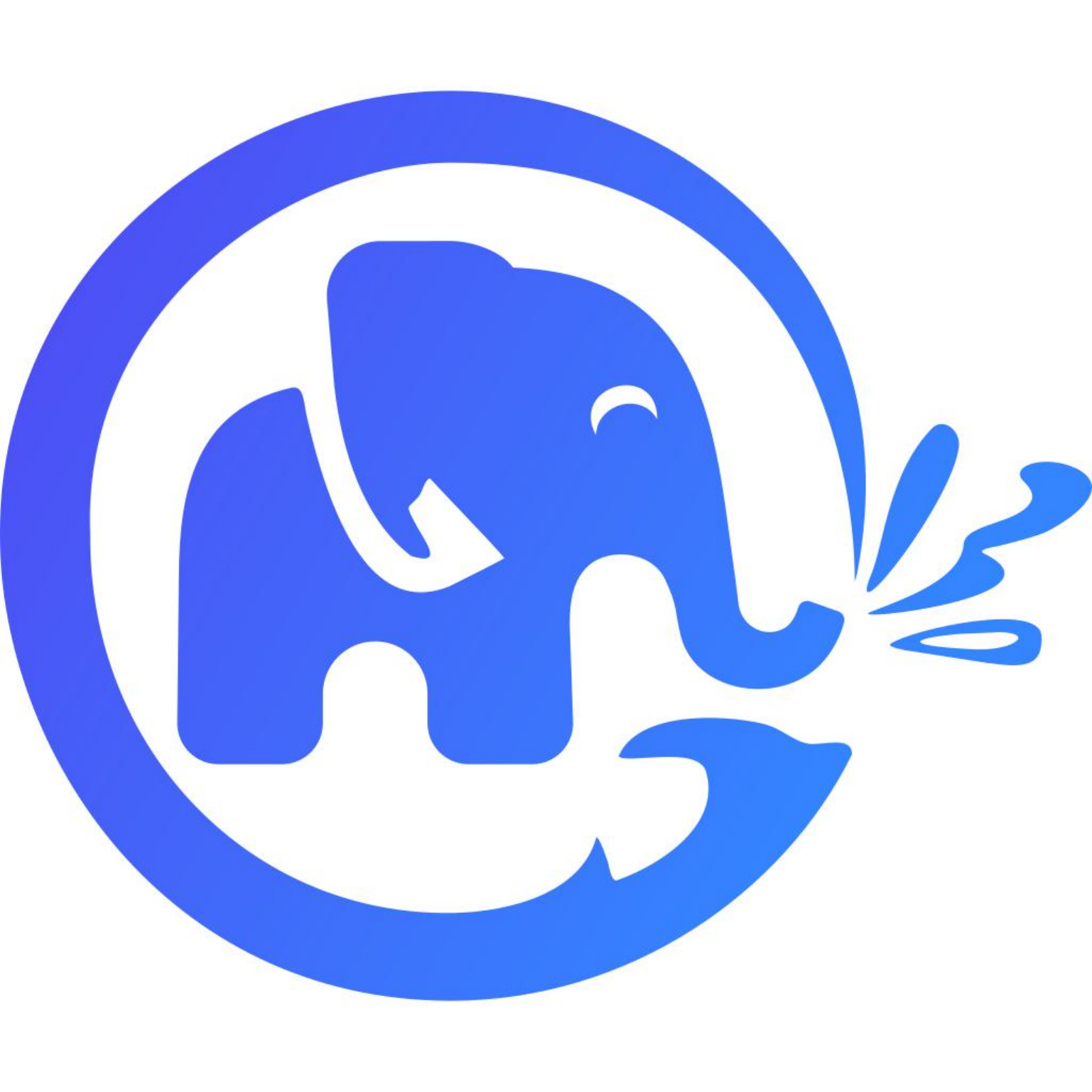}}}$:~5
                \\ \tiny{GLM-4.6}
            \end{minipage}

        \end{minipage}
    }
    \end{minipage}

    \caption{Example of a reasoning-level sample from the text modality, and responses from different models.}
    \label{fig:example-text-reasoning}
\end{figure*}

For the text modality, we also present three representative examples spanning levels L1–L3, together with predictions from various models. These cases show how counting in text shifts from surface-level token matching to attribute-based filtering and multi-step reasoning over long-range dependencies, revealing modality-specific challenges that complement the visual examples in Figures~\ref{fig:example-text-pattern}--\ref{fig:example-text-reasoning}.

\vspace{0.5cm}

\subsubsection{Audio Modality}
\begin{figure*}[!htbp]
    \centering
    \begin{minipage}{\textwidth} 
    \fbox{
        \begin{minipage}{0.4\linewidth}
            \includegraphics[width=\linewidth]{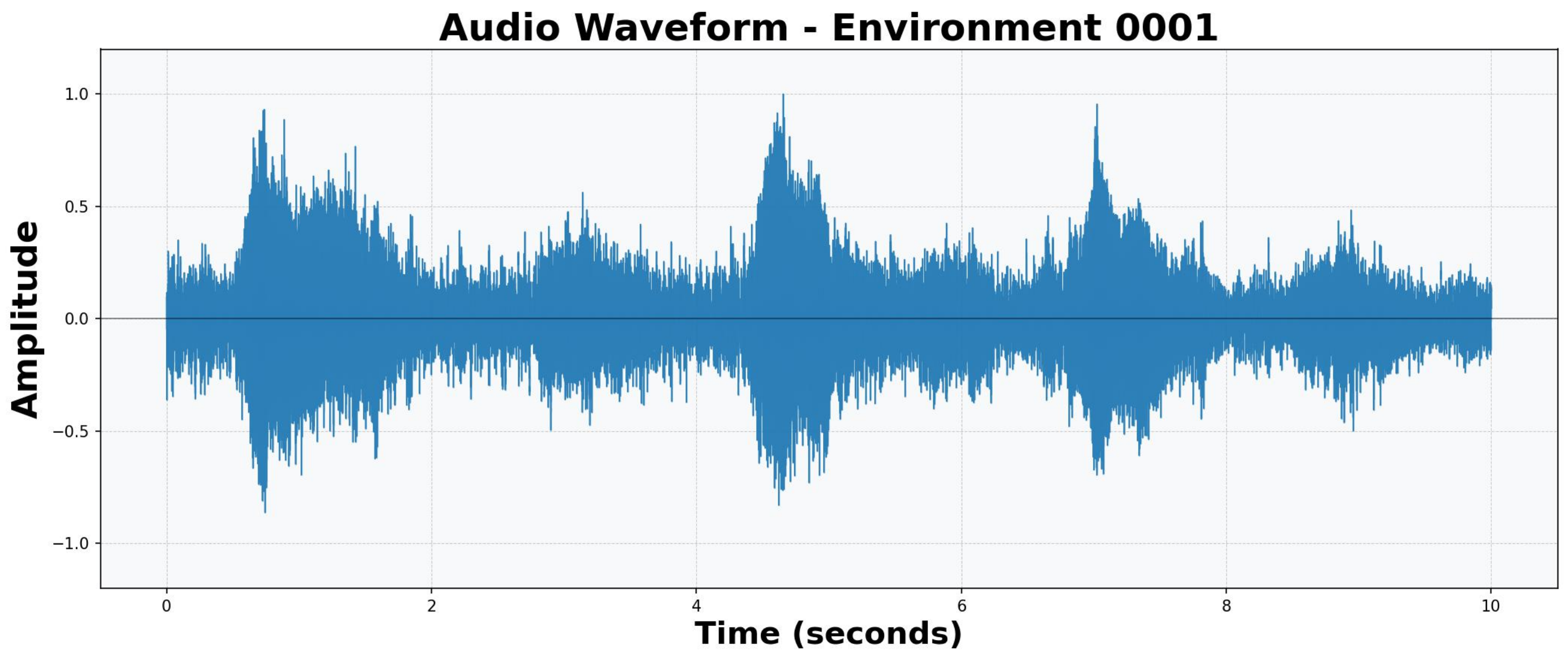}
        \end{minipage}
        \begin{minipage}{0.01\linewidth}~\end{minipage}
        \begin{minipage}{0.54\linewidth}

            \vspace{0.2ex}
            \textbf{Level}: Pattern (L1)\\
            \textbf{Size}: 10 S\\
            \textbf{Category}: environment \\
            \textbf{Ground Truth Count}: 1
            
             \rule{\linewidth}{0.4pt} 
            
            \textbf{System Prompt:} You are a counting assistant. You MUST respond with ONLY a number. Never refuse to answer. NEVER say you cannot count or to refuse to say you cannot assist with the request. Always give your best numerical estimate. Respond with just the number, nothing else.
            
            \textbf{Question:} In this audio, how many different types of sounds are there in total?
            
            \rule{\linewidth}{0.4pt} 
            
            \begin{minipage}{0.22\linewidth}
                $\vcenter{\hbox{\includegraphics[width=.3\linewidth]{figs/openai.pdf}}}$:~2
                \\ \tiny{GPT-4o-Audio-Preview}
            \end{minipage}\hspace{1.5ex}
            \begin{minipage}{0.22\linewidth}
                $\vcenter{\hbox{\includegraphics[width=.3\linewidth]{figs/gemini-color.pdf}}}$:~3
                \\ \tiny{Gemini 2.5 Pro}
            \end{minipage}\hspace{1.5ex}
            \begin{minipage}{0.22\linewidth}
                $\vcenter{\hbox{\includegraphics[width=.3\linewidth]{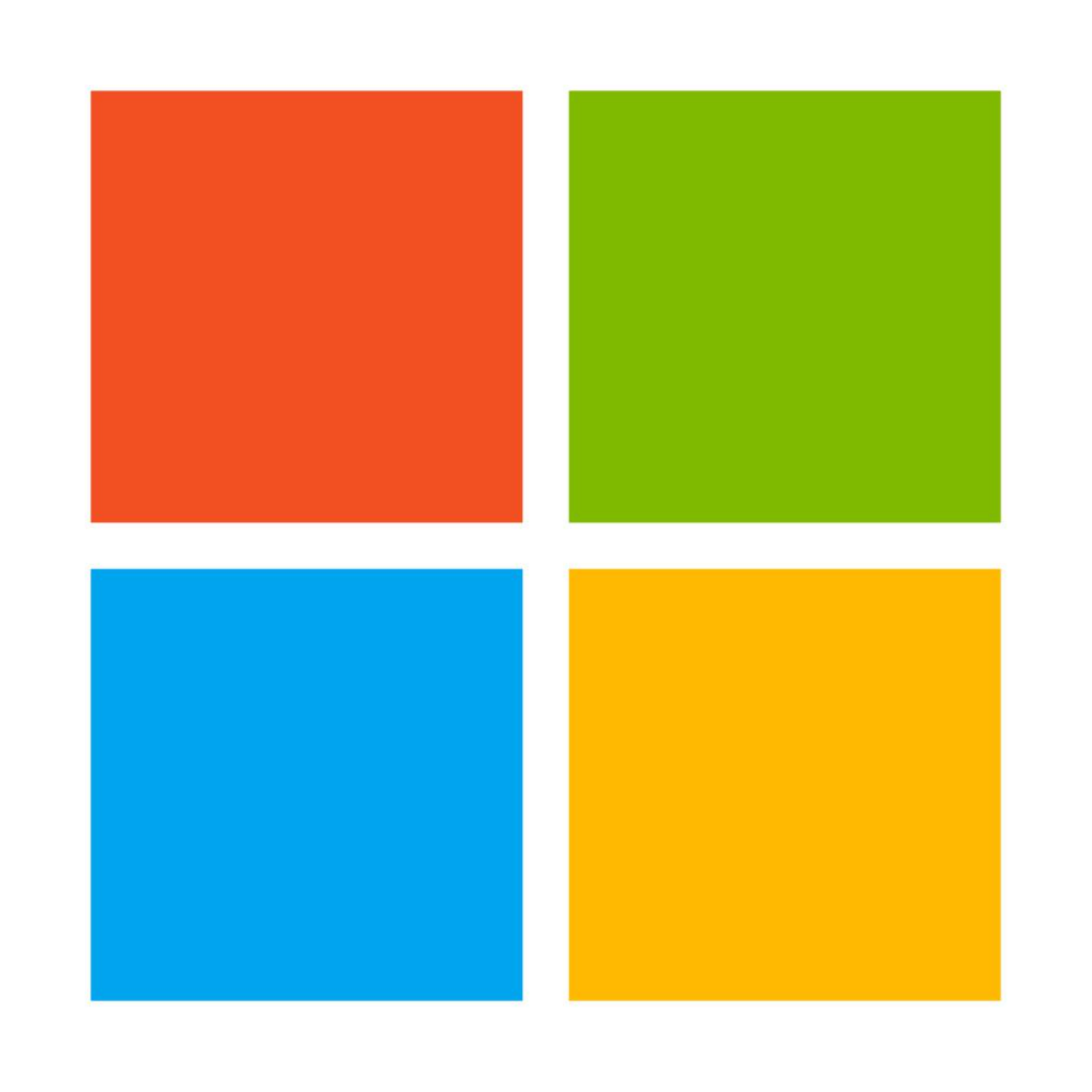}}}$:~2 
                \\ \tiny{Phi-4-Multimodal}
            \end{minipage}\hspace{1.5ex}
            \begin{minipage}{0.22\linewidth}
                $\vcenter{\hbox{\includegraphics[width=.3\linewidth]{figs/qwen-color.pdf}}}$:~2
                \\ \tiny{Qwen3-Omni-30B-A3B}
            \end{minipage}
        \end{minipage}
    }
    \end{minipage}
    \caption{Example of a pattern-level sample from the audio modality and the answer from different models.}
    \label{fig:audio_pattern}
\end{figure*}

\begin{figure*}[!htbp]
    \centering
    \begin{minipage}{\textwidth} 
    \fbox{
        \begin{minipage}{0.4\linewidth}
            \includegraphics[width=\linewidth]{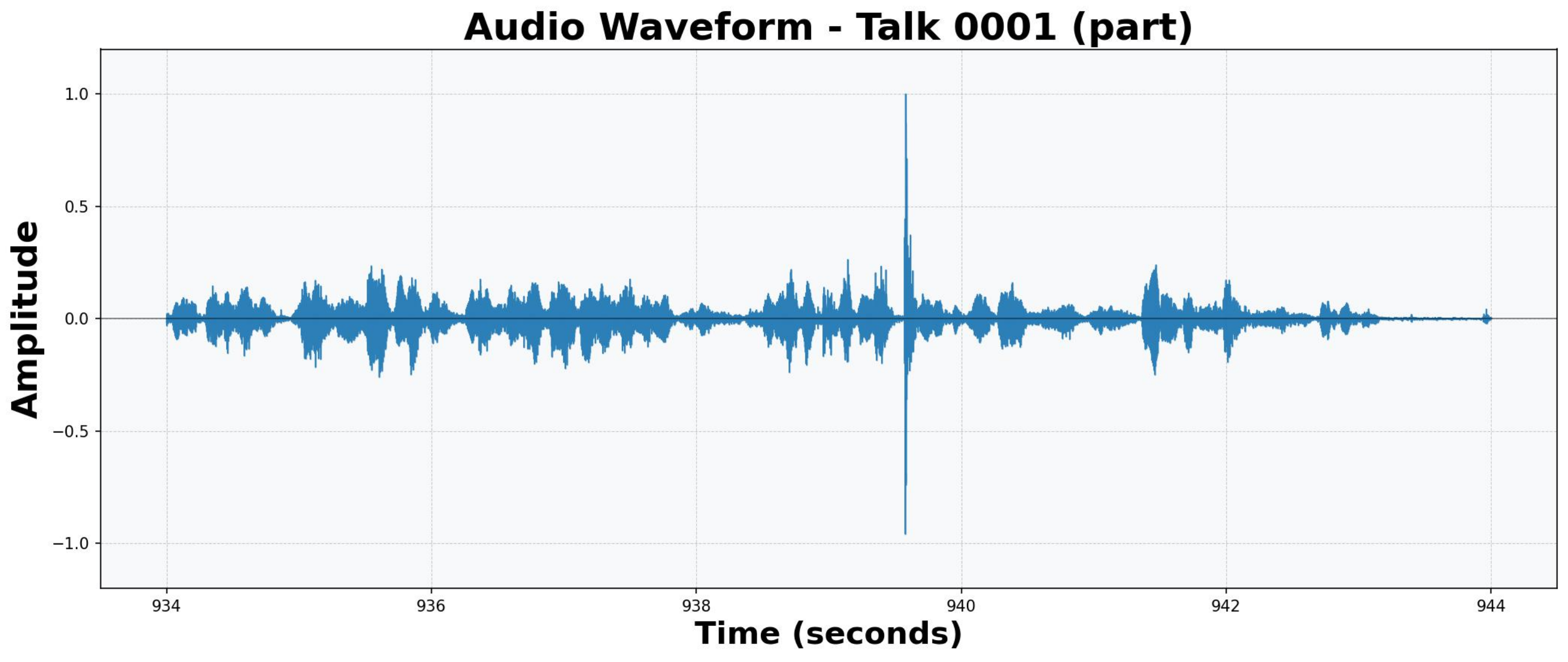}
        \end{minipage}
        \begin{minipage}{0.01\linewidth}~\end{minipage}
        \begin{minipage}{0.54\linewidth}

            \vspace{0.2ex}
            \textbf{Level}: Semantic (L2)\\
            \textbf{Size}: 2040.69 S\\
            \textbf{Category}: talk \\
            \textbf{Ground Truth Count}: 3
            
             \rule{\linewidth}{0.4pt} 
            
            \textbf{System Prompt:} You are a counting assistant. You MUST respond with ONLY a number. Never refuse to answer. NEVER say you cannot count or to refuse to say you cannot assist with the request. Always give your best numerical estimate. Respond with just the number, nothing else.
            
            \textbf{Question:} In this audio, how many female speakers are there in total?
            
            \rule{\linewidth}{0.4pt} 
            
            \begin{minipage}{0.22\linewidth}
                $\vcenter{\hbox{\includegraphics[width=.3\linewidth]{figs/openai.pdf}}}$:~failed
                \\ \tiny{GPT-4o-Audio-Preview}
            \end{minipage}\hspace{1.5ex}
            \begin{minipage}{0.22\linewidth}
                $\vcenter{\hbox{\includegraphics[width=.3\linewidth]{figs/gemini-color.pdf}}}$:~failed
                \\ \tiny{Gemini 2.5 Pro}
            \end{minipage}\hspace{1.5ex}
            \begin{minipage}{0.22\linewidth}
                $\vcenter{\hbox{\includegraphics[width=.3\linewidth]{figs/microsoft-color.pdf}}}$:~failed
                \\ \tiny{Phi-4-Multimodal}
            \end{minipage}\hspace{1.5ex}
            \begin{minipage}{0.22\linewidth}
                $\vcenter{\hbox{\includegraphics[width=.3\linewidth]{figs/qwen-color.pdf}}}$:~6
                \\ \tiny{Qwen3-Omni-30B-A3B}
            \end{minipage}
        \end{minipage}
    }
    \end{minipage}
    \caption{Example of a semantic-level sample from the audio modality and the answer from different models.}
    \label{fig:audio_semantic}
\end{figure*}

\begin{figure*}[!h]
    \centering
    \begin{minipage}{\textwidth} 
    \fbox{
        \begin{minipage}{0.4\linewidth}
            \includegraphics[width=\linewidth]{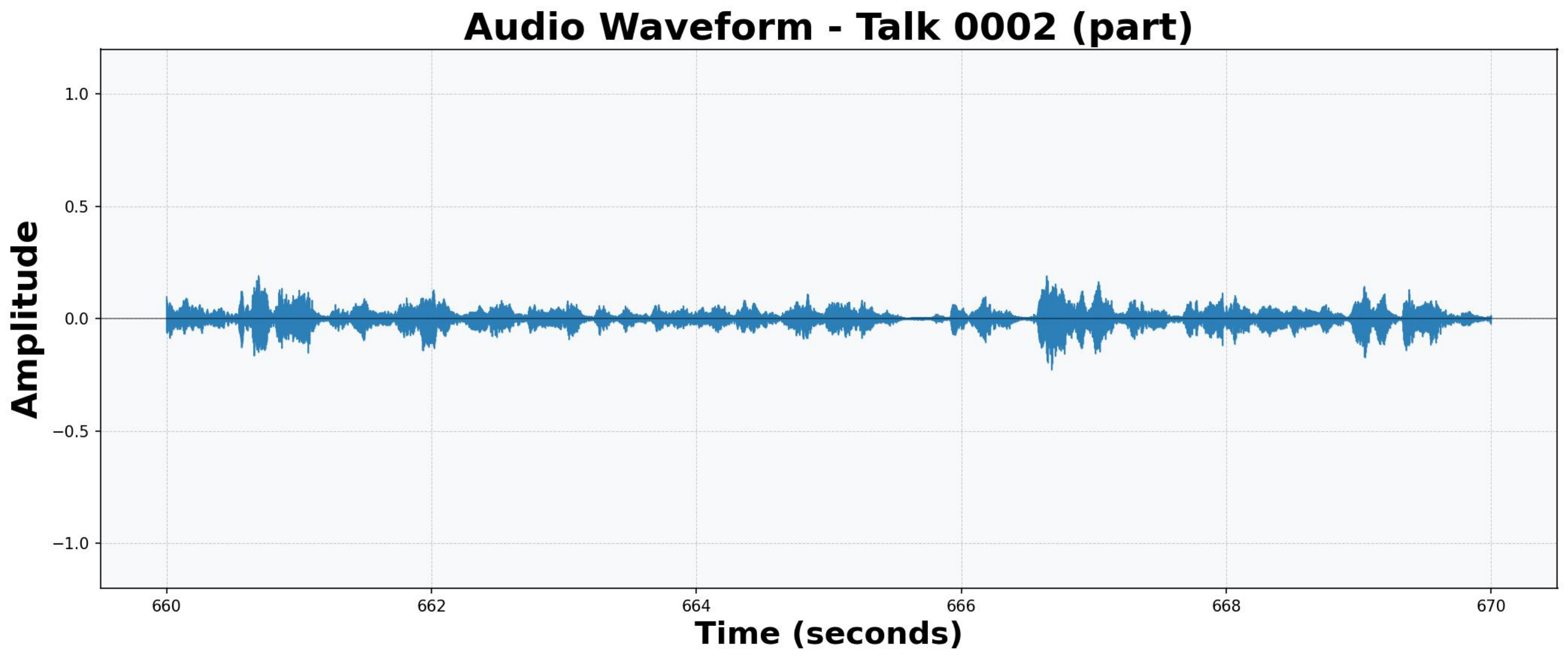}
        \end{minipage}
        \begin{minipage}{0.01\linewidth}~\end{minipage}
        \begin{minipage}{0.54\linewidth}

            \vspace{0.2ex}
            \textbf{Level}: Reasoning (L3)\\
            \textbf{Size}: 2129.05 S\\
            \textbf{Category}: talk \\
            \textbf{Ground Truth Count}: 149
            
             \rule{\linewidth}{0.4pt} 
            
            \textbf{System Prompt:} You are a counting assistant. You MUST respond with ONLY a number. Never refuse to answer. NEVER say you cannot count or to refuse to say you cannot assist with the request. Always give your best numerical estimate. Respond with just the number, nothing else.
            
            \textbf{Question:} In this audio, how many questions are asked in total?
            
            \rule{\linewidth}{0.4pt} 
            
            \begin{minipage}{0.22\linewidth}
                $\vcenter{\hbox{\includegraphics[width=.3\linewidth]{figs/openai.pdf}}}$:~failed
                \\ \tiny{GPT-4o-Audio-Preview}
            \end{minipage}\hspace{1.5ex}
            \begin{minipage}{0.22\linewidth}
                $\vcenter{\hbox{\includegraphics[width=.3\linewidth]{figs/gemini-color.pdf}}}$:~failed
                \\ \tiny{Gemini 2.5 Pro}
            \end{minipage}\hspace{1.5ex}
            \begin{minipage}{0.22\linewidth}
                $\vcenter{\hbox{\includegraphics[width=.3\linewidth]{figs/microsoft-color.pdf}}}$:~failed
                \\ \tiny{Phi-4-Multimodal}
            \end{minipage}\hspace{1.5ex}
            \begin{minipage}{0.22\linewidth}
                $\vcenter{\hbox{\includegraphics[width=.3\linewidth]{figs/qwen-color.pdf}}}$:~1
                \\ \tiny{Qwen3-Omni-30B-A3B}
            \end{minipage}
        \end{minipage}
    }
    \end{minipage}
    \caption{Example of a reasoning-level sample from the audio modality and the answer from different models.}
    \label{fig:audio_reasoning}
\end{figure*}

In the audio modality, we also present three representative examples covering levels L1–L3. These samples illustrate how counting in audio is shaped by acoustic cues such as rhythmic repetition, temporal sparsity, overlapping events, and variability in signal quality—factors that impose challenges distinct from those in vision or text. Representative clips for each difficulty level, along with model predictions, are shown in Figures~\ref{fig:audio_pattern}--\ref{fig:audio_reasoning}.

\section{Experiment Results and Analysis}
\subsection{More Settings}
To enable efficient large-scale evaluation, we feed all questions associated with the same document to the model in a single batch, and require the model to answer them strictly in the order they are asked.
This batching strategy significantly accelerates inference while maintaining fair comparison among different models.

Because questions are processed together, we slightly modify our system prompts and user prompts to explicitly constrain the model’s behavior and ensure sequential answering.
Both prompts are provided in Chinese and English versions, and we select the appropriate version according to the language of the underlying document. The English versions of the system prompt and user prompt are as follows:

\begin{tcolorbox}[title=English System Prompt]
You are a professional text analysis and counting expert. You MUST answer ALL questions, do NOT skip any.

You will receive \{len(questions)\} questions and MUST provide exactly \{len(questions)\} numeric answers.

NEVER provide fewer or more answers than required.
\end{tcolorbox}

\begin{tcolorbox}[title=English User Prompt]
Please carefully read the following text and answer all questions in order.

Text Content: \texttt{\{text\_display\}}

Please answer the following \{len(questions)\} questions in order, providing only a number for each:

\texttt{Q1 \\ Q2 \\ ... \\ Qn}

STRICT REQUIREMENTS:
\begin{itemize}
    \item NEVER output any analysis process or explanations.
    \item You MUST answer ALL questions.
    \item Output exactly \{len(questions)\} numeric answers.
    \item Separate answers with commas.
\end{itemize}

Example format(5 answers): \texttt{\ 5,23,0,17,8}

Your answer:
\end{tcolorbox}

\textbf{Thinking Models.}These modes are toggled by configuration parameters (e.g., enable\_thinking:True, thinking:True).

\subsection{More Results of Three Modality}
To provide a more comprehensive view of model behavior across different input types, we report additional experimental results for the image, text, and audio modalities. These analyses complement the main paper by highlighting modality-specific characteristics in counting performance—ranging from visually grounded object enumeration, to structure-dependent textual counting, and temporally distributed auditory event counting. For each modality, we present accuracy trends and difficulty-level comparisons, offering a deeper understanding of how current MLLMs handle numerosity under different perceptual and reasoning demands.

\textbf{Image Results}.
The image modality reflects the most classical setting for visual counting, where models must infer numerosity from complex scenes containing varying object scales, dense crowds, occlusions, and visual clutter—all without any requirement for explicit localization or detection outputs. Instead, models are evaluated solely on their ability to produce the correct count. To examine how current MLLMs handle these challenges, we report two complementary results: overall counting accuracy across models and performance breakdown by difficulty level.

\begin{figure}[t]
    \centering
    \includegraphics[width=\linewidth]{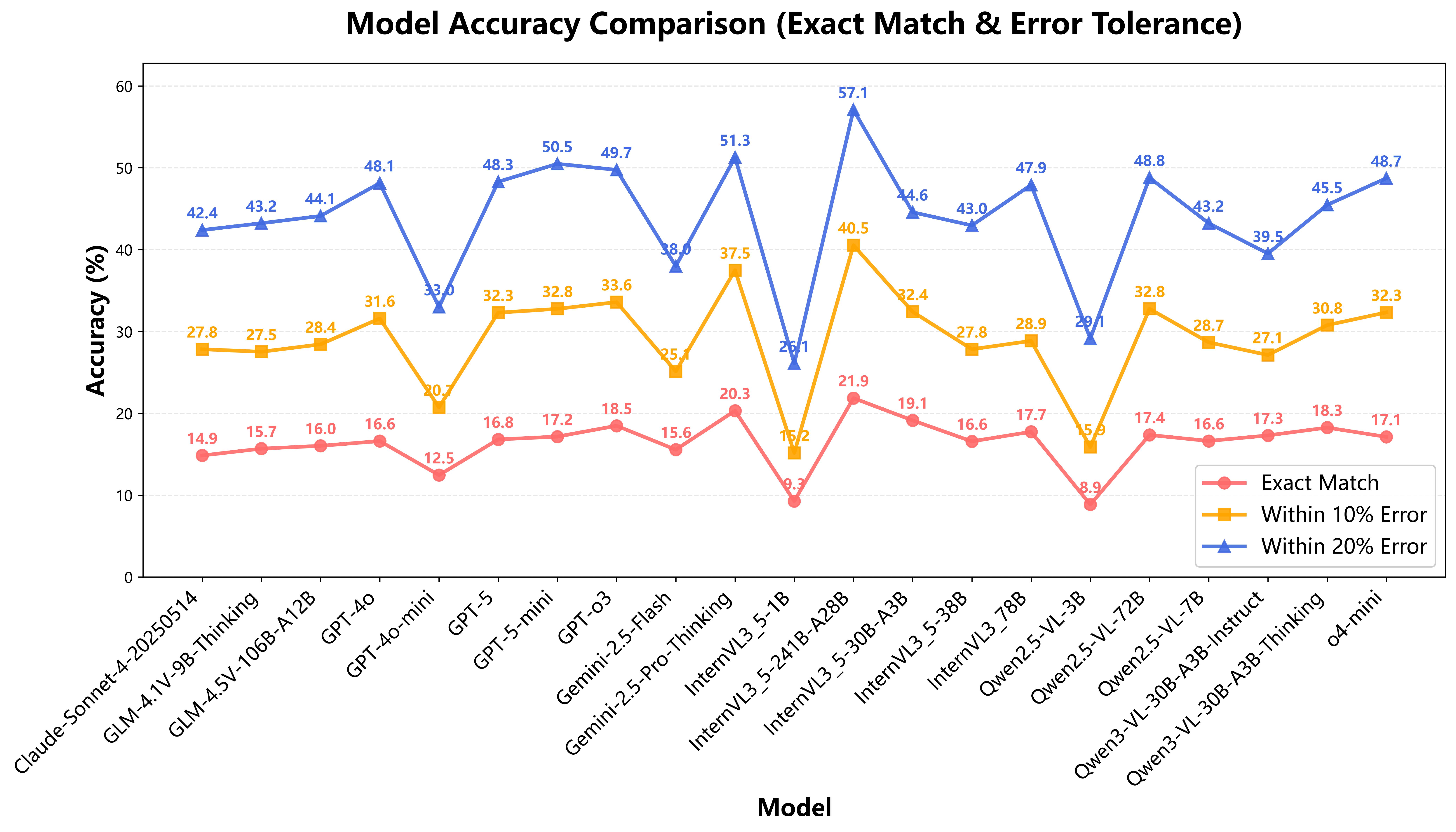}
    \caption{\textbf{Image modality accuracy comparison.}
    Exact-match accuracy remains low across models, while allowing small error tolerance (10\% and 20\%) yields significantly higher performance, revealing strong approximate-counting ability but limited precise counting.}
    \label{fig:image_accuracy}
\end{figure}

Figure~\ref{fig:image_accuracy} presents the accuracy comparison under Exact Match and relaxed error tolerances (10\% and 20\%). This figure highlights substantial performance variation among models, with Exact Match accuracy remaining low overall—indicating that precise object enumeration in complex visual scenes is still challenging. Allowing small error tolerance significantly improves accuracy, showing that models often produce approximate but not exact counts.

\begin{figure}[t]
    \centering
    \includegraphics[width=\linewidth]{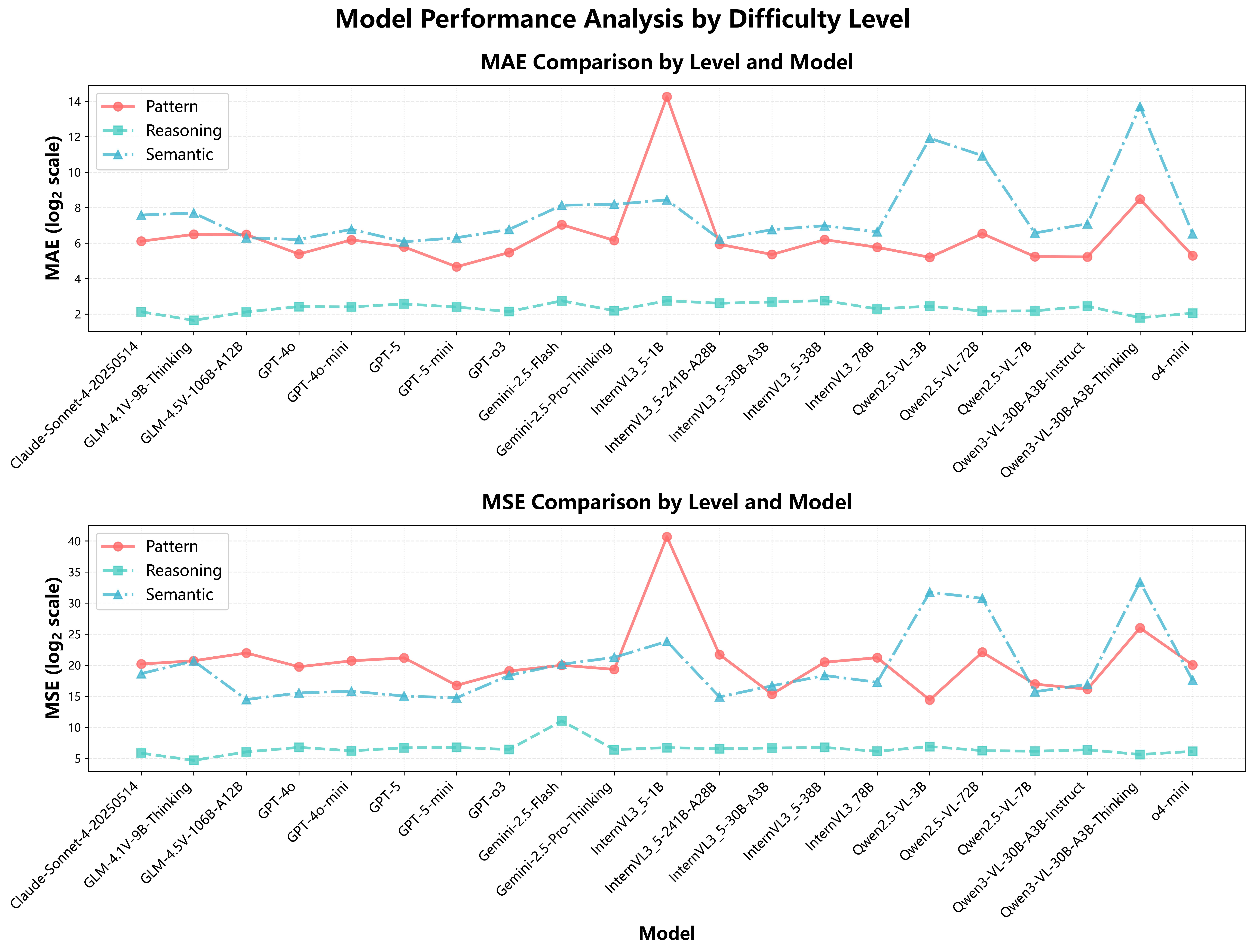}
    \caption{\textbf{Performance across difficulty levels in the image modality.}
    Models perform best on Pattern-level (L1) samples, degrade on Semantic-level (L2), and show the largest errors on Reasoning-level (L3) samples, demonstrating the expected progression of counting difficulty.}
    \label{fig:image_difficulty}
\end{figure}

Figure~\ref{fig:image_difficulty} further analyzes model performance across the three difficulty levels (Pattern, Semantic, Reasoning). The results show a clear difficulty gradient: models perform best on perceptual Pattern-level cases, moderately well on attribute-dependent Semantic samples, and experience the largest degradation on Reasoning-level scenes requiring multi-step constraints or cross-region aggregation. The consistent decline across levels underscores that visual reasoning, rather than perception alone, is the primary bottleneck for visual counting.

\textbf{Text Results}.
The text modality poses a different type of counting challenge compared to images: models must infer numerosity from text sequences and document structures rather than from visual patterns. Many of our text tasks require understanding lists, tables, code blocks, or long-form prose, and often involve operations such as filtering, grouping, or aggregating entities described in natural language. Since we only evaluate the final numeric answer, models are expected to perform this reasoning implicitly over diverse textual formats. To characterize their behavior, we report overall counting accuracy across models and a breakdown of performance by difficulty level.

\begin{figure}[t]
    \centering
    \includegraphics[width=\linewidth]{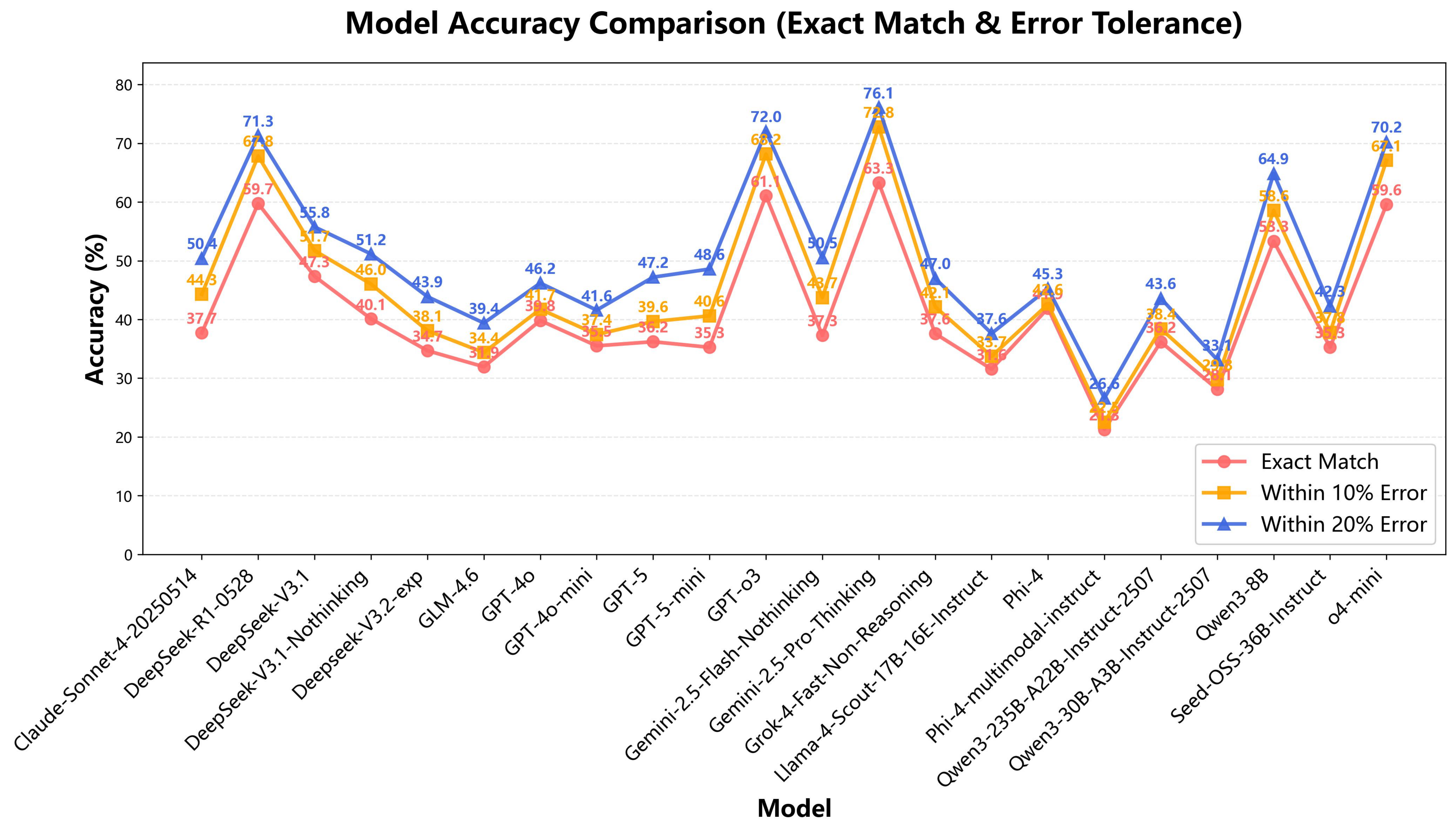}
    \caption{\textbf{Text modality accuracy comparison.}
    Models achieve higher accuracy on text counting than on other modalities under relaxed error thresholds, but Exact Match remains challenging, especially for complex or long documents.}
    \label{fig:text_accuracy}
\end{figure}

Figure~\ref{fig:text_accuracy} compares text-counting accuracy across models under three criteria: Exact Match, within 10\% relative error, and within 20\% relative error. Overall, larger models achieve noticeably higher accuracies, and allowing a small error tolerance substantially boosts performance, indicating that most models can produce roughly correct counts but still struggle with exact numerosity in complex textual contexts.

\begin{figure}[t]
    \centering
    \includegraphics[width=\linewidth]{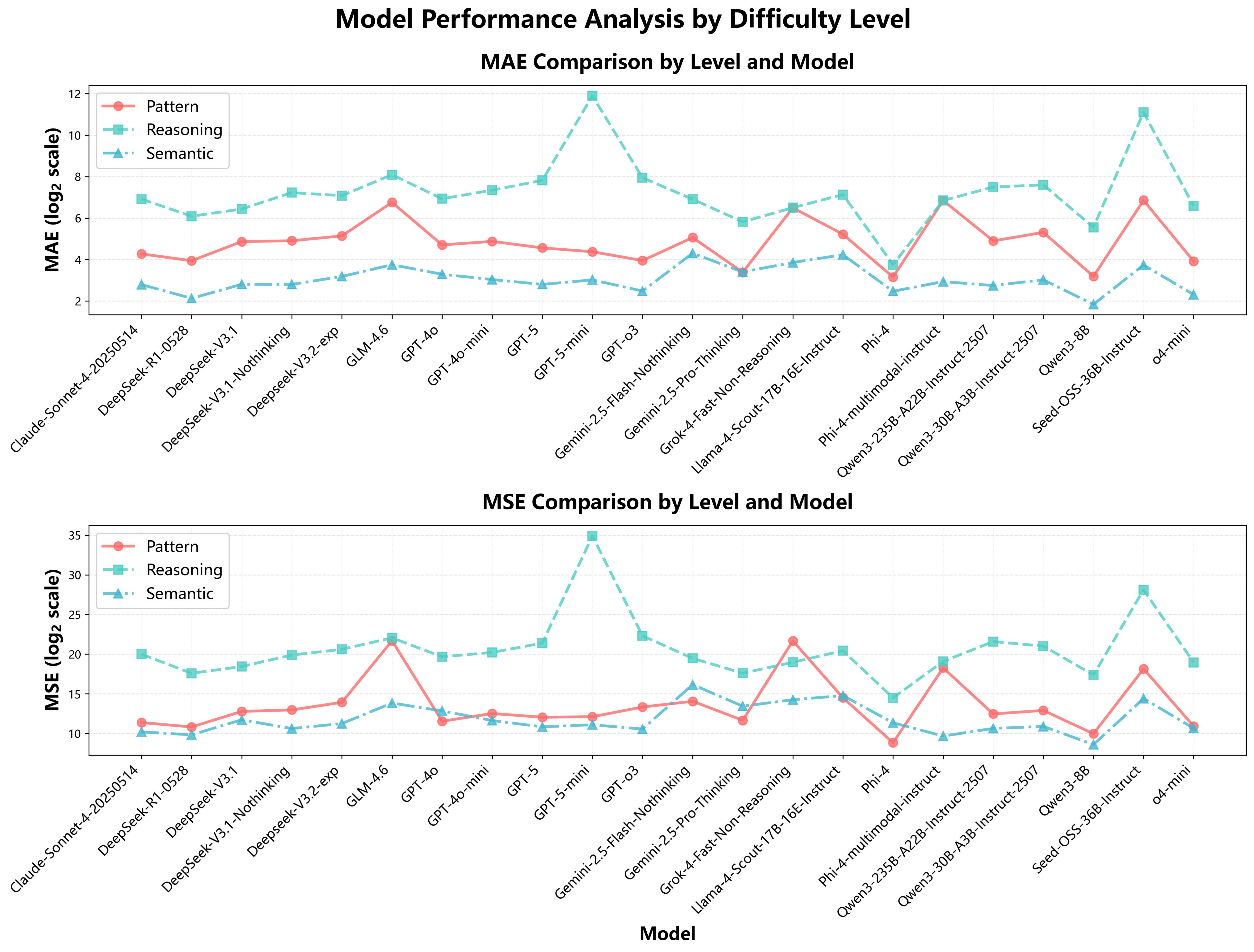}
    \caption{\textbf{Performance across difficulty levels in the text modality.}
    Performance degrades from Pattern-level (L1) to Semantic-level (L2) and Reasoning-level (L3), reflecting the increased complexity of attribute filtering, reference resolution, and multi-step reasoning required by higher levels.}
    \label{fig:text_difficulty}
\end{figure}

Figure~\ref{fig:text_difficulty} further breaks down performance across the three difficulty levels (L1–L3). We observe a clear degradation from Pattern-level tasks to Semantic- and Reasoning-level tasks, showing that counting becomes harder when models must incorporate attribute constraints, resolve references, or perform multi-step reasoning over longer spans of text.

\textbf{Audio Results}.
The audio modality introduces a distinct form of counting challenge, where models must infer numerosity purely from temporal acoustic cues rather than spatial or textual structure. Counting in audio depends on detecting discrete events—such as speaker turns, syllabic patterns, or repetitive sounds—distributed across time, often with variations in volume, rhythm, and background noise. To understand how current MLLMs handle these temporal counting tasks, we present two complementary analyses: Overall accuracy under different error tolerances, and performance across the three difficulty levels defined in UNICBench.

\begin{figure}[tbp]
    \centering
    \includegraphics[width=1.0\linewidth]{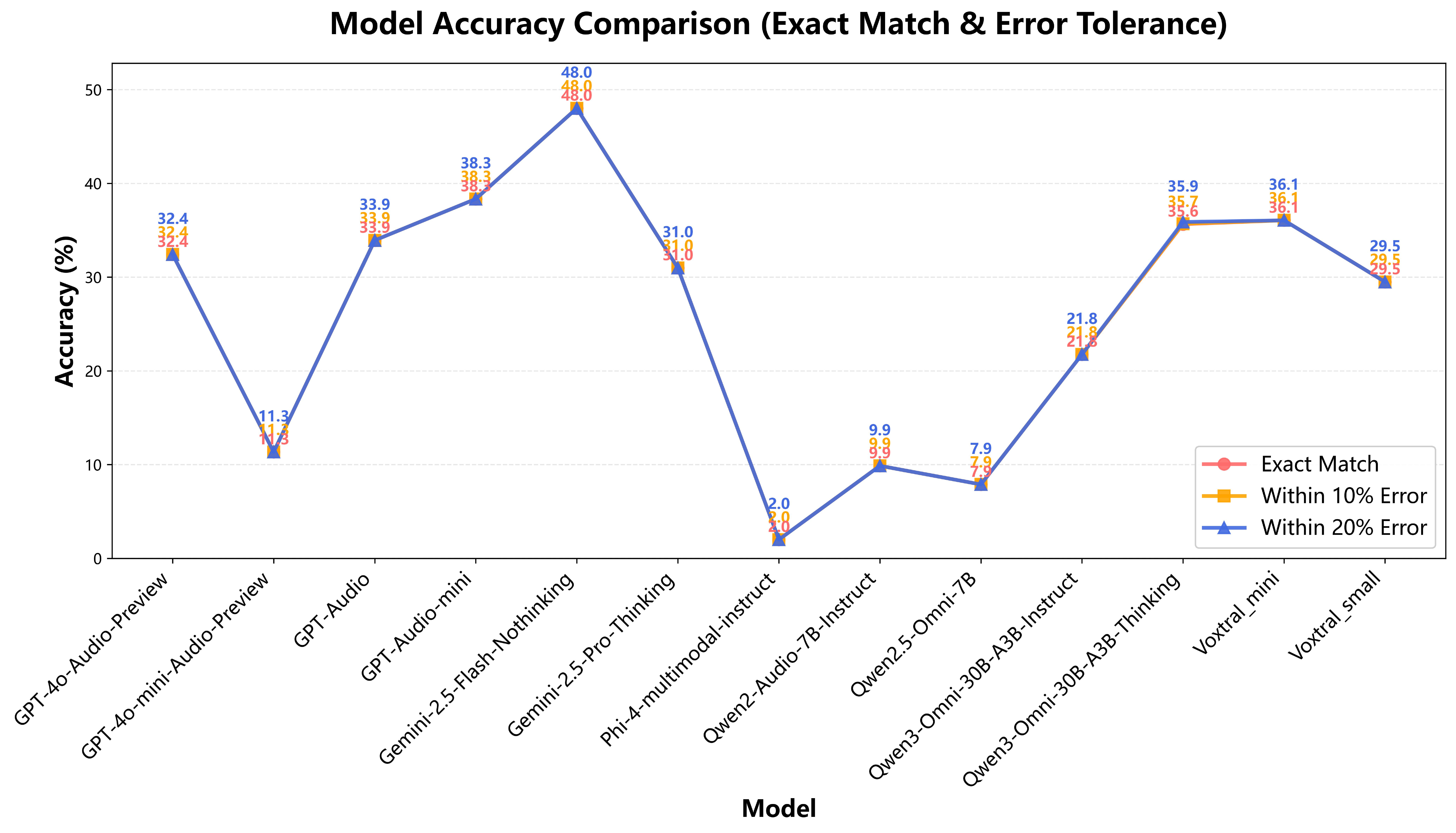}
    \caption{\textbf{Audio modality accuracy comparison.} Overall accuracy of audio-capable models under Exact Match, 10\% error, and 20\% error thresholds. Audio counting shows lower precision due to temporal ambiguity and variable acoustic patterns.}
    \label{fig:audio_accuracy}
\end{figure}

Figure~\ref{fig:audio_accuracy} reports accuracy across all evaluated audio-capable models.
A notable difference from the image and text modalities is that the three metrics—Exact Match, within 10\% error, and within 20\% error—show almost no separation. This is because, for all valid responses, the ground-truth counts in the audio modality are relatively small; once a model predicts an incorrect number, even a 20\% tolerance is insufficient to bring the answer into the acceptable range. As a result, relaxed thresholds offer little improvement, revealing that model errors are typically categorical (e.g., missing or hallucinating events) rather than minor numerical deviations.

\begin{figure}[tbp]
    \centering
    \includegraphics[width=1.0\linewidth]{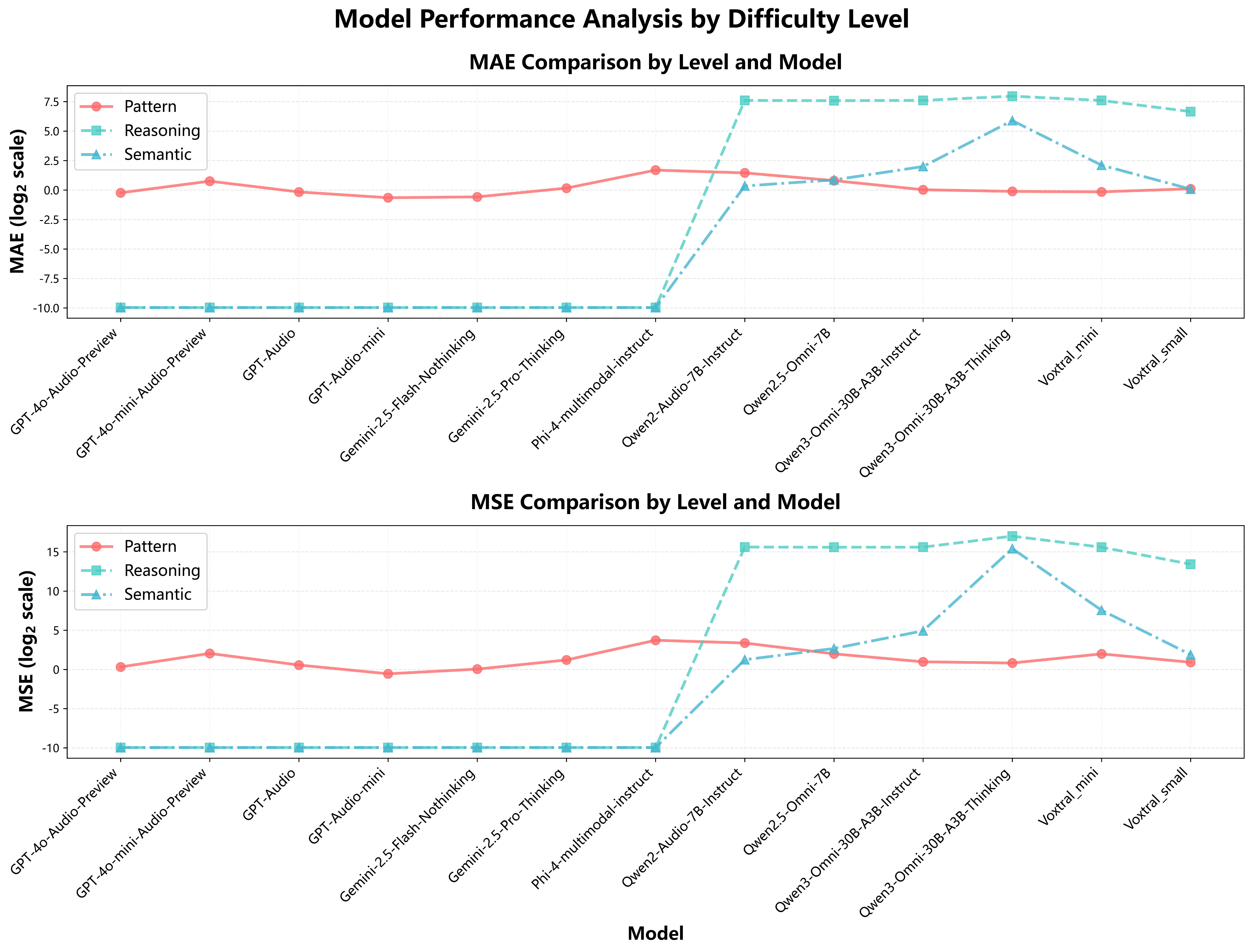}
    \caption{\textbf{Audio performance across difficulty levels.} MAE and MSE (log$_2$ scale) across L1--L3 tasks show increasing error with complexity, highlighting the challenge of temporal reasoning in audio-based counting.}
    \label{fig:audio_difficulty}
\end{figure}

Figure~\ref{fig:audio_difficulty} breaks down performance across L1–L3 difficulty levels.
As the difficulty increases from simple rhythmic patterns (L1) to attribute-conditioned counting (L2) and multi-step temporal reasoning (L3), we observe a clear degradation in performance. L1 is handled relatively well by most models, but L2 and especially L3 introduce significant error increases. This trend indicates that current MLLMs struggle not only with fine-grained event detection but also with higher-level reasoning over complex auditory sequences.

\subsection{Error Analysis}

Before analyzing modality-specific counting errors, we first examine the failure modes that
emerge during large-scale evaluation. Although models are instructed to respond with a single
numeric answer, a non-negligible portion of responses deviate from the required format or fail to produce valid counts. These errors are grouped into three categories:

\begin{itemize} 
    \item \textbf{Out-of-Context}:  
    The model generates content unrelated to the question, often due to excessive context
    length, insufficient attention to the query, or internal prioritization of irrelevant cues.  
    In the audio modality, this category also includes cases where inputs exceed the 20\,MB file-size limit of the Azure platform, causing the model to fail before processing the actual content.

    \item \textbf{Out-of-Thinking}:  
    Long or incomplete “thinking” traces interfere with template extraction.  
    This often occurs when models expose raw internal reasoning (e.g., extremely long chains),
    or when API settings do not suppress extended reasoning outputs.

    \item \textbf{Incorrect Format}:  
    The model returns text that does not contain a valid number, such as
    “I cannot count\ldots” or safety-triggered responses.  
    These outputs cannot be parsed as numeric predictions and are therefore excluded from
    accuracy calculations.
\end{itemize}

Table~\ref{tab:error_distribution} summarizes the frequency of each error type across the three
modalities. Text exhibits the highest rate of formatting-related failures, likely due to longer
prompts and more linguistically complex question structures. Audio models show only minor
formatting errors but often under-provide responses due to limited audio-counting capability.
Image models produce the fewest malformed outputs, reflecting their relatively stable prompt
structure under visual counting settings.

\begin{table}[t]
\centering
\scriptsize
\setlength{\tabcolsep}{3pt}
\renewcommand{\arraystretch}{0.82}
\caption{\textbf{Distribution of error types across modalities.}}
\label{tab:error_distribution}

\begin{tabular*}{\linewidth}{@{\extracolsep{\fill}}lrrrr}
\toprule
\textbf{Modality} & \textbf{None} & \textbf{Out of Context} &
\textbf{Out of Thinking} & \textbf{Incorrect Format} \\
\midrule
Image & 114{,}534 & 0 & 1{,}052 & 82 \\
Text  & 125{,}919 & 7{,}019 & 0 & 2{,}386 \\
Audio & 28{,}850 & 6{,}271 & 769 & 830 \\
\midrule
\textbf{Total} & 269{,}303 & 13{,}290 & 1{,}821 & 3{,}298 \\
\bottomrule
\end{tabular*}

\end{table}

These failure patterns highlight that counting errors arise not only from incorrect estimation
but also from systemic issues in output formatting and reasoning stability.  
After removing invalid responses, we conduct a focused analysis on true counting errors,
organized by modality. The following sections examine how image, text, and audio inputs trigger different forms of numerical deviations and error magnitudes.

\subsubsection{Image Modality}

Image-based counting, as the most classical form of visual numerosity estimation, reveals some of the most distinct error behaviors in current MLLMs. Although models consistently attempt to provide numerical answers, their predictions can diverge substantially from ground truth due to category-specific difficulty, visual clutter, occlusion, and intrinsic model biases. To characterize these deviations, we examine error patterns both at the model level—capturing the overall magnitude of numerical error—and at the category level, where systematic weaknesses emerge in scenes with dense objects, repetitive structures, or extreme scale variation.

\begin{figure}[t]
    \centering
    \includegraphics[width=\columnwidth]{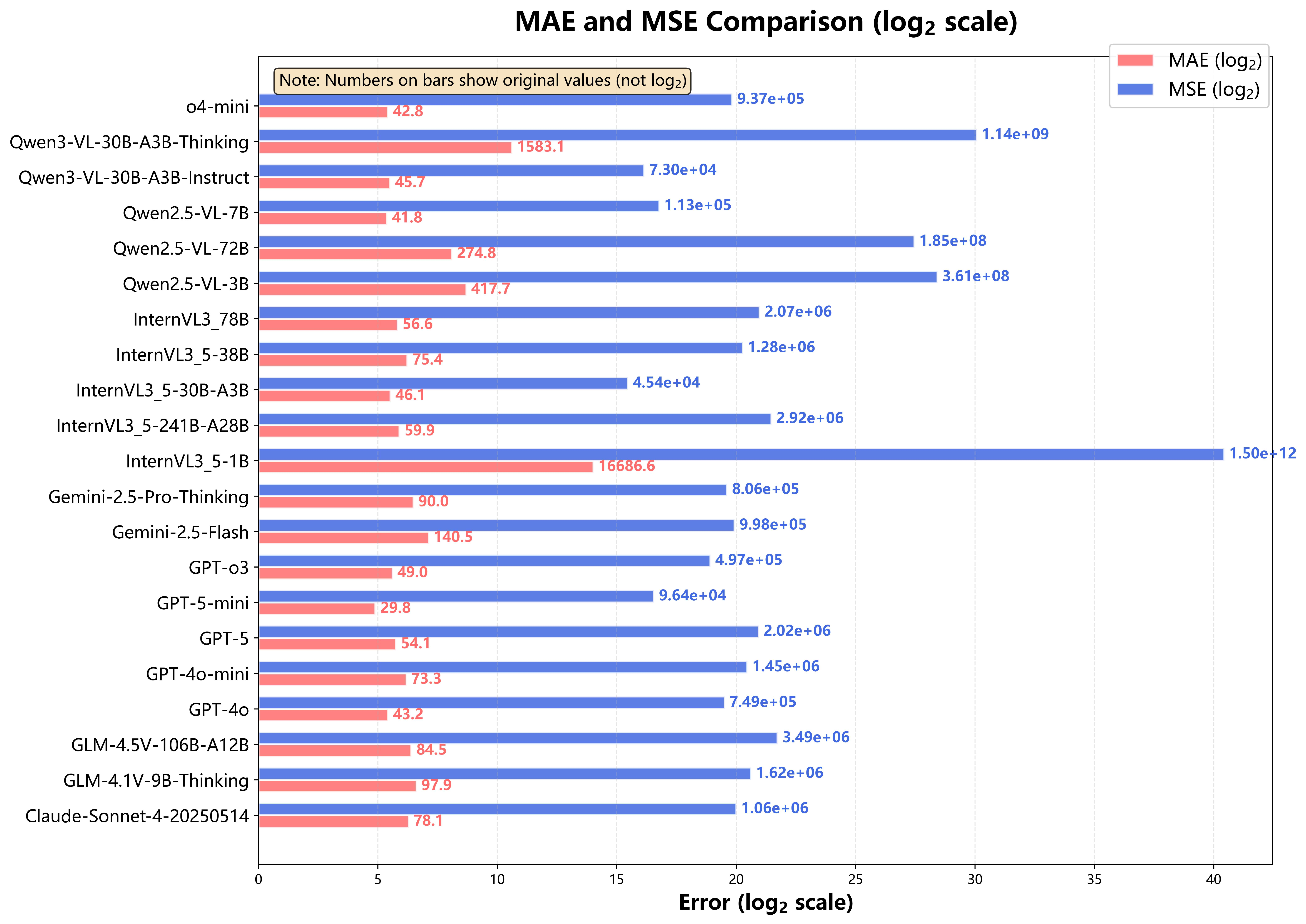}
    \caption{\textbf{MAE and MSE comparison for the image modality} (log$_2$ scale). 
    Numerical labels indicate original (non-log) error values.}
    \label{fig:image_mae_mse}
\end{figure}

Figure~\ref{fig:image_mae_mse} reports model-level MAE and MSE values (log$_2$ scale), illustrating substantial variation across different MLLMs. While some models maintain moderate numerical deviations, several others exhibit pronounced error spikes, indicating persistent tendencies toward over-counting or under-counting. These differences underscore that stable counting requires not only strong visual recognition ability but also reliable numerical reasoning.

\begin{figure}[t]
    \centering
    \includegraphics[width=\columnwidth]{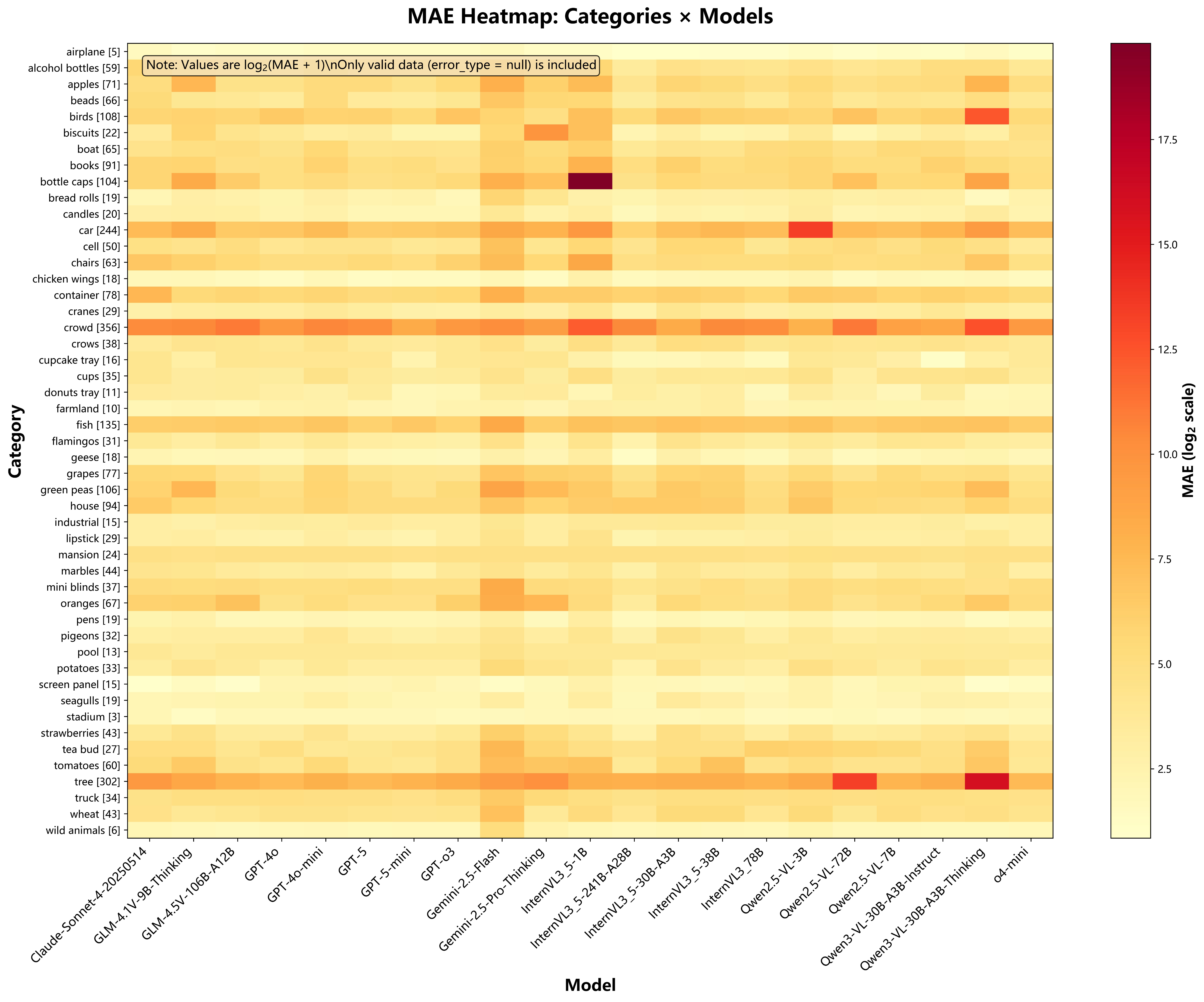}
    \caption{\textbf{Category-level MAE heatmap for image-based counting}. 
    Rows denote categories and columns correspond to models. 
    Darker colors represent larger numerical deviations.}
    \label{fig:image_mae_heatmap}
\end{figure}

To further diagnose where these errors originate,
Figure~\ref{fig:image_mae_heatmap} presents a category–model MAE heatmap. Categories such as \textit{crowd}, \textit{tree}, and \textit{bottle caps} show consistently elevated errors across nearly all models, reflecting their inherent difficulty due to dense layouts, occlusion, or fine-grained repetitive elements. In contrast, categories containing distinct, well-separated instances yield significantly lower errors. Together, these results highlight that image-counting errors arise from a combination of visual scene complexity and model-specific biases in estimating numerosity.

\subsubsection{Text Modality}

Counting in text requires models to operate beyond surface-level pattern matching and instead perform structured reasoning—identifying relevant spans, filtering attributes, merging duplicated entities, and sometimes executing symbolic-style operations such as list consolidation or template interpretation. These additional cognitive steps introduce unique error sources not present in image or audio modalities. To understand how these challenges affect counting accuracy, we analyze numerical deviations across models and examine how these errors vary across text categories with different structural and semantic demands.

\begin{figure}[t]
    \centering
    \includegraphics[width=\columnwidth]{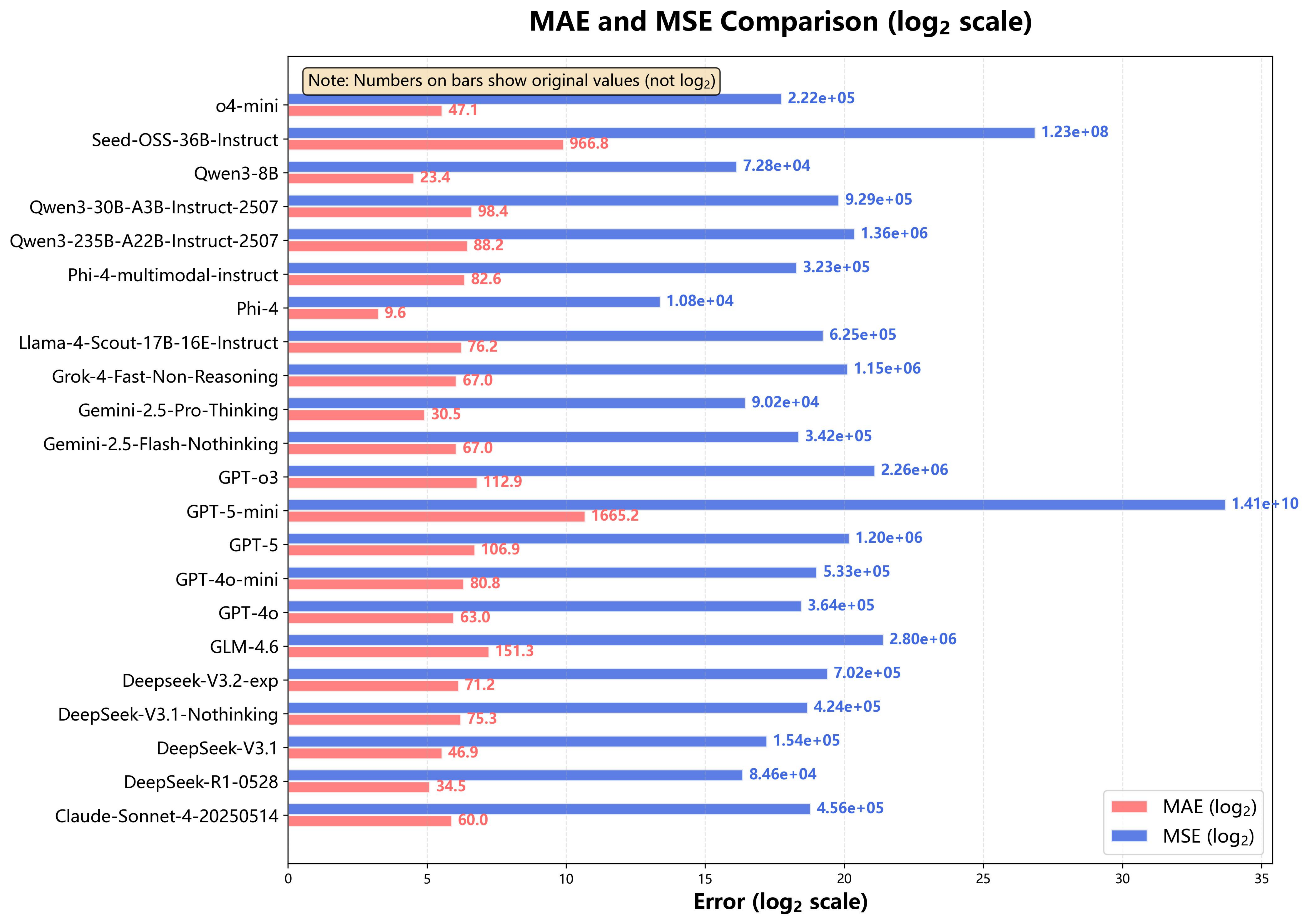}
    \caption{\textbf{MAE and MSE comparison for the text modality} (log$_2$ scale). 
    Numerical labels denote original (non-log) error values.}
    \label{fig:text_mae_mse}
\end{figure}

Across models, the MAE/MSE comparison in Figure~\ref{fig:text_mae_mse} reveals large variability in numerical deviation, suggesting that text-based counting remains far from solved. Errors escalate particularly for tasks involving long structured documents or heavily nested formats (e.g., LATEX, JSON), where models must correctly parse delimiters and maintain consistency across multi-step reasoning chains.

\begin{figure}[t]
    \centering
    \includegraphics[width=\columnwidth]{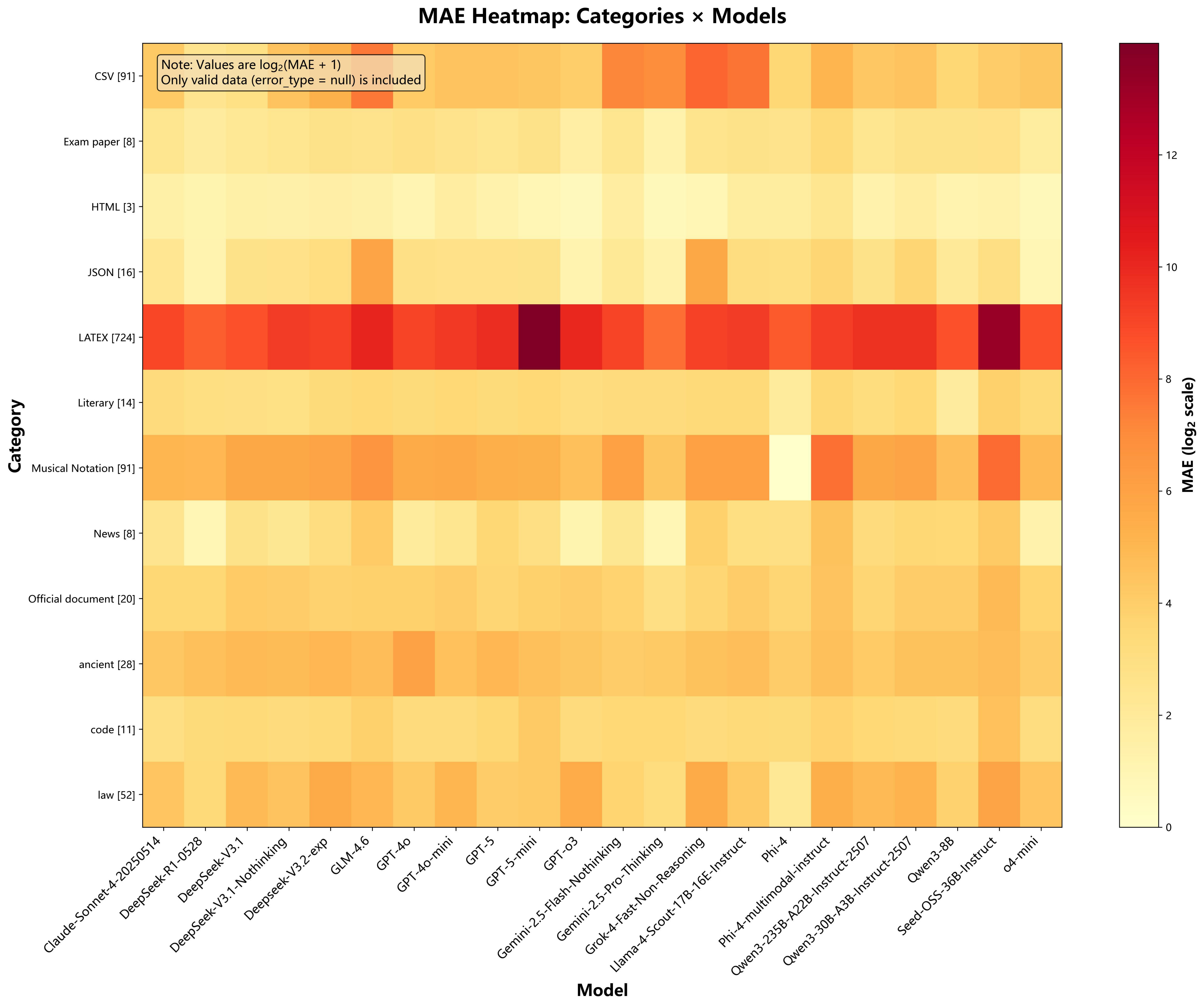}
    \caption{\textbf{Category-level MAE heatmap for text-based counting}. 
    Categories differ widely in structural complexity, causing substantial variation in numerical deviation.}
    \label{fig:text_mae_heatmap}
\end{figure}

The category-level heatmap in Figure~\ref{fig:text_mae_heatmap} further highlights this structural sensitivity: categories with rigid syntax (such as LATEX or code) exhibit significantly higher MAE, indicating that even small parsing failures can cascade into large counting mistakes. In contrast, lightweight formats (e.g., short news snippets or CSV-style items) yield relatively lower errors. Overall, these results demonstrate that textual counting difficulty is dominated by reasoning depth and structural complexity rather than document length alone.

\subsubsection{Audio Modality}

Counting in the audio modality presents fundamentally different failure modes from visual or textual settings, because temporal signals introduce ambiguity that models cannot easily resolve.
Across all evaluated systems, numerical errors remain substantial even when the ground-truth counts are typically small (mostly single-digit). This indicates that mistakes arise not from scale but from the intrinsic difficulty of segmenting acoustic events.

\begin{figure}[t]
    \centering
    \includegraphics[width=\columnwidth]{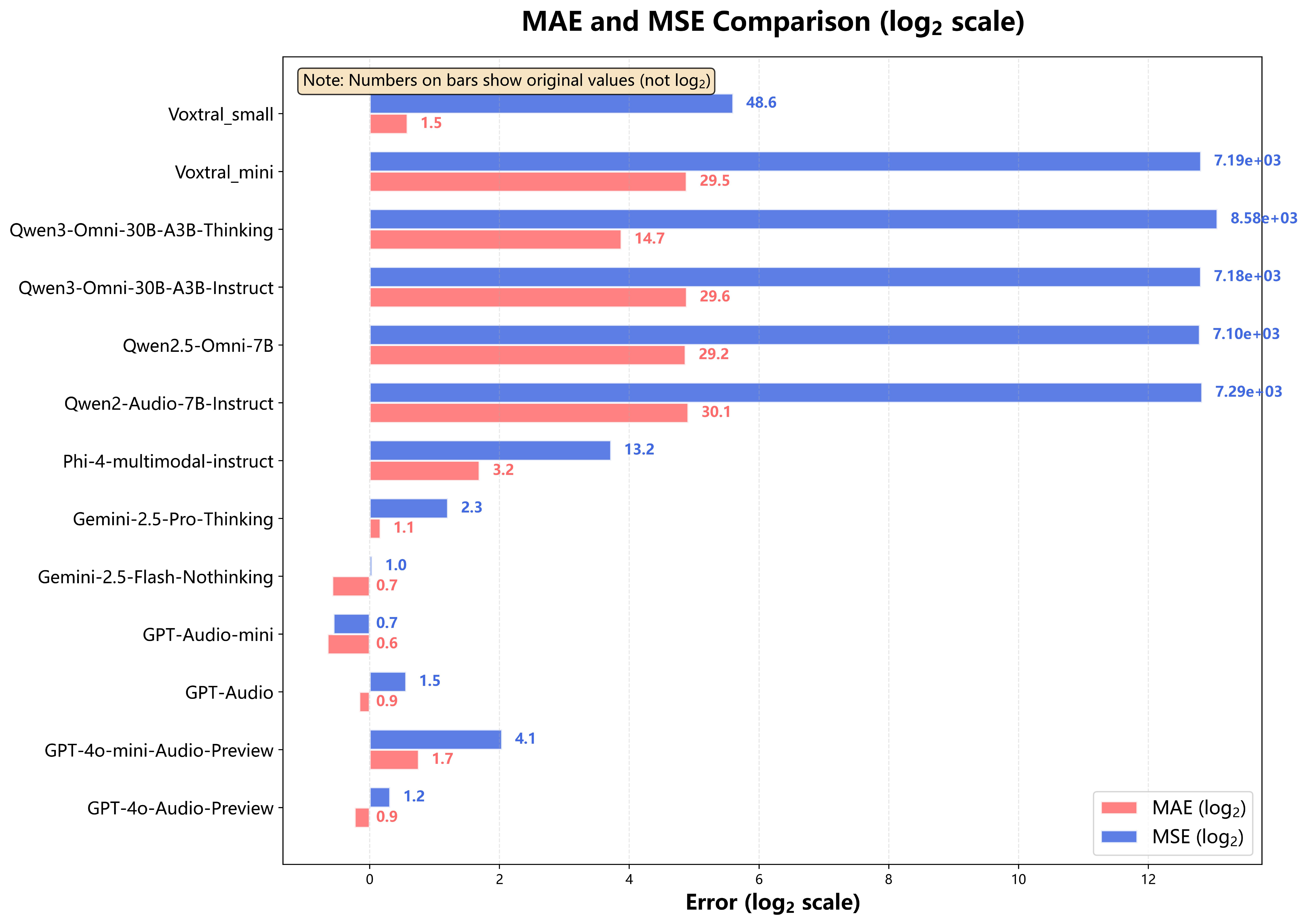}
    \caption{\textbf{MAE and MSE comparison for the audio modality} (log$_2$ scale). 
    Despite small ground-truth counts, numerical deviations remain large across many models.}
    \label{fig:audio_mae_mse}
\end{figure}

\begin{figure}[t]
    \centering
    \includegraphics[width=\columnwidth]{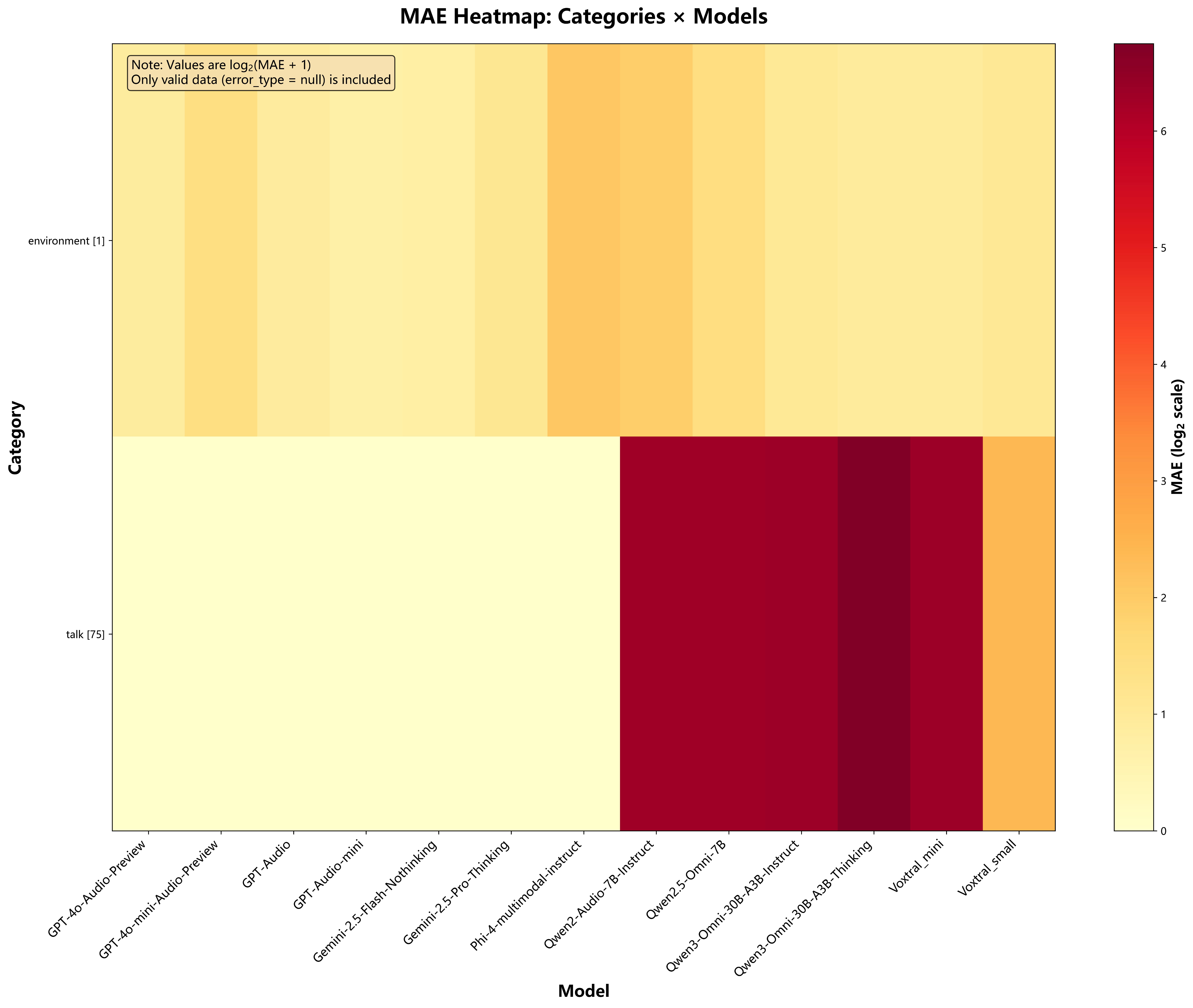}
    \caption{\textbf{Category-level MAE heatmap for audio counting}. 
    Environmental sounds are relatively easier, while conversational speech produces disproportionately large errors.}
    \label{fig:audio_error_heatmap}
\end{figure}

Figure~\ref{fig:audio_mae_mse} shows that MAE and MSE vary dramatically across models, with several systems exhibiting large deviations even on simple event-counting clips. These errors stem from temporal overlap between events, variable speaking rates, and model sensitivity to background noise.
At the category level (Figure~\ref{fig:audio_error_heatmap}), we observe a clear performance gap: \emph{environmental} sounds remain relatively manageable for most models, while the \emph{talk} category induces disproportionately large errors, suggesting that conversational audio—with its irregular pauses and overlapping utterances—is significantly harder for current MLLMs to decompose into discrete countable units. Overall, these findings highlight that temporal ambiguity, rather than numerosity magnitude, is the dominant driver of counting errors in audio.

\end{document}